%% file: main.tex
\PassOptionsToPackage{numbers, compress}{natbib}

\documentclass{article}

\usepackage[main, final]{neurips_2026}

\usepackage[utf8]{inputenc}
\usepackage[T1]{fontenc}
\usepackage{microtype}

\usepackage{xcolor}
\usepackage{url}
\usepackage{hyperref}
\usepackage{enumitem}
\usepackage{xspace}

\usepackage{amsmath}
\usepackage{amssymb}
\usepackage{amsfonts}
\usepackage{amsthm}
\usepackage{mathtools}
\usepackage{nicefrac}
\usepackage{bm}

\usepackage{graphicx}
\usepackage{booktabs}
\usepackage{multirow}
\usepackage{diagbox}
\usepackage{float}
\usepackage{caption}
\usepackage{subcaption}
\usepackage{tikz}
\usepackage{pgfplots}

\usepackage{algorithm}
\usepackage{algorithmic}

\usepackage[capitalize,noabbrev]{cleveref}

\usepackage[textsize=tiny]{todonotes}

\theoremstyle{plain}
\newtheorem{theorem}{Theorem}[section]
\newtheorem{proposition}[theorem]{Proposition}
\newtheorem{lemma}[theorem]{Lemma}
\newtheorem{corollary}[theorem]{Corollary}
\theoremstyle{definition}

\theoremstyle{remark}
\newtheorem{remark}[theorem]{Remark}
\theoremstyle{definition}
\newtheorem{example}{Example}

\title{Reformulating Neural Operators in $d+1$ Dimensions \\
for Embedding Evolution}

\author{%
  Haoze Song$^{1}$\quad Zhihao Li$^{1}$\quad Xiaobo Zhang$^{3}$\quad
  Zecheng Gan$^{1,2}$ \quad Zhilu Lai$^{1,2,\dagger}$\quad Wei Wang$^{1,2,\dagger}$\\
  $^{1}$The Hong Kong University of Science and Technology (Guangzhou)\\
  $^{2}$The Hong Kong University of Science and Technology\\
  $^{3}$Southwest Jiaotong University\\
  \texttt{\{hsong492, zli416\}@connect.hkust-gz.edu.cn}\quad
  \texttt{zhangxb@swjtu.edu.cn}\\
  \texttt{\{zechenggan, zhilulai, weiwcs\}@ust.hk}
}

\begin{document}
\maketitle

\renewcommand{\thefootnote}{\fnsymbol{footnote}}
\footnotetext[2]{Corresponding authors.}   
\renewcommand{\thefootnote}{\arabic{footnote}}


\input{Sections/0_Abstract}

\input{Sections/1_Introduction}

\input{Sections/2_d_plus_1_dimensional_operator_learning}

\input{Sections/3_SKNO}

\input{Sections/4_Experiments}

\input{Sections/5_Conclusion}

\bibliographystyle{plainnat}
\bibliography{reference}

\input{checklist}

\newpage
\appendix

\input{Appendix/1_Claim_of_LLM}

\input{Appendix/2_Experiment_Details}

\input{Appendix/5_Related_Work}

\input{Appendix/4_Proof}

\input{Appendix/0_Table_of_Notations}

\input{Appendix/3_Visualization}


\end{document}

%% file: Sections/0_Abstract.tex
\begin{abstract}

Neural Operators (NOs) are powerful architectures for learning mappings between function spaces. While most advances focus on refining kernel parameterizations over the $d$-dimensional physical domain, the evolution of lifted embeddings remains underexplored, which often drives models toward computationally expensive embedding-scaling designs to improve approximation.

In this paper, we introduce an auxiliary function dimension that models embedding evolution in operator form, thereby reformulating the NO pipeline in $d+1$ dimensions. We instantiate this framework via Fourier-based operators acting jointly on the physical and auxiliary domains, yielding a basis-diversified auxiliary evolution module as an alternative to brute-force embedding scaling. Across more than ten increasingly challenging benchmarks, ranging from the 1D heat equation to the highly nonlinear 3D Rayleigh–Taylor instability, our model consistently achieves the lowest relative $L_2$ error among the evaluated baselines. Crucially, this advantage is empirically supported by (1) controlled budget-aware comparisons against scaled and ablated baselines; (2) robustness under mixed-resolution training and super-resolution inference; and (3) zero-shot generalization to unseen temporal regimes. In addition, we present a broader set of design choices for lifting and recovery operators, demonstrating their impact on our model’s predictive performance.

\end{abstract}

%% file: Sections/1_Introduction.tex

\section{Introduction} \label{Section_1}

Neural Operators (NOs) \cite{20:GNO, 21:fno, 22:transformer_2, 23:function_space_no, 23:LSM, 23:geofno, 23:uno, 24:NMO, 24:local_fno,25:AMG} parameterize mappings between function spaces. As a new paradigm for learning solution operators of Partial Differential Equations (PDEs), they often improve accuracy relative to standard neural networks \cite{20:GNO, 25:AMG}.

A natural function-space analogue of matrix multiplication in standard Neural Networks \cite{89:universal_nn, 98:ANN_for_ODEPDE} is kernel integration. Concretely, NOs define kernel integrals acting on embedding functions defined on the $d$-dimensional physical domain, whose pointwise values are lifted from the input at each location \cite{20:GNO}. Building on this formulation, successive variants have improved the parameterization of these kernel integrals through richer aggregation mechanisms, such as graph aggregation \cite{20:MGNO, 25:AMG}, spectral convolution \cite{21:fno, 22:ufno}, and attention \cite{24:Transolver, 24:ONO}.

However, when the lifted embedding is not sufficiently expressive, existing architectures \cite{23:oFormer, 24:Transolver, 25:AMG} typically compensate by enlarging the embedding width or adding heads,while leaving the mechanism by which embeddings evolve largely unchanged. This strategy becomes computationally expensive as the embedding size grows, since dense channel mixing scales quadratically with the embedding width, while head-wise factorization only partially alleviates this cost, since it induces a block-diagonal interaction structure that weakens direct cross-head coupling \cite{17:transformer, 21:Choose}.

As a result, improving expressivity often relies on brute-force embedding enlargement rather than a more direct design of how embeddings evolve. This raises a natural question:

\textit{Can we improve how embeddings evolve by applying the operator-design principle along an auxiliary embedding coordinate, instead of merely increasing the embedding width?}

\input{Figures/code/heat_equation}

In this paper, we answer this question by introducing an auxiliary embedding dimension $p$, and reformulating the latent NO pipeline on the $(d+1)$-dimensional product domain: instead of only updating embeddings over the physical domain, the proposed formulation evolves latent functions over both the physical and the auxiliary embedding coordinate. 

This auxiliary-domain propagation provides an explicit operator-based mechanism for embedding evolution. In our Fourier-based instantiation, the auxiliary profile is updated through two coordinate views: raw \(p\)-coordinate mixing and Fourier-\(p\)-coordinate mixing. We refer to this as basis-diversified auxiliary
evolution, and evaluate it through controlled scaled and ablated baselines. Specifically, our main contributions are summarized as follows:

\begin{itemize}[topsep=0pt, leftmargin=2.3em]

    \item We reformulate the latent NO pipeline to evolve via kernel integral operators over both physical and auxiliary coordinates, establishing an operator-level mechanism for embedding evolution. We instantiate this framework with Fourier-based kernels into a specific model, the Schrödingerised Kernel
    Neural Operator (SKNO), whose auxiliary update mixes the lifted embedding in both raw-\(p\) and Fourier-\(p\) coordinates (dual-basis auxiliary operator).

    \item We evaluate SKNO across more than ten benchmarks spanning linear, nonlinear, high-dimensional, and spatio-temporal dynamics. Across these tasks, SKNO consistently achieves the lowest relative \(L_2\) error among the evaluated baselines. Controlled comparisons accounting for parameters, FLOPs, and wall-clock time, together with ablated and scaled ablations, indicate that the gain is not explained by additional same-basis capacity alone. Further tests on zero-shot super-resolution, temporal generalization, lifting, recovery, and auxiliary-domain operators support the robustness and scalability of the proposed framework.
\end{itemize}

%% file: Figures/code/heat_equation.tex

\begin{figure}
    \centering
    {\includegraphics[width=\linewidth]{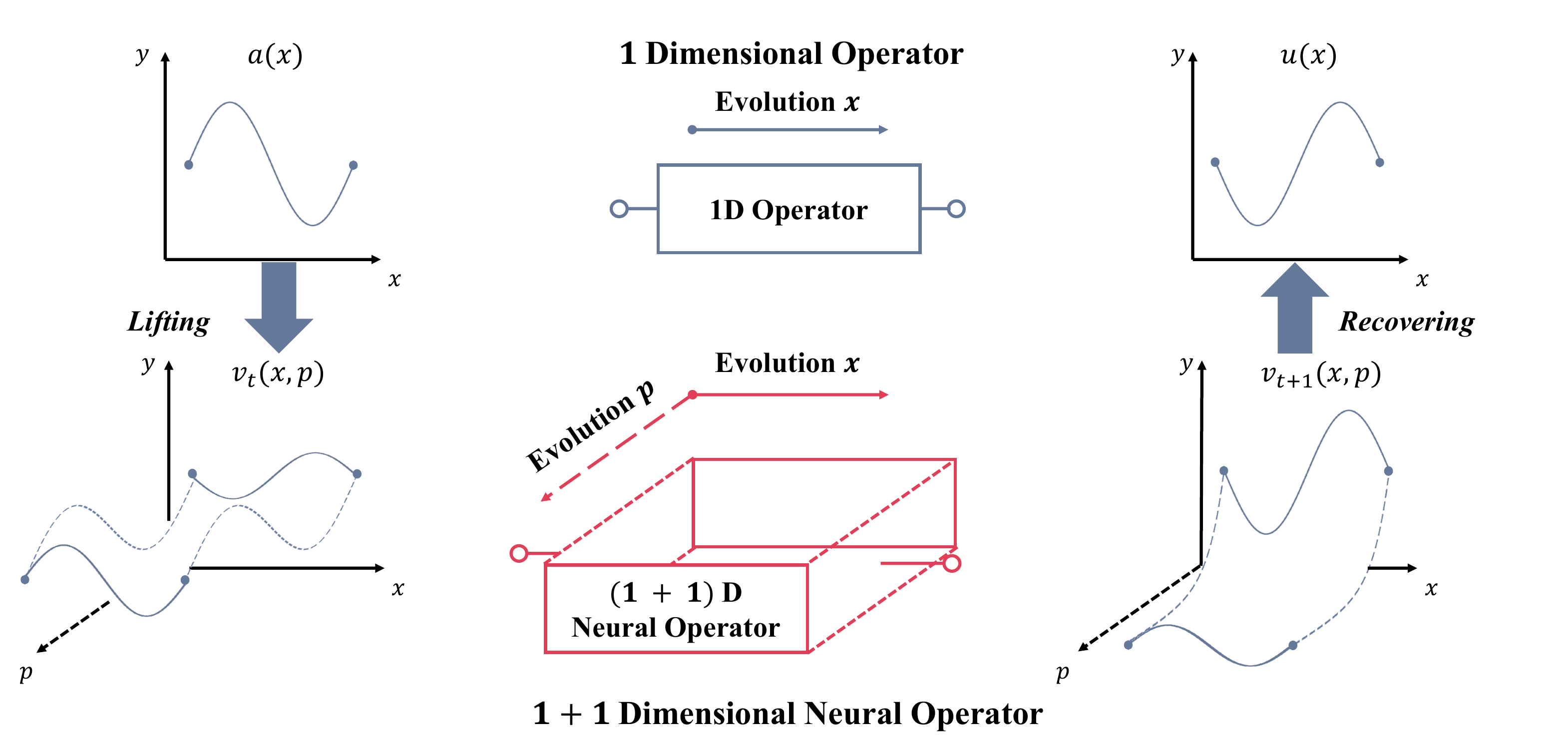}}
    \caption{Different $d$-dimensional operators can be parameterized by the $(d+1)$-dimensional Neural Operator which flexibly represents the evolution on both $x$ and $p$.} \label{fig1:intro}
\end{figure}

%% file: Sections/2_d_plus_1_dimensional_operator_learning.tex
\newpage

\section{Neural Operator in $d+1$ Dimensions} \label{Section_2}

\paragraph{Problem setup.}
Let \(D_x \subset \mathbb{R}^d\) be a bounded physical domain. Given an input field
\(a(x) \in \mathcal{A}(D_x;\mathbb{R}^{d_a})\), our goal is to learn an operator
\begin{equation}
    \mathcal{G}: a \mapsto u = \mathcal{G}[a],
\end{equation}
where \(u(x) \in \mathcal{U}(D_x;\mathbb{R}^{d_u})\). Here \(x=(x_1,\ldots,x_d)\)
denotes the physical coordinate, and $\mathcal{A}$ and $\mathcal{U}$ are (suitable subsets of) Banach spaces, where functions $a$ and $u$ take values in $\mathbb{R}^{d_{{a}}}$ and $\mathbb{R}^{d_{{u}}}$ ($d_a, d_u \in \mathbb{N}$), respectively.

\paragraph{Auxiliary function dimension.} Previous neural operators first lift the input to a vector-valued embedding field, and then evolve it over the physical domain \(D_x\). In our framework, we take a different view. We introduce an auxiliary domain $D_p \subset \mathbb{R}$ and lift the input to a scalar latent function $v_0(x,p) \in \mathcal{V}(D_x \times D_p;\mathbb{R})$. The variable \(p\) is an auxiliary embedding coordinate. Figure~\ref{fig6:compare} summarizes this design distinction between previous NO variants and ours.

\paragraph{Lifting.} To prepare the $(d+1)$-dimensional scalar function $v(x,p) \in \mathcal{V}(D_x \times D_p; \mathbb{R)}$ for subsequent evolving, the lifting operator \(\mathcal{P}\) first maps the input field \(a(x)\) to a latent function on the product domain \(D_x \times D_p\):
\begin{equation}
    v_0(x,p) := \mathcal{P}[a](x,p).
\end{equation}
A simple separated lifting is
\begin{equation}\label{Eq_sep_map}
    v_0(x,p) = w^\top(p) a(x),
\end{equation}
where \(w(p) \in \mathbb{R}^{d_a}\) is usually learnable. If positional features are used, they are concatenated with \(a(x)\) before lifting.

\paragraph{$(d+1)$-dimensional evolution.} The lifted function is evolved by iterating learnable linear operators $\mathcal{L}$ and nonlinear maps $\sigma$, where the core block is the kernel integral operator $\mathcal{K}$. For the $l$-th updated latent function \(v_l(x,p)\), we define the kernel integral operator on \(D_x \times D_p\) as follows:
\begin{equation}\label{Eq_kernel}
    \mathcal{K}_l[v_l](x,p)
    :=
    \int_{D_x}\int_{D_p}
    \kappa_l(x,y,p,p')\,v_l(y,p')\,dp'\,dy ,
\end{equation}
where \(\kappa_l: \mathbb{R}^{2(d+1)} \to \mathbb{R}\) is the $l$-th learnable kernel function and $l = 0, 1, \cdots, L-1$ for $\kappa$.

\paragraph{Recovery.}
After the latent function has evolved, the recovery operator \(\mathcal{Q}\) maps it back
to the original output domain through:
\begin{equation}
    u_{\text{pred}}(x)
    :=
    \mathcal{Q}[v_L](x)
    =
    \int_{D_p} \chi(p)\,v_L(x,p)\,dp ,
\end{equation}
where \(\chi(p) \in \mathbb{R}^{d_u}\) is usually learnable.

Given all the definitions introduced above, we assemble our neural operator framework to approximate $\mathcal{G}$ in the problem setup as follows:
\begin{align}\label{Equ_10} 
    \mathcal{G}[{a}] &\approx (\mathcal{Q} \circ \mathcal{M} \circ \mathcal{P})[{a}] \nonumber \\
    &= (\mathcal{Q} \circ \left(\sigma_{L-1} \circ \mathcal{L}_{L-1}) \circ \cdots \circ (\sigma_0 \circ \mathcal{L}_0\right) \circ \mathcal{P})[{a}],
\end{align}
Here, each $\mathcal{L}_l$ denotes a general learnable $(d+1)$-dimensional linear operator, and $\sigma_l$ can be a nonlinear activation function or a shallow MLP, providing the nonlinear composition needed to approximate nonlinear target operators \(\mathcal{G}\).

%% file: Sections/3_SKNO.tex
\newpage

\section{Schrödingerised Kernel Neural Operator} \label{Section_3}

We now instantiate the $(d+1)$-dimensional framework in Section~\ref{Section_2}
with a Fourier-based model, called the Schrödingerised Kernel Neural Operator
(SKNO). Specifically, SKNO evolves the lifted latent function \(v(x,p)\) by using Fourier-based operators over the physical coordinate \(x\) and the
auxiliary \(p\).

\textit{Naming.} The term ``Schrödingerised'' reflects the use of Fourier-domain propagation on the introduced auxiliary phase-like coordinate as in the Schrödingerised solver. In our work, Schrödingerisation only serves as design inspiration for building structured operator evolution along the auxiliary coordinate.

An overview of the SKNO architecture is shown in Figure ~\ref{fig:arc_D_x}. Beyond the lifting and recovering modules, SKNO is built from signal propagators that instantiate the kernel integration in Eq.~\eqref{Eq_kernel} through structured physical-domain and auxiliary-domain updates, together with bias terms and residual connections.

\input{Figures/code/architecture}

\paragraph{Lifting and Recovery.}
We consider several choices for \(\mathcal{P}\) and \(\mathcal{Q}\), and evaluate their effects in Section~\ref{Section_4_3_2}. Unless otherwise stated, SKNO uses a pointwise linear layer for the separated lifting map in Eq.~\eqref{Eq_sep_map}, which balances representation flexibility and computational efficiency. For recovery, we use an MLP-based projection whose final linear layer implements the quadrature over \(D_p\). This choice keeps the lifting/recovering modules comparable to common NO implementations, so that the main architectural difference lies in the proposed \((d+1)\)-dimensional evolution.

\paragraph{Signal propagation on $D_x$.} \label{Section_3_1_2} As illustrated in Figure~\ref{fig:arc_D_x}, we employ $(L-1)$ global propagators and one last local propagator for modeling the underlying $(d+1)$-dimensional function evolution on $D_x$.

\textit{Spectral Convolution Operator.} \label{Section_3_1_3} Differential operators with respect to $x$, such as \(\partial_x\) and \(\partial_{xx}\), can be diagonalized in Fourier domain of $x$ according to the properties of Fourier Transform (FT). Based on that, we leverage the implementation of spectral convolution operator in \cite{21:fno}, which uses a learnable complex vector equivalent to a diagonal matrix in the Fourier domain of $D_x$. In practice, the complex vector is truncated for working on different grid sizes: \begin{equation}  \label{Equ_11}
    \mathrm{SpectralConvOp}(\boldsymbol{v}_l) = F_x^{-1}(\mathrm{trun\_diag}(F_x(\boldsymbol{v}_l))),
\end{equation} where $\boldsymbol{v}_l$ denotes the input latent signal fitting $v(x,p)$, and $F_x$ and $F_x^{-1}$ denote Discrete Fourier Transform (DFT) and Inverse DFT (accelerated by Fast Fourier Transform (FFT) in practice). The $\mathrm{trun\_diag}()$ represents the element-wise multiplication between a learnable truncated complex vector and signal slices of $\boldsymbol{v}_l$ on the transformed grid $\xi$ after $F_x$.

\textit{Differential Operator.} \label{Section_3_1_4} Since the truncation in \textit{Spectral Convolution Operator} prefers to propagate signals in a relatively global scope, we compensate for possible local information lost by propagating the finite difference derivative of $x$ on the local region based on \cite{24:local_fno}:

\begin{align}\label{Equ_12}
\mathcal{K}_{L-1}[v_{L-1}]({x},p) &= \frac{\iint_{D_{loc} \times D_p} \kappa_{L-1}({x},{y}, p, p')(v_{L-1}({y},p')-v_{L-1}({x},p'))d{y}dp'}{\iint_{D_{loc} \times D_p} \kappa_{L-1}({x},{y}, p, p')({y}-{x},p')d{y}dp'},
\end{align}

where $D_{loc} \subset D_x$ denotes the local region around the query position $x$. Related ablation results are reported in Tab.~\ref{tab:ablation} and Tab.~\ref{tab:darcy_local_props}.

\paragraph{Evolution along the auxiliary dimension} \label{Section_3_2}

For each spatial location \(x_i=(x_{i_1},\ldots,x_{i_d})\), the discretized signal along auxiliary dimension \(v_{l}(x_i, \cdot)\in\mathbb R^{N_p}\) is updated by a basis-diversified auxiliary operator:
\[
    v_{l}(x_i, \cdot)
    \mapsto
    \big(F_p^{-1}\widetilde A_lF_p+B_l\big)v_{l}(x_i, \cdot),
\]
where \(F_p\) is the DFT matrix along the auxiliary axis, \(\widetilde A_l\) mixes the signal in Fourier-\(p\) coordinates, and \(B_l\) mixes it in the raw \(p\)-coordinates. This design introduces a finite-budget architectural bias: the two branches parameterize the auxiliary evolution through different coordinate views, and subsequent nonlinear/residual updates can exploit features formed from both views.

\paragraph{Linear block with residuals.} \label{Section_3_1_5} With the above two implementations of $(d+1)$-dimensional kernel integral operators, we specified the linear block $\mathcal{L}$ in Eq.\ref{Equ_10} with two parts, kernel integral operator $\mathcal{K}_l$ and residual operator $\bar{\mathcal{W}}_l$:
\begin{align} \label{Equ_13}
    \mathcal{L}_l[v_{l}]({x}, p) &= ({\mathcal{K}}_l + {\bar{\mathcal{W}}}_l)[v_l]({x}, p),
\end{align} where $\bar{\mathcal{W}}_l[v_l](x,p)= (\mathcal{I} + \mathcal{W}_l)[v_l](x,p) = v_l(x,p) + \int_{D_p} w_l(p,p')v_l({x},p')dp'$. Here, $\mathcal{W}_l$ is implemented by a shallow MLP and $\mathcal{I}$ is implemented by a residual connection. If moving the residual connection from right to left, Eq.~\ref{Equ_13} implements a forward finite-difference (Euler) update.

In the discrete forward pass, all paths (Figure~\ref{fig:skno_flow})
and the identity are summed before activation, the local propagator (Eq. \ref{Equ_12}) acts jointly on
$x$ and $p$ once at $l = L-1$, and the only exact commutation used is $F_xF_p = F_pF_x$.

\input{Figures/code/evolution_p}

As illustrated in Figure~\ref{fig:evolution_p}, while standard channel-mixing designs update the embedding through a single coordinate view, SKNO updates the auxiliary profile through both raw-\(p\) and Fourier-\(p\) views. By iteratively combining physical-domain propagation with this auxiliary-domain update, SKNO instantiates the \((d+1)\)-dimensional operator framework while keeping the additional cost controlled through FFT-based operations. To help readers connect the above formulations to the implementation, we further clarify the discrete instantiation of the linear blocks in Section \ref{subsec_Model_plot}.

%% file: Figures/code/architecture.tex
\begin{figure*}[ht]
    \centering
    \includegraphics[width=0.8\textwidth]{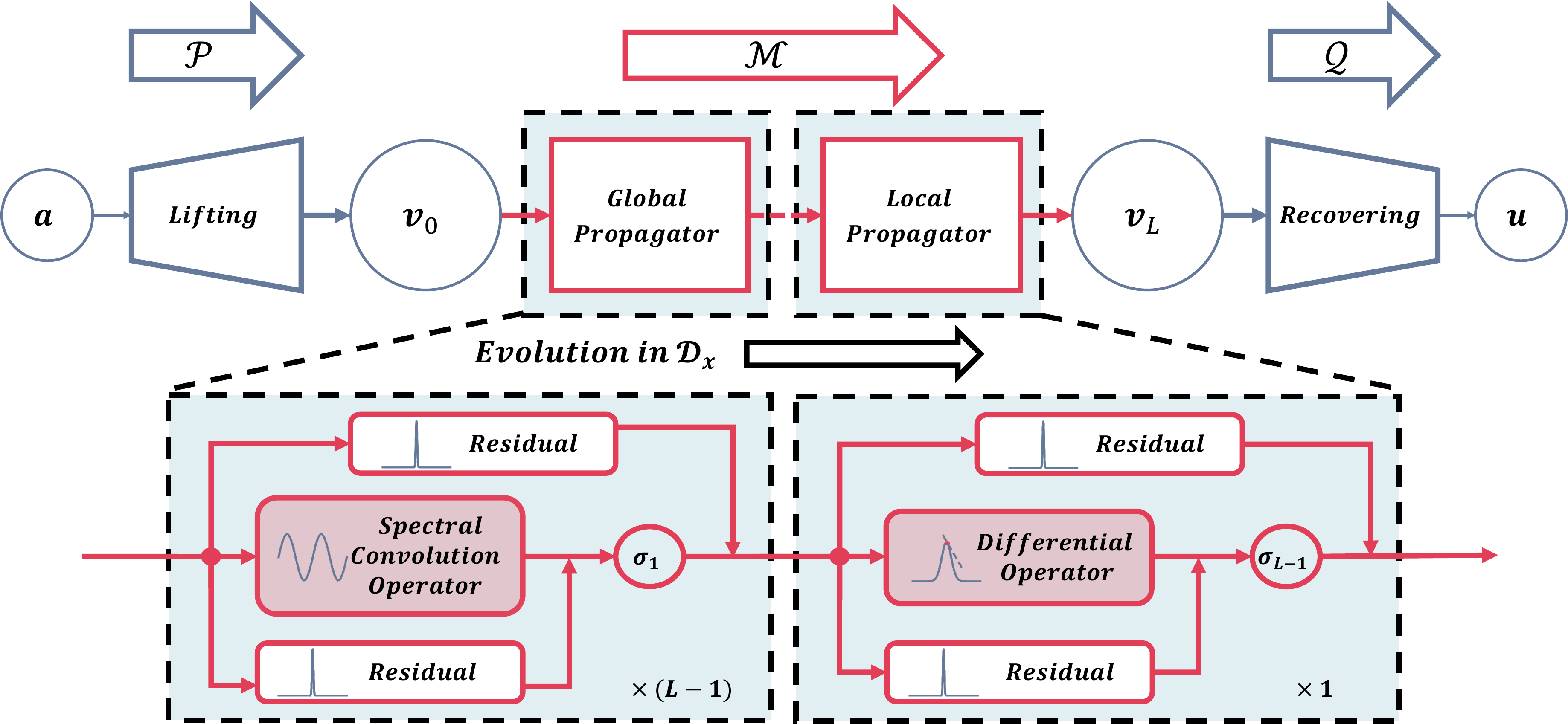}
    \caption{Overview of the Schrödingerised Kernel Neural Operator (SKNO) architecture.  The input signal $\boldsymbol{a}$ is lifted to a $(d+1)$-dimensional signal $\boldsymbol{v}_0$ through the preparation lifting module. Such a $(d+1)$-dimensional signal evolves through $(L-1)$ global propagators and one local propagator in $D_x$, capturing both global and local signal changes. The measuring module recovers the output signal $\boldsymbol{u}$ from evolved $\boldsymbol{v}_L$.}
    \label{fig:arc_D_x}
\end{figure*}

%% file: Figures/code/evolution_p.tex
\begin{figure*}
    \centering
    \includegraphics[width=\textwidth]{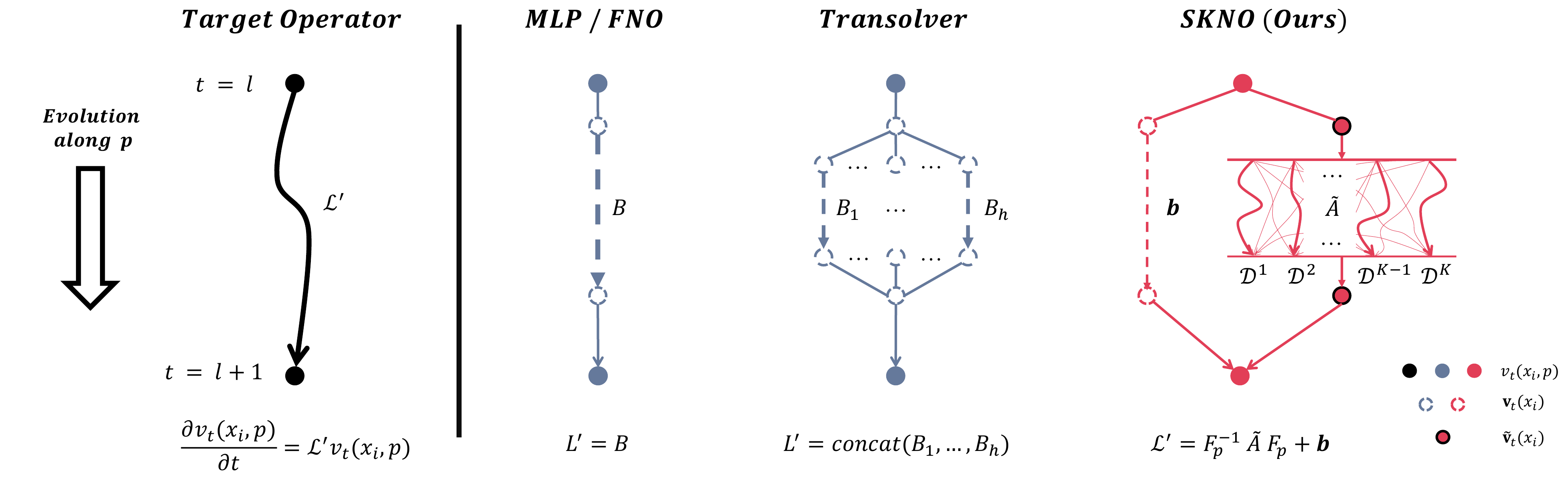}
    \caption{Evolution along the auxiliary dimension and corresponding neural operator implementations. \textit{Left}: The target operator $\mathcal{L}'$ governs the evolution along $p$. \textit{Right}: Whereas prior methods rely on discretized matrices at each $p$ location, our model captures this evolution using operator modules acting on dual spaces.}
    \label{fig:evolution_p}
\end{figure*}

%% file: Sections/4_Experiments.tex
\newpage

\section{Experiments} \label{Section_4}

\subsection{General Setting}  \label{Section_4_1}

\paragraph{Benchmarks} \label{Section_4_1_1} 
We evaluate the proposed model on more than ten diverse benchmarks spanning a wide spectrum of PDE dynamics. To highlight the effectiveness of our evolving kernel integral design, we first compare different kernel parameterizations on the PDE operators including the 1D heat equation~\cite{22:PDNO} and the 1D advection equation~\cite{22:PDEBench}. We then evaluate full models on a broader set of PDE systems, conducting extensive experiments on PDEs including the 1D Burgers equation; 2D Darcy flow with discontinuous coefficients; 2D incompressible Navier–Stokes at low viscosity ($1e{-5}$)~\cite{21:fno, 24:Transolver}; 2D Gray–Scott Reaction–Diffusion~\cite{24:boosting}; and the highly unstable 3D Rayleigh–Taylor Instability~\cite{24:thewell}. A summary of all benchmarks is in Table~\ref{tab20:sum_bench}. Per-benchmark details and additional results are provided in Appendix~\ref{appendix:bmandresults}.

\paragraph{Baselines and Configurations} \label{Section_4_1_2} 
We compare our model with various $d$-dimensional neural implementations, including DeepONet \cite{21:deeponet_new}, Fourier Neural Operator (FNO) \cite{21:fno}, Convolutional Neural Operator (CNO) \cite{23:CNO}, and transformer-based model Transolver \cite{24:Transolver}. All models are trained using the recommended or reported configurations from their papers and official implementations when available. For benchmarks such as 1D Burgers, 2D Darcy Flow, and 2D Gray–Scott, we train each model for 500 epochs with no physical prior regularization, using only relative $L_2$ loss. Because the compared architectures differ in parameter count, FLOPs, and training speed, we do not rely solely on epoch-matched comparisons. For the incompressible 2D Navier–Stokes, we ensure fairness by matching comparable training time scales, accounting for differences in training speed between transformer-based and other operator architectures. All experiments are run on a single \emph{NVIDIA GeForce RTX 4090} GPU. Complete configurations are provided in Table~\ref{tab:conf_supple}.

\input{Tables/benchmark}

\subsection{Main Results and Analysis}

\paragraph{Results} \label{Section_4_2_1}  
SKNO attains the lowest relative \(L_2\) error among the compared baselines across the evaluated benchmarks. Table~\ref{tab:Linear_blocks} summarizes the performance of different $\mathcal{L}$ parameterizations across representative models: the Fourier integral operator in FNO, the physics-attention block in Transolver, and SKNO’s spectral convolution operator acting on $(d+1)$-dimensional functions. All modules are evaluated with their original lifting and recovering components and without additional positional encoding (see Appendix~\ref{appendix:baseline}).

For the heat equation experiment, the solution at $t=1$ is predicted with the same truncation settings as in practice (i.e., truncated modes or slices, both of which parameterize aggregation on the retained coefficient or slice subspace). For the advection equation, these kernels are tested in an autoregressive setting to validate performance on sequential prediction with concatenated input quantities; the reported error corresponds to predicting the next 10 time steps autoregressively given the previous 10.

Relative to FNO, SKNO yields an approximately \(37.1\%\) relative \(L_2\) error reduction on 2D incompressible Navier–Stokes (\(\nu=10^{-5}\)) using the six-seed mean, and a \(42.1\%\) reduction on 2D Gray–Scott relative to the default FNO configuration. On 3D Rayleigh–Taylor instability, SKNO attains an approximately \(14.3\%\) error reduction, while the compared baselines remain in a similar error range. Across the benchmarks in Table~\ref{tab:main_result}, Table~\ref{tab:other_exp_bench} and Table~\ref{tab:era5}, SKNO attains the lowest reported error. Visualizations in Figure~\ref{fig12:GS} and Figure~\ref{fig:showcase_darcy} further illustrate that SKNO better captures sharp local structures, such as discontinuities at material interfaces and vorticity patterns in low-viscosity fluid flows. For complete experimental settings, case studies, and full results, see Appendix~\ref{Appendix_Experiment_Settings}.

In addition to the main benchmark results, we report parameter counts, FLOPs, wall-clock training time, and controlled scaled/ablated variants where applicable.

\paragraph{Capacity efficiency.}
To test whether the improvement comes merely from additional embedding capacity, we compare SKNO against systematically scaled FNO variants, parallel stacked FNO variants, and a same-basis SKNO ($B+B$) control. The $B+B$ variant keeps the dual-branch structure but replaces the Fourier-\(p\) branch with a duplicate raw-\(p\) branch. Thus, $B+B$ controls for extra same-basis capacity, while $A+B$ tests the basis-diversified auxiliary update. As shown in Table~\ref{tab11:capacity_efficiency}, SKNO($A+B$) achieves the lowest error with fewer parameters and FLOPs than the best scaled FNO configuration, while SKNO($B+B$) fails to match $A+B$ on 2D Gray--Scott. This supports the view that the gain is not explained by widening or duplicating the same channel-mixing path alone.

\input{Tables/linear_compare}

\paragraph{Resolution Invariance} \label{Section_4_3_1} 
A key advantage of neural operators is their ability to generalize across discretizations. To evaluate this property, we design experiments where training data of mixed resolutions are combined. Table~\ref{tab:other_exp_bench} shows that SKNO consistently achieves the best accuracy, even under randomly mixed training resolutions, demonstrating robustness. In addition, we further evaluate super-resolution performance on the 1D Burgers equation, 2D Darcy flow, and the challenging ERA5 wind field prediction task. All models are trained on a fixed resolution and tested on lower and higher ones. As shown in Tables~\ref{tab14:superes_table} and~\ref{tab:superes_table_era5}, SKNO consistently achieves lower relative $L_2$ errors across resolutions than competing models, whereas the others exhibit substantial error fluctuations as resolution increases. This indicates their instability in generalizing across resolutions. In contrast, SKNO delivers more stable and reliable predictions across varying scales.


\input{Tables/error}

\paragraph{Different Lifting and Recovery Operators} \label{Section_4_3_2}  
We explore a broader set of lifting and recovery operators to test their impact on the output solution. Specifically, we fix $\mathcal{P}$ as a linear layer and vary the integration weights $\chi(p)$ for $\mathcal{Q}$. Table~\ref{tab:Q_implement} shows that all recovering designs perform well on the heat equation, where the Mean and Linear can also be properly scaled by the symmetry of $w(p)$. Some predefined recovery operators also remain effective on the more complex Darcy cases, suggesting that the learned \((d+1)\)-dimensional evolution can provide useful latent structure before the final integration over \(D_p\). Combining results from Table~\ref{tab:Q_implement} and Table~\ref{tab:P_Q_combination} on 2D Darcy, we find that for functions with complex topology on $D_x$, a nonlinear transformation on $(d+1)$-dimensional signals, either after the first lifting or before the last recovery, helps adjust the propagated \((d+1)\)-dimensional latent structure before recovery. The best performance with the MLP--Linear combination is consistent with the MLP--Linear implementation in Transolver \cite{24:Transolver} for handling complex topology.

\paragraph{Ablation} \label{Section_4_3_3}
We conduct a detailed ablation study on 2D Darcy (Table~\ref{tab:ablation}) to evaluate the contribution of different components. Removing both linear and nonlinear residual layers causes a significant performance degradation, indicating their importance for modeling complex \((d+1)\)-dimensional evolution. Omitting either the Fourier-\(p\) branch \(\widetilde A\) or the raw-\(p\) branch \(B\) also increases the error, suggesting that both coordinate views are useful in this setting. Finally, we separate the contributions of the global and local propagators: the global propagator captures long-range dependencies, while the local propagator refines predictions in regions with sharp transitions, as reflected by the increase in error when the corresponding component is removed.

\input{Tables/error_1}

\paragraph{Other Experiments} Additional experimental results and visualizations are provided in the Appendix, mainly including (i) zero-shot generalization to unseen temporal regimes: SKNO maintains the lowest error among the compared baselines on 2D incompressible Navier--Stokes in Figure~\ref{fig5:time_unseen}; and (ii) other operators choices along $p$: physics-attention \cite{24:Transolver} also benefits from adopting the $(d+1)$-dimensional operator design, see Table~\ref{tab12:darcy_NO_family}.

\input{Figures/code/zero_shot_time}

%% file: Tables/benchmark.tex
\begin{table*}[tb!]
    \centering
    \caption{Main experimental results across 1D, 2D, and 3D benchmarks. Each benchmark block reports Rel. $L_2$ Error and Training Time. For 2D Stress and 2D Strain, the training time is the reported average over the two benchmarks.}
    \label{tab:main_result}
    \renewcommand\arraystretch{1.15}
    \setlength{\tabcolsep}{4pt}

    \resizebox{\textwidth}{!}{%
    \begin{tabular}{l|cc|cc|cc|cc}
    \toprule
    \textbf{Model}
    & \multicolumn{2}{c|}{\textbf{1D Burgers}}
    & \multicolumn{2}{c|}{\textbf{2D Stress}}
    & \multicolumn{2}{c|}{\textbf{2D Strain}}
    & \multicolumn{2}{c}{\textbf{2D Darcy}} \\
    \cmidrule(lr){2-3}
    \cmidrule(lr){4-5}
    \cmidrule(lr){6-7}
    \cmidrule(lr){8-9}
    & Rel. $L_2$ & Time
    & Rel. $L_2$ & Time
    & Rel. $L_2$ & Time
    & Rel. $L_2$ & Time \\
    \midrule

    DeepONet
    & $8.991\mathrm{e}{-2}$ & \textbf{2.07 min}
    & $3.891\mathrm{e}{-1}$ & \textbf{4.15 min}
    & $5.439\mathrm{e}{-1}$ & \textbf{4.15 min}
    & $6.097\mathrm{e}{-2}$ & \textbf{5.20 min} \\

    FNO
    & $6.479\mathrm{e}{-4}$ & 6.86 min
    & $6.221\mathrm{e}{-2}$ & 9.43 min
    & $5.753\mathrm{e}{-2}$ & 9.43 min
    & $6.201\mathrm{e}{-3}$ & 10.21 min \\

    Transolver
    & $6.277\mathrm{e}{-3}$ & 94.23 min
    & $4.812\mathrm{e}{-2}$ & 117.39 min
    & $4.657\mathrm{e}{-2}$ & 117.39 min
    & $5.853\mathrm{e}{-3}$ & 115.48 min \\

    \textbf{SKNO}
    & $\mathbf{5.475\mathrm{e}{-4}}$ & 15.51 min
    & $\mathbf{4.724\mathrm{e}{-2}}$ & 19.36 min
    & $\mathbf{4.635\mathrm{e}{-2}}$ & 19.36 min
    & $\mathbf{5.555\mathrm{e}{-3}}$ & 17.59 min \\

    \midrule
    \midrule

    \textbf{Model}
    & \multicolumn{2}{c|}{\textbf{2D Gray-Scott}}
    & \multicolumn{2}{c|}{\textbf{2D Inc. NS}}
    & \multicolumn{2}{c|}{\textbf{3D Comp. NS}}
    & \multicolumn{2}{c}{\textbf{3D RT}} \\
    \cmidrule(lr){2-3}
    \cmidrule(lr){4-5}
    \cmidrule(lr){6-7}
    \cmidrule(lr){8-9}
    & Rel. $L_2$ & Time
    & Rel. $L_2$ & Time
    & Rel. $L_2$ & Time
    & Rel. $L_2$ & Time \\
    \midrule

    DeepONet
    & $1.528\mathrm{e}{-1}$ & \textbf{3.01 min}
    & $3.448\mathrm{e}{-1}$ & 865.75 min
    & $5.628\mathrm{e}{-1}$ & \textbf{3.90 min}
    & $5.731\mathrm{e}{-2}$ & \textbf{1.56 min} \\

    FNO
    & $2.425\mathrm{e}{-2}$ & 9.64 min
    & $1.280\mathrm{e}{-1}$ & 773.33 min
    & $2.631\mathrm{e}{-1}$ & 10.12 min
    & $5.219\mathrm{e}{-2}$ & 5.12 min \\

    Transolver
    & $3.573\mathrm{e}{-2}$ & 234.96 min
    & $1.002\mathrm{e}{-1}$ & 1263.33 min
    & $2.947\mathrm{e}{-1}$ & 361.41 min
    & $5.723\mathrm{e}{-2}$ & 135.03 min \\

    \textbf{SKNO}
    & $\mathbf{1.298\mathrm{e}{-2}}$ & 20.11 min
    & $\mathbf{8.717\mathrm{e}{-2}}$ & \textbf{708.93 min}
    & $\mathbf{2.376\mathrm{e}{-1}}$ & 23.33 min
    & $\mathbf{4.471\mathrm{e}{-2}}$ & 10.31 min \\

    \bottomrule
    \end{tabular}%
    }
\end{table*}

%% file: Tables/linear_compare.tex


\begin{table*}[t]
\centering

\begin{minipage}[t]{0.50\textwidth}
\centering
\caption{Different $\mathcal{L}$ parameterizations' performance on learning operators
governed by 1D Heat and Advection Equations.}
\label{tab:Linear_blocks}
\vspace{2mm}
\setlength{\tabcolsep}{6pt}
\renewcommand{\arraystretch}{1.05}

\resizebox{\linewidth}{!}{%
\begin{tabular}{lc}
    \toprule
    Linear Blocks     & Error       \\
    \midrule
    \multicolumn{2}{c}{1D Heat Equation} \\
    \midrule
       FNO (1 mode, 4 grid $p$) & 9.305e-1 \\
       Transolver (1 slice, 4 grid $p$ / 2 heads) & 9.060e-1 \\
       Transolver (4 slices, 4 grid $p$ / 2 heads) & 8.988e-1 \\
       Transolver (4 slices, 4 grid $p$ / 1 head) & 8.987e-1 \\
       \textbf{Ours (1 mode, 4 grid $p$)} & \textbf{2.637e-3} \\
    \midrule
    \multicolumn{2}{c}{1D Advection Equation} \\
    \midrule
       Transolver (8 slices, 16 grid $p$ / 4 heads) & 7.118e-1 \\
       Transolver (8 slices, 16 grid $p$ / 2 heads) & 6.984e-1 \\
       Transolver (8 slices, 16 grid $p$ / 1 head) & 6.828e-1 \\
       FNO (8 modes, 16 grid $p$) & 6.012e-2 \\
       \textbf{Ours (8 modes, 16 grid $p$)} & \textbf{1.979e-2} \\
    \bottomrule
\end{tabular}%
}
\end{minipage}%
\hfill
\begin{minipage}[t]{0.40\textwidth}
\centering
\caption{Ablation study for SKNO on 2D Darcy. The Linear Residual is $\bar{\mathcal{W}}_l$ in Linear Block $\mathcal{L}$, while the non-linear one is out of $\mathcal{L}$.}
\label{tab:ablation}
\vspace{2mm}
\setlength{\tabcolsep}{6pt}
\renewcommand{\arraystretch}{1.05}

\resizebox{\linewidth}{!}{%
\begin{tabular}{lc}
       \toprule
       Configuration & Error \\
       \midrule
       w.o. Double Res. & 1.363e-2 \\
       w.o. Linear Res. & 8.284e-3 \\
       w.o. Non-linear Res. & 7.621e-3 \\
       \midrule
       w.o. $\tilde{A}$ along $p$ dim & 8.804e-3 \\
       w.o. $\boldsymbol{b}$ along $p$ dim & 6.827e-3 \\
       \midrule
       w.o. Global Propagators & 1.584e-1 \\
       w.o. Local Propagator & 5.715e-3 \\
       \midrule
       \textbf{Baseline (SKNO)} & \textbf{5.555e-3} \\
       \bottomrule
\end{tabular}%
}
\end{minipage}

\end{table*}

%% file: Tables/error.tex



\begin{table*}[t]
\centering
\caption{The error on 1D Heat and 2D Darcy with different $\mathcal{Q}$ implementations.}
\label{tab:Q_implement}
\vspace{2mm}
\resizebox{0.85\textwidth}{!}{%
\begin{tabular}{l|cccccc}
\toprule
$\mathcal{Q}$ & Delta & Step & Mean & Linear & MLP & MLP (Dropout)\\
\midrule
\multicolumn{7}{c}{1D Heat Equation} \\
\midrule
Error & \textbf{2.538e-3} & \textbf{2.538e-3} & \textbf{2.538e-3} & \textbf{2.538e-3} & \underline{2.637e-3} & 7.928e-3 \\
\midrule
\multicolumn{7}{c}{2D Darcy Flow} \\
\midrule
Error & 5.752e-3 & 1.402e-2 & 5.717e-3 & 5.741e-3 & \textbf{5.555e-3} & \underline{5.678e-3} \\
\bottomrule
\end{tabular}%
}
\end{table*}

%% file: Tables/error_1.tex
\begin{table*}[ht]
\centering
\caption{The error on 2D Darcy with different $\mathcal{P}$, $\mathcal{Q}$ combinations.}
\label{tab:P_Q_combination}
\vspace{2mm}
\resizebox{0.65\textwidth}{!}{%
\begin{tabular}{c|cccc}
\toprule
\diagbox{$\mathcal{Q}$}{$\mathcal{P}$} 
& Constant & Linear & MLP & MLP (Dropout) \\
\midrule
Linear 
& 8.738e-3 & 5.741e-3 & 5.952e-3 & \textbf{5.451e-3} \\
MLP 
& 8.791e-3 & \textbf{5.555e-3} & 6.614e-3 & 5.845e-3 \\
\bottomrule
\end{tabular}%
}
\end{table*}

%% file: Figures/code/zero_shot_time.tex
\begin{figure*}[ht]
    \centering
    \includegraphics[width=0.8\textwidth]{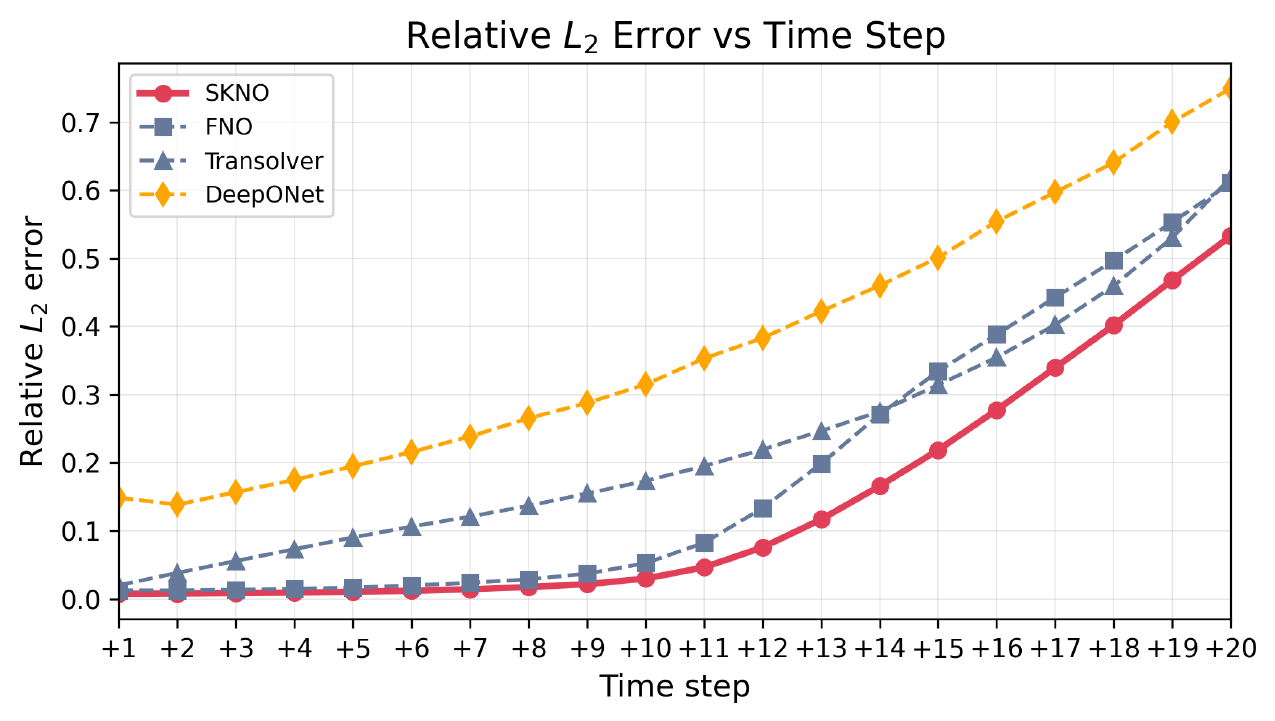}
    \caption{Visualization of zero-shot predictions at future time steps. For the 2D incompressible Navier–Stokes with viscosity $1\times 10^{-4}$ \cite{21:fno}, all models are autoregressively evaluated over the twenty time steps following the input time. SKNO maintains its better performance on both the seen temporal range ($+1$ to $+10$) and the unseen temporal range ($+11$ to $+20$), achieving a substantial performance margin over competing methods. Compared with FNO, Transolver, and DeepONet, SKNO achieves average error reductions of 34.5$\%$, 58.7$\%$, and 74.5$\%$, respectively, across all shown predictions. Note that even on the first predicted step, SKNO attains an error of $0.0076$, outperforming FNO ($0.0127$), Transolver ($0.0202$), and DeepONet ($0.1488$).}
    \label{fig5:time_unseen}
\end{figure*}

%% file: Sections/5_Conclusion.tex
\newpage

\section{Conclusion, Limitations and Future work} \label{Section_5}

In this work, we reformulate neural operators within a \((d+1)\)-dimensional framework, in which latent functions live on \(D_x \times D_p\) and are evolved by operators acting jointly over the physical and auxiliary coordinates. Building on this formulation, we propose the Schrödingerised Kernel Neural Operator (SKNO) as one Fourier-based instantiation with residual linear blocks and basis-diversified auxiliary evolution along \(D_p\). Across more than ten benchmarks, SKNO attains the lowest relative \(L_2\) error among the compared baselines under the reported settings. Its resolution robustness, zero-shot temporal generalization, and controlled ablation studies support the empirical effectiveness of the proposed \((d+1)\)-dimensional design. Taken together, these results suggest that auxiliary-domain operator evolution is a promising alternative to brute-force embedding scaling.

\paragraph{Multiple auxiliary coordinates.} We restrict to one auxiliary
coordinate ($d+k=d+1$) to stay within the standard NO setting. Here, we discuss two directions for future neural operator works with $k\ge2$.
(i) \emph{Structural}: the auxiliary coordinates mirror physical ones and the recovery integral coincides with
moment-taking; $k$ then stays small (2--3), and coarse or low-rank
representations along $p$ keep the cost manageable.
(ii) \emph{Auxiliary}: a second lift $w(x,p,q)$ of the latent function of a
pretrained operator $\mathcal{G}_1$ could learn an operator-to-operator mapping
to another operator $\mathcal{G}_2$ under the same input, with $q$ the
coordinate that selects which solution map is realized. More broadly, $p$ can be read as a gauge-like internal
degree of freedom that makes internal states explicit.

Another important future direction is to develop quantitative criteria for comparing neural operators beyond final test error alone. Because training navigates a high-dimensional error landscape over a large set of kernel parameters, optimization trajectories and attained local minima can vary substantially across architectures. While existing models (e.g., \cite{21:fno, 23:GNOT, 24:Transolver}) and related theoretical analyses (e.g., \cite{21:universal_fno, 25:nonlocality, 25:quanti_NO_parabolic}) provide approximation-oriented results, they offer limited insight into which minima are selected by practical training. Future work will therefore study how different aggregation designs and propagation parameters affect neighborhood-averaged training and test errors around optimizer-attained solutions.

\newpage

\section{Acknowledgments}

This work was financially supported by the Advanced Materials-National Science and Technology Major Project (Grant No. 2025ZD0620100). Wei Wang was supported by HKUST(GZ)-IEIP-RoP (G01RF000256), the National Key R\&D Program of China (No. 2024YFA1012700), the Guangdong Provincial Key Laboratory of Integrated Communication, Sensing and Computation for Ubiquitous Internet of Things (No. 2023B1212010007), and the Key Research and Development Program in Sichuan Province of China (No. 2024YFFK0410). Zhilu Lai was supported by the Guangdong Provincial Fund-Special Innovation Project (2024KTSCX038) and the Research Grants Council of Hong Kong through the Research Impact Fund (R5006-23). Xiaobo Zhang was supported by the Key Research and Development Program in Sichuan Province of China (No. 2024YFFK0410). Zecheng Gan was supported by NSFC grants 12201146, the Guangdong Natural Science Foundation (Grant No. 2023A1515012197), and the Guangzhou-HKUST(GZ) Joint Research Project (Grants Nos. 2024A03J0606).

%% file: checklist.tex
\section*{NeurIPS Paper Checklist}

\begin{enumerate}

\item {\bf Claims}
    \item[] Question: Do the main claims made in the abstract and introduction accurately reflect the paper's contributions and scope?
    \item[] Answer: \answerYes{}
    \item[] Justification: The abstract and introduction state the proposed $(d+1)$-dimensional neural-operator framework, the SKNO instantiation, and the empirical scope. These claims are supported by the main results in Section 4 and the additional experiments in Appendix C.

\item {\bf Limitations}
    \item[] Question: Does the paper discuss the limitations of the work performed by the authors?
    \item[] Answer: \answerYes{}
    \item[] Justification: Limitations and future directions are discussed in Section 5, including the need for quantitative criteria beyond final test error and a better understanding of optimization behavior across neural-operator architectures.

\item {\bf Theory assumptions and proofs}
    \item[] Question: For each theoretical result, does the paper provide the full set of assumptions and a complete (and correct) proof?
    \item[] Answer: \answerYes{}
    \item[] Justification: The theoretical results, assumptions, and proofs are provided in Appendix E, including the auxiliary-dimension neural-operator result, sparsity/energy-capture analysis, and adjoint/backward-propagation properties, etc.

\item {\bf Experimental result reproducibility}
    \item[] Question: Does the paper fully disclose all the information needed to reproduce the main experimental results of the paper to the extent that it affects the main claims and/or conclusions of the paper (regardless of whether the code and data are provided or not)?
    \item[] Answer: \answerYes{}
    \item[] Justification: Section 4 and Appendix C describe the benchmarks, baselines, losses, data splits, model settings, and evaluation protocols. The supplementary materials include the code package, repository materials, license, and detailed configurations.

\item {\bf Open access to data and code}
    \item[] Question: Does the paper provide open access to the data and code, with sufficient instructions to faithfully reproduce the main experimental results, as described in supplemental material?
    \item[] Answer: \answerYes{}
    \item[] Justification: The supplementary materials include the code package, repository files, license, and detailed configuration files needed to reproduce the reported experiments. The paper also states that the implementation and related configurations are included in the supplementary materials.

\item {\bf Experimental setting/details}
    \item[] Question: Does the paper specify all the training and test details (e.g., data splits, hyperparameters, how they were chosen, type of optimizer) necessary to understand the results?
    \item[] Answer: \answerYes{}
    \item[] Justification: The main experimental setup is described in Section 4.1, and Appendix C provides benchmark-specific data splits, optimizer choices, learning rates, epochs, batch sizes, modes/slices, positional features, and other configurations.

\item {\bf Experiment statistical significance}
    \item[] Question: Does the paper report error bars suitably and correctly defined or other appropriate information about the statistical significance of the experiments?
    \item[] Answer: \answerYes{}
    \item[] Justification: The paper reports multi-seed results, standard deviations, min--max ranges, and mean errors for representative experiments and capacity studies, including the six-seed evaluations in the appendix. These results quantify variability across random seeds for key empirical claims.

\item {\bf Experiments compute resources}
    \item[] Question: For each experiment, does the paper provide sufficient information on the computer resources (type of compute workers, memory, time of execution) needed to reproduce the experiments?
    \item[] Answer: \answerYes{}
    \item[] Justification: The paper reports that experiments are run on a single NVIDIA GeForce RTX 4090 GPU and provides training times, parameter counts, FLOPs, and memory-related constraints in the main tables and Appendix C.

\item {\bf Code of ethics}
    \item[] Question: Does the research conducted in the paper conform, in every respect, with the NeurIPS Code of Ethics \url{https://neurips.cc/public/EthicsGuidelines}?
    \item[] Answer: \answerYes{}
    \item[] Justification: The work uses publicly available or synthetically generated scientific datasets and does not involve human subjects, personal data, or sensitive information. The experiments and release materials are intended for reproducible scientific research.

\item {\bf Broader impacts}
    \item[] Question: Does the paper discuss both potential positive societal impacts and negative societal impacts of the work performed?
    \item[] Answer: \answerYes{}
    \item[] Justification: The work may positively impact scientific machine learning by improving efficient PDE/operator learning for physical simulation. Possible negative impacts are limited but include over-reliance on learned surrogates in safety-critical scientific or engineering settings without proper validation.

\item {\bf Safeguards}
    \item[] Question: Does the paper describe safeguards that have been put in place for responsible release of data or models that have a high risk for misuse (e.g., pre-trained language models, image generators, or scraped datasets)?
    \item[] Answer: \answerNA{}
    \item[] Justification: The paper does not release high-risk generative models, scraped datasets, or models with obvious dual-use misuse risks. The released materials are code and configurations for scientific PDE/operator-learning experiments.

\item {\bf Licenses for existing assets}
    \item[] Question: Are the creators or original owners of assets (e.g., code, data, models), used in the paper, properly credited and are the license and terms of use explicitly mentioned and properly respected?
    \item[] Answer: \answerYes{}
    \item[] Justification: Existing datasets, baselines, and implementations are cited in the paper, and the supplementary repository package includes license information. The experiments use publicly available or synthetically generated datasets with credited sources.

\item {\bf New assets}
    \item[] Question: Are new assets introduced in the paper well documented and is the documentation provided alongside the assets?
    \item[] Answer: \answerYes{}
    \item[] Justification: The new code package and repository materials are documented in the supplementary materials with detailed configurations and license information. These materials are provided to support reproduction of the proposed SKNO experiments.

\item {\bf Crowdsourcing and research with human subjects}
    \item[] Question: For crowdsourcing experiments and research with human subjects, does the paper include the full text of instructions given to participants and screenshots, if applicable, as well as details about compensation (if any)? 
    \item[] Answer: \answerNA{}
    \item[] Justification: The paper does not involve crowdsourcing, user studies, or research with human subjects. No participant instructions, screenshots, or compensation details are applicable.

\item {\bf Institutional review board (IRB) approvals or equivalent for research with human subjects}
    \item[] Question: Does the paper describe potential risks incurred by study participants, whether such risks were disclosed to the subjects, and whether Institutional Review Board (IRB) approvals (or an equivalent approval/review based on the requirements of your country or institution) were obtained?
    \item[] Answer: \answerNA{}
    \item[] Justification: The paper does not involve human subjects, personal data, or crowdsourcing. Therefore, IRB approval or equivalent human-subject review is not applicable.

\item {\bf Declaration of LLM usage}
    \item[] Question: Does the paper describe the usage of LLMs if it is an important, original, or non-standard component of the core methods in this research? Note that if the LLM is used only for writing, editing, or formatting purposes and does \emph{not} impact the core methodology, scientific rigor, or originality of the research, declaration is not required.
    \item[] Answer: \answerNA{}
    \item[] Justification: LLMs only contribute to the writing on this paper.

\end{enumerate}

%% file: Appendix/1_Claim_of_LLM.tex
\section{Statement on LLM Usage}

In this paper, we used a large language model (LLM) to provide suggestions on word choice, grammar, and logical consistency. While some sentences were lightly rephrased by the LLM, all scientific content remains original and authored by us.

\section{Ethics and Reproducibility Statements} \label{Appendix_supp_material}

This work does not involve human subjects, personal data, or other sensitive information. All experiments were conducted on publicly available or synthetically generated datasets. To ensure reproducibility, we provide detailed descriptions of the experimental setup, including model architectures, training procedures, and evaluation details. The code and datasets used in this study have been made publicly accessible. For readers, the code implementations and related configurations are included in the supplementary materials.

%% file: Appendix/2_Experiment_Details.tex
\section{Model, Experiment Settings and Results}\label{Appendix_Experiment_Settings}

\subsection{Model} \label{subsec_Model_plot}

\paragraph{Averaging Neural Operator on $D_p$} \label{ANO} In experiments governed by non-linear equations, we utilize ${F}_p^{-1}\tilde{A}{F}_p + B$, where $\tilde{A}$ and $B$ are both matrices with diagonally enhanced initialization. Such kind of implementation can be thought as an Averaging Neural Operator (ANO) \cite{25:nonlocality} along $p$, which governs the universal approximation of the PDE operator on $D_p$.

Below, we clarify the discrete instantiation of the linear blocks, including how the auxiliary operator nests within the kernel integral operator.

\input{Figures/code/added_1}

\subsection{Experiment Settings and Results}

\paragraph{Baselines} \label{appendix:baseline} The principle of our experiment configurations for different baselines is to keep the best configuration provided in their papers or codes, while ensuring fairness and not exploding the \emph{Nvidia GeForce RTX 4090} GPU memory (limited to 24564MiB). Detailed configurations of all baselines in different experiments are shown in Table \ref{tab:conf_supple}. And some explaination notes are listed below:

\begin{itemize}[topsep=0pt, leftmargin=1.5em]

    \item Mainly two types of Positional Features (Pos. Feat.), absolute coordinates (Abs.) and reference grid (Ref.), are concatenated into input signals:
    \begin{itemize}
        \item FNO \cite{21:fno} concatenates the input function value $a \in \mathbb{R}^{d_a}$ with the normalized absolute position coordinate $x \in \mathbb{R}^d$, usually scaled between 0 and 1.
        \item Transolver \cite{24:Transolver}, on the other hand, uses an advanced version which constructs a coarse reference grid and computes Euclidean distances from each input point to each point in this reference grid, concatenating these relative positions $x_{ref} \in \mathbb{R}^{(d_{ref})^d}$.
        \item In our work, we utilize the same absolute positional features as FNO to ensure a fair comparison.
    \end{itemize}

    \item Since Transolver employs the multi-head in its implementation, it creates sub-grid with a equal size from the complete one.

    \item For Transolver, the optimizers and scheduler type is AdamW \cite{19:AdamW} and OneCycleLR, while for FNO and SKNO are Adam \cite{14:adam} and Cosine Annealing. All the learning rates are initially set to $0.001$. More configurations can be found in the code provided in our supplementary materials.

\end{itemize}

\paragraph{Loss} Denote the parameterized neural operator as $\mathcal{G}_{\theta}$, we aim to approximate the solution operator $\mathcal{G}$ in Eq.~\ref{Equ_10} by optimizing the model parameters $\theta \in \Theta$ through relative $L_2$ loss below:
\begin{align} \label{loss_l_2}
    \min_{\theta \in \Theta} \mathcal{L}(\theta) &:= \min_{\theta \in \Theta} \frac{1}{S} \sum_{s=1}^{S} \left[ \frac{\lVert {\mathcal{G}}_{\theta}[a_s] - u_s\rVert_2}{\lVert u_s \rVert_2}  \right ] \\
    & = \min_{\theta \in \Theta} \frac{1}{S} \sum_{s=1}^{S} \left[ \left ( \frac{\sum_{j}({\mathcal{G}}_{\theta}[a_s](x_j) - u_s(x_j))^2}{\sum_{j} (u_s(x_j))^2}  \right )^{\frac{1}{2}} \right ],
\end{align}
where $\Theta$, $S$, and $u_s=u_s(x)$ denote the parameter space, the number of function samples, and the $i$-th output function, respectively.

\paragraph{Benchmarks and Results} \label{appendix:bmandresults} Benchmarks and the corresponding experimental results are summarized below:

\begin{enumerate}[leftmargin=*]

\item \textit{1D Heat Equation} \label{bm:1dheat}

Consider the 1D Heat Equation as follows:
\begin{equation}
    \partial_tu = c\partial_{xx}u,
\end{equation}
where $c=0.05$ as in \cite{22:PDNO}. We generate 1200 samples (1000 for the training set, and 100 for testing) in a denser grid with size $2^{10} = 1024$, and down-sample them to size $2^8 = 256$. Our goal is to predict solution $u_1 = u(x, t=1)$ from $u_0 = u(x, t=0)$. Results are shown in Table~\ref{tab:Linear_blocks}.

\item \textit{1D Advection Equation} \label{bm:1dadvection}

Consider 1D Advection Equation:
\begin{equation}
    \partial_tu = \beta\partial_xu,
\end{equation}
where $\beta$ is set with $1$. We use the data from \cite{22:PDEBench} with 2000 samples of $u_0 = u(x, t=0)$. Our goal is to predict the next 10 time steps from the last 10 of the solution u(x, t). The number of training and testing samples is 1000 and 200, respectively. Results are shown in Table~\ref{tab:Linear_blocks}

\item \textit{1D Burgers Equation} \label{bm:1dburgers}

The 1D Burger’s Equation on the unit torus is formed as  \cite{21:fno}:
\begin{align}
    \partial_t u + u\partial_xu &= \nu \partial_{xx}u,
\end{align} where $\nu=0.1$ with periodic boundary conditions. Our goal is to build a solution operator to predict the solution $u(x)$ at $t=1$. The training and testing grid size is set with $2^7=128$. Results are listed in Table~\ref{tab:main_result}.

\item \textit{2D Gray-Scott System} \label{bm:2dgs}

Consider a pair of coupled reaction-diffusion PDEs as follows:
\begin{align}
    \partial_tu &= D_u\Delta u - uv^2 + F(1-u), \nonumber \\
    \partial_tv &= D_v\Delta v - uv^2 - (F+k)v,
\end{align}
where $D_u = 0.2097, D_v = 0.105$ denote the diffusion coefficients respectively for $u, v$. And $F = 0.03, k = 0.062$ are the reaction parameters. We follow the data generation in \cite{24:boosting} to make a training set with 200 samples of $u, v$ evolution and a testing set with 20 in a $32 \times 32$ size grid with periodic boundary conditions. Our goal is to predict the next 10 time steps from the last 10 of $u, v$. See Table~\ref{tab:main_result} for results.

\item \textit{2D Shallow-Water Equations} \label{bm:2dshallow}

The 2D shallow-water equations govern the evolution of fluid flows under the assumption that horizontal length scales are much larger than the vertical depth. They can be expressed as the following system of hyperbolic PDEs:

\begin{align}
     \partial_t h + \nabla h {\bf u} &= 0, \\
     \partial_t h {\bf u} + \nabla \left( {\bf u}^2 h + \frac{1}{2} g_r h^2 \right) &= - g_r h \nabla b \, , 
\end{align}

where $h = h(x,t)$ denotes the fluid depth, $\mathbf{u} = \mathbf{u}(x,t)$ is the 2D velocity field, $g_r$ is the (reduced) gravitational acceleration, $b = b(x)$ represents the bathymetry characterizing the bottom topography. For this benchmark, we employ the dataset from \cite{22:PDEBench}, which is generated under Neumann boundary conditions. To assess resolution-invariant stability during training, models are trained on 500 samples with resolutions randomly drawn from $\{32, 64, 128\}$ and evaluated on 300 test samples at the highest resolution 128. The prediction task is to forecast the water depth over the next 10 time steps, given the previous 10. See Table~\ref{tab:other_exp_bench} for results.

\item \textit{2D Darcy Flow} \label{bm:2ddarcy}

The 2D Darcy flow models steady flow through a heterogeneous porous material,
such as a rock formation or a two-phase composite, and is governed by the
second-order elliptic equation
\begin{align}
    -\nabla \cdot(a\nabla u) &= f,
\end{align}
where $a(x)$ is the permeability field of the material, $u(x)$ the pressure,
and $f=1$ a uniform source. The same equation describes steady heat conduction
in a composite with conductivity $a(x)$, so the task is a microstructure-to-response
map. Following \cite{21:fno}, $a(x)$ is
piecewise constant, so the material has sharp phase interfaces. Our goal is to
approximate the mapping operator from $a(x)$ to $u(x)$ with zero Dirichlet
boundary conditions \cite{21:fno}. The results are based on models trained on 1000 training samples, which are down-sampled to a size of ($85 \times 85$), and tested on 100 samples. Results are listed in Table~\ref{tab12:darcy_NO_family}.

\item \textit{2D Incompressible Navier-Stokes Equations} \label{bm:2dNS}

The 2D incompressible Navier-Stokes (NS) Equations in vorticity form on a unit torus are shown below:
\begin{align}
    \partial_t \omega + {u} \cdot \nabla \omega - \nu \Delta \omega &= f, \\
    \nabla \cdot {u} = 0.
\end{align}
Here $\nu=1e{-5}$ and $f = 0.1(sin(2 \pi (x_1 + x_2)) +
cos(2 \pi (x_1 + x_2)))$. Our goal is to predict the next 10 time steps from the last 10 ($d_{{a}}=d_{{u}}=10)$ of the solution $w({x}, t)$ with periodic boundary conditions on the 2D grid with size \(64 \times 64\) \cite{21:fno}. All models are trained on 1000 training samples and tested on 200 test samples. Results are listed in Table~\ref{tab:main_result} and Table~\ref{tab13:ns_stat}.

\item \textit{3D Compressible Navier–Stokes Equations} \label{bm:3dns}

The 3D compressible NS equations describe fluid dynamics with density changes, given by:
\begin{align}
 \partial_t \rho + \nabla \cdot (\rho \textbf{v}) &= 0, \label{eq:cnast-1}
 \\
 \rho (\partial_t \textbf{v} + \textbf{v} \cdot \nabla \textbf{v}) &= - \nabla p + \eta \triangle \textbf{v} + (\zeta + \eta/3) \nabla (\nabla \cdot \textbf{v}),
 \label{eq:cnast-2}\\
 \partial_t \left[ \epsilon + \frac{\rho v^2}{2} \right] &+ \nabla \cdot \left[ \left(\epsilon + p + \frac{\rho v^2}{2} \right) \bf{v} - \bf{v} \cdot \sigma' \right] = 0,\label{eq:cnast-3}
\end{align}
where $\textbf{v} = \textbf{v}(x, t)$, $\rho = \rho(x, t)$, and $p = p(x, t)$ denote the velocity vector field, fluid density (mass per unit volume), and pressure, respectively. We consider the challenging transonic flow benchmark from \cite{22:PDEBench}, with shear and bulk viscosity coefficients $\eta = \zeta = 1 \times 10^{-8}$, and internal energy density $\epsilon = \tfrac{3p}{2}$. The viscous stress tensor $\sigma' = \sigma'(\textbf{v}; \eta, \zeta)$ is determined by the velocity field under the Newtonian fluid assumption. Our objective is to predict the velocity, density, and pressure fields three time steps ahead, given the previous three steps ($d_{a} = d_{u} = 3 \times 5 = 15$). The simulations are performed on a downsampled 3D grid of size $32 \times 32 \times 32$ with periodic boundary conditions. The initial conditions consist of turbulent velocity fields with uniform mass density and pressure. All benchmark models are trained on 500 samples and evaluated on 100 test samples. See Table~\ref{tab:main_result} for results.

\item \textit{3D Rayleigh-Taylor Instability} \label{bm:3dRayTaylorInstab}

The Rayleigh–Taylor (RT) instability occurs when a heavier fluid overlies a lighter one under the influence of gravity, which is extremely sensitive to perturbations of the interface and governed by the following PDEs:

\begin{align}
    \partial_t\rho + \nabla\cdot(\rho \boldsymbol{u}) &= 0,\label{continuity}\\
    \partial_t(\rho \boldsymbol{u})+\nabla\cdot(\rho \boldsymbol{u} \boldsymbol{u}) &= -\nabla p + \nabla\cdot \boldsymbol{\tau} +\rho \boldsymbol{g}, \label{momentumEqn}\\
     \nabla\cdot\boldsymbol{u} &= -\kappa\nabla\cdot\left(\frac{\nabla\rho}{\rho}\right). \label{incompressible}
\end{align}

where $\rho = \rho(x, t)$ and $\boldsymbol{u} = \boldsymbol{u}(x,t)$ denote the density and the velocity vector field, respectively. $p$ is the pressure, $\boldsymbol{\tau}=\boldsymbol{\tau}(\boldsymbol{u}, \rho; \nu)$ is the eviatoric stress tensor with kinematic viscosity $\nu$, $\boldsymbol{g}$ represents the gravity, and $\kappa$ is the coefficient of molecular diffusivity. A key dimensionless parameter in the RT simulations is the Atwood number $A = (\rho_h - \rho_l) / (\rho_h + \rho_l)$, which quantifies the density contrast between the two fluids. For example, miscible fluids with greater density contrasts exhibit stronger instabilities. In this work, we adopt the dataset with $A = 0.25$ provided in \cite{24:thewell} to train a solution operator $vec(\rho, \boldsymbol{u}; t = t_0) \mapsto vec(\rho, \boldsymbol{u}; t = t_0 + 20)$ which evolves the system from rest states (with small perturbations) to mixed states after 20 time steps. The models are trained on 220 mapping pairs and tested on 40 pairs from distinct trajectories, using a downsampled grid of size $32 \times 32 \times 32$. Horizontal periodic boundary conditions and impermeable vertical walls are imposed. See Table~\ref{tab:main_result} for results.

\item \textit{ERA5 Wind Field Prediction} \label{bm:era5}

To compare the baseline models in a global context of spatio-temporal dynamics, we evaluate their performance on predicting the next-hour ERA5 global wind field \cite{20:era5}. All models are trained using data from the previous four days to predict the next-hour wind field at a grid size of $512 \times 512$, and are tested on data from a new day with the same grid size. See Table~\ref{tab:era5} for results.

\end{enumerate}

We emphasize the broader learning difficulty spectrum of the benchmarks we selected. As shown in Figure~\ref{fig9:time_unseen} which illustrates the spectral decay of truncation modes, our chosen benchmarks \ref{bm:2ddarcy} and \ref{bm:2dNS} contain significantly more high-frequency modes than simpler benchmarks such as \ref{bm:1dburgers} or other incompressible NS benchmarks with $\nu > 1 \times 10^{-4}$. To further evaluate the capability of these efficient and lightweight neural operators in capturing complex non-linear dynamics in 3D space, we include two additional, extremely challenging benchmarks: \ref{bm:3dns} and \ref{bm:3dRayTaylorInstab}. Moreover, we also compare the performance of all baseline models in the rapid prediction of global wind fields, providing a comprehensive evaluation across both highly nonlinear and large-scale spatio-temporal dynamics.

\paragraph{Zero-shot Super-resolution} To assess the resolution-invariant inference capability of SKNO, we evaluate the model on three benchmarks: 1D Burgers’ equation (trained on a grid of size $128$), 2D Darcy Flow (trained on a grid of size $85 \times 85$), and ERA5 Global Wind Field Prediction (trained on a grid of size $32 \times 32$). As shown in Table~~\ref{tab14:superes_table} and Table~~\ref{tab:superes_table_era5}, SKNO maintains stability and accuracy even when tested on extremely fine grids, including $8192$, $(421)^2$, and $(512)^2$. All experiments are conducted on a single RTX 4090 GPU, consistent with the setup used in other sections.

\begin{table}[ht]
\centering
\caption{Complete recommended training configurations}
\label{tab:conf_supple}
\resizebox{\linewidth}{!}{
\begin{tabular}{l|cccccc}
\hline
Model & Epochs & Batch Size & \# Modes \textit{OR} Slices & \# Layers & \# Pos. Feat. / Type & \# Sub-grid $\times$ size on $D_p$\\
\hline
\multicolumn{7}{c}{1D Heat (Linear Block)} \\ 
\hline
Transolver & 500 & 20 & 4 & 1 & N.A. & 1 $\times$ 4 \\
FNO & 500 & 20 & 1 & 1 & N.A. & 1 $\times$ 4 \\
\textbf{SKNO} & 500 & 20 & 1 & 1 & N.A. & 1 $\times$ 4 \\ 
\hline
\multicolumn{7}{c}{1D Advection (Linear Block)} \\ 
\hline
Transolver & 500 & 4 & 8 & 1 & N.A. & 1 $\times$ 16 \\
FNO & 500 & 20 & 8 & 1 & N.A. & 1 $\times$ 16 \\
\textbf{SKNO} & 500 & 20 & 8 & 1 & N.A. & 1 $\times$ 16 \\ 
\hline
\multicolumn{7}{c}{1D Burgers} \\ 
\hline
Transolver & 500 & 4 & 64 & 8 & 64 / Ref. & 8 $\times$ 32 \\
FNO & 500 & 20 & 16 & 4 & 1 / Abs. & 1 $\times$ 64 \\
\textbf{SKNO} & 500 & 20 & 16 & 4 & 1 / Abs. & 1 $\times$ 64 \\ 
\hline
\multicolumn{7}{c}{2D Gray-Scott} \\ 
\hline
Transolver & 500 & 2 & 32 & 8 & $8^2 = 64$ / Ref. & 8 $\times$ 32 \\
FNO & 500 & 10 & 12 & 4 & 2 / Abs. & 1 $\times$ 24 \\
\textbf{SKNO} & 500 & 10 & 12 & 4 & 2 / Abs. & 1 $\times$ 24 \\ 
\hline
\multicolumn{7}{c}{2D Shallow-Water} \\ 
\hline
Transolver & 100 & 2 & 32 & 8 & $8^2 = 64$ / Ref. & 8 $\times$ 32 \\
FNO & 100 & 20 & 8 & 4 & 2 / Abs. & 1 $\times$ 24 \\
\textbf{SKNO} & 100 & 20 & 8 & 4 & 2 / Abs. & 1 $\times$ 24 \\ 
\hline
\multicolumn{7}{c}{2D Darcy Flow} \\ 
\hline
Transolver & 500 & 4 & 64 & 8 & $8^2 = 64$ / Ref. & 8 $\times$ 16 \\
FNO & 500 & 20 & 12 & 4 & 2 / Abs. & 1 $\times$ 32 \\
\textbf{SKNO} & 500 & 20 & 12 & 4+1 & 2 / Abs. & 1 $\times$ 32 \\ 
\hline
\multicolumn{7}{c}{2D Incompressible Navier-Stokes} \\ 
\hline
Transolver & 500 & 2 & 32 & 8 & $8^2 = 64$ / Ref. & 8 $\times$ 32 \\
FNO & 8000 & 20 & 12 & 4 & 2 / Abs. & 1 $\times$ 24 \\
\textbf{SKNO} & 3000 & 20 & 12 & 4+1 & 2 / Abs. &  1 $\times$ 24 \\
\hline
\multicolumn{7}{c}{3D Compressible Navier-Stokes} \\ 
\hline
Transolver & 100 & 2 & 32 & 8 & $8^3 = 512$ / Ref. & 8 $\times$ 32 \\
FNO & 100 & 10 & 16 & 4 & 3  / Abs. & 1 $\times$ 32 \\
\textbf{SKNO} & 100 & 10 & 16 & 4 & 3 / Abs. &  1 $\times$ 32 \\
\hline
\multicolumn{7}{c}{3D Rayleigh-Taylor Instability} \\ 
\hline
Transolver & 250 & 2 & 32 & 8 & $8^3 = 512$ / Ref. & 8 $\times$ 32 \\
FNO & 250 & 10 & 16 & 4 & 3  / Abs. & 1 $\times$ 32 \\
\textbf{SKNO} & 250 & 10 & 16 & 4+1 & 3 / Abs. &  1 $\times$ 32 \\
\hline
\multicolumn{7}{c}{ERA5 Wind Field Prediction} \\ 
\hline
Transolver & 50 & 2 & 64 & 4 & $8^2 = 64$ / Ref. & 8 $\times$ 8 \\
FNO & 50 & 2 & 10 & 4 & 2 / Abs. & 1 $\times$ 20 \\
\textbf{SKNO} & 50 & 2 & 10 & 4 & 2 / Abs. &  1 $\times$ 20 \\
\hline
\end{tabular}}
\end{table}

%% file: Figures/code/added_1.tex
\begin{figure*}[ht]
    \centering
    \includegraphics[width=\textwidth]{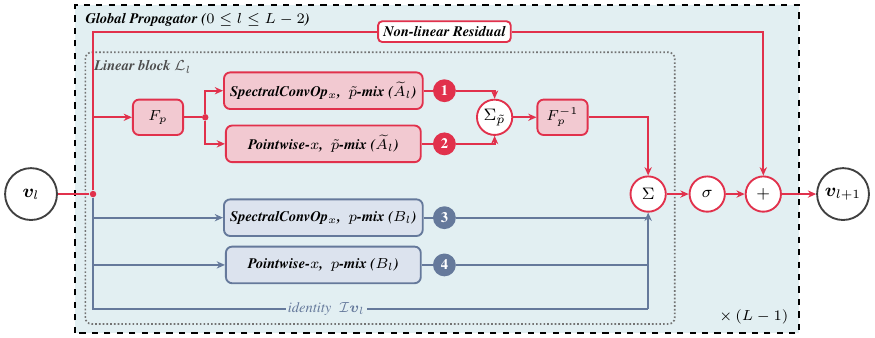}
    \includegraphics[width=0.7\textwidth]{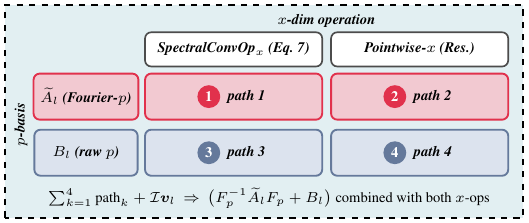}
    \caption{Forward pass of one global propagator ($0 \le l \le L-2$). The latent tensor
    $\bm{v}_l \in \mathbb{R}^{N_p \times N_x}$ is processed by paths~1--2 in Fourier-$p$
    coordinates (red) and by paths~3--4 in raw $p$ coordinates (blue), alongside the identity. All
    paths are summed within the linear block $\mathcal{L}_l$ before the activation $\sigma$ and
    the (non-linear) residual.}
    \label{fig:skno_flow}
\end{figure*}

%% file: Appendix/5_Related_Work.tex
\newpage

\section{Related Work} \label{Appendix_related_work}

\paragraph{Adding one dimension to solve PDEs} Unlike traditional methods \cite{04:rom,11:pgd} that rely on dimension reduction techniques, such as symmetry analysis \cite{02:symmetry_analy}, to represent equations within their original domain, approaches like the level set method \cite{01:level_set, 04:levelset, 95:fast_level_set, 03:level_riemann} and the recent Schrödingerisation method \cite{24:analog_JIN, 23:schroedingerisation_JIN, 24:schroedingerisation_JIN_PRL} propose extending original PDEs to one more higher dimensional domains to create a pipe for functions to evolve. Similar "adding one dimension" ideas have also been used in signal processing like wavelets \cite{92:wavelets} which employ scale and shift dimensions to represent original signals, and in quantum mechanics like the Wigner functions \cite{32:wigner} which utilize a single function displaying the probability (quasi-)distribution in the phase space.

In this paper, Schrödingerisation is used only as design inspiration for introducing an auxiliary function coordinate and Fourier-domain evolution along that coordinate. We do not claim that Schrödingerisation itself provides a classical numerical acceleration mechanism for the nonlinear well-posed PDE benchmarks considered here. The theoretical and empirical claims concern the resulting NO architecture.

\paragraph{Neural Networks for PDEs} In scenarios where PDEs are unknown or training data is limited, supervised learning methods are employed to approximate solution operators, which recently shows their capacity to universally represent solution operators for partial differential equations (PDEs) \cite{89:universal_nn,21:universal_fno, 21:deeponet_new, 98:ANN_for_ODEPDE}. Existing approaches mainly fall into two paradigms: Common Neural Networks and Neural Operators.

Common NNs approximate solution operators working on a fixed grid size, which maps discretized input to output signals using architectures such as MLPs \cite{16:resnet, 18:neuralode, 25:yi_PINN}, CNNs \cite{22:cnn_solver, 23:CNN_PDE_solver}. For example, \cite{20:cfd_net, 21:parametric_PDE_ANN} utilizes a typical CNN structure to map the input to the output image, which is composed of the solution of numerical solvers.

Neural operators (NOs) are designed to map between function spaces, enabling generalization across varying input discretizations without retraining \cite{23:function_space_no, 23:nature}. One of the most popular NO definitions is the kernel integral operator \cite{21:fno, 23:convno, 24:laplace}. Early kernel function implementations employed graph-based message-passing networks \cite{20:GNO, 20:MGNO}, where dense edge connections encoded these kernels. However, these methods faced scalability challenges due to long training times and inefficiencies in modeling interactions across multiple spatial and temporal scales. Transformer-based models \cite{22:transformer_2}, similar to graphs, also encounter limitations in computational efficiency due to quadratic time complexity. To address these challenges, tricky computation strategies on kernels have been proposed. Galerkin-based methods \cite{21:Choose, 04:galerkin_fem} employ linear projections to parameterize kernels, while \cite{24:ONO} introduces a decomposition constrained by positive semi-definite assumptions. However, such methods suffer from the problems of reduced expressiveness due to the absence of non-linear activations like softmax. Recent SOTA model transolver \cite{24:Transolver} proposes a physics-attention module to approximate the kernel function with the aid of a self-attention-based aggregation on flattened data. While these graph or transformer-based advanced methods try to escape from the heavy quadratic complexity by tricky implementations on the kernel functions originally defined in \cite{20:GNO, 21:fno}, they still pay the similar computational price for their neglect of the embedding designs. While Mamba-based methods such as \cite{25:LaMO} can further reduce the computational cost over the domain after aggregation, the similar embedding scaling needs remain unaddressed. Instead, empirically designed fully connected or multi-head architectures are typically retained by default to model the evolution along the auxiliary dimension. 

Additionally, Fourier Neural Operators (FNO) \cite{21:fno} efficiently implement spectral kernel representations by global convolutions via the Fast Fourier Transform (FFT). While training fast, it also faces the representation bottleneck like other NO methods because of the unaligned evolution in embedding spaces. Subsequent works \cite{22:afno, 22:ufno, 23:FFNO, 24:TFNO, 25:bregmanNO} take efforts to extend the Fourier Transform to other analytically well-explored transformations. For example, Complex Neural Operator \cite{23:CoNO} generalizes the Fourier Transformation with an extra learnable rotation parameter to model the $d$-dimensional kernel integration.

There is another NO definition: DeepONet \cite{21:deeponet_new} employs a branch-trunk architecture inspired by the universal operator approximation theorem \cite{95:universal_no}. These representations are combined through a dot product for the embeddings composed with functional outputs to predict the solution values. Recent advancements, such as integrating numerical solvers with DeepONet \cite{24:blending}, have improved convergence by capturing multi-frequency components, enhancing performance on complex PDE systems. However, its general-purpose design may limit its effectiveness in scenarios requiring high accuracy and PDE-aligned structural representations.

\paragraph{Embedding augmentation and auxiliary coordinates.}
Augmenting a latent representation is a common design principle in continuous-depth models and operator learning. For example, augmented neural ODEs \cite{19:ANODE} enlarge an embedding vector so that continuous flows can represent maps that may be difficult to realize in the original size. Similarly, most neural operators lift pointwise input function values \(a(x)\) into an enlarged embedding before evolving them to improve approximation. In our formulation, the auxiliary variable \(p\) is treated as a function coordinate where the latent scalar function \(v(x,p)\) on \(D_x\times D_p\) evolves, and the update is parameterized by operators acting over this product domain. This makes the auxiliary direction available for kernel design.

\paragraph{Demonstrations of the Schrödingerisation}
We demonstrate how the $d$-dimensional linear PDEs are transformed into their $d+1$-dimensional versions in Schrödingerisation after introducing an auxiliary dimension $p$:

\begin{example} [Heat Equation \cite{22:schrodingerisation_JIN_tech, 23:schroedingerisation_JIN}] \label{example_1} Consider the initial-value problem of the heat equation: \begin{align}
        &\partial_t u = \partial_{xx} u, \qquad \text{($d$ dimensional heat equation)} \label{Equ_1} \\
    &u(x, t=0) = u_0, \qquad \text{($d$ dimensional input)} \label{Equ_2}
\end{align} where $u = u(x, t), \ x = (x_1, x_2, ..., x_d) \in \mathbb{R}^d, \ t \geq 0$. Let $d=1$, which can be easily generalized to arbitrary $d$.  Our goal is to construct a $d+1$ dimensional pipe that evolves the initial condition $u_0 = u(x, t = 0)$ to the solution $u_1 = u(x, t = 1)$.

Step 1: Warped Phase Transformation. Define a transformation from $u(x, t)$ to $v(x, p, t)$ by introducing an auxiliary dimension $p \in (-\infty, \infty)$: \begin{equation} 
    v(x, p, t=0) = v_0(x, p) = e^{-|p|} u_0.  \label{Equ_3}
\end{equation} Step 2: $d+1$ Dimensional Phase-space Heat Equation. Combine Eq.\ref{Equ_1}, \ref{Equ_2} and \ref{Equ_3}, obtain: \begin{equation}
    \partial_t v(x, p, t) = - \partial_p \partial_{xx} v(x, p, t). \label{Equ_4}
\end{equation} Step 3: Recovering to $d$ Dimensional Solution. After evolving $v_0 \mapsto v_1:=v(x, p, t=1)$ by Eq.\ref{Equ_4}, Recover $u_1$ by integrating along $p$: \begin{equation}
    u_1(x) = \int^{-\infty}_{\infty} \chi(p) v_1(x, p)dp,  \label{Equ_5}
\end{equation} where $\chi(p)$ has multiple choices and depends on the introduced transformation form in Step 1. Here $\chi(p)$ can be formed as the step function or delta function.

\end{example}

For Eq.\ref{Equ_3}, if taking fourier transform on $p$ ($\mathcal{F}_p: p\rightarrow \eta$), one gets a group of uncoupled Schrödinger equations (operators), over all $\eta$; If taking fourier transform on $x$ ($\mathcal{F}_x: x \rightarrow \xi$), one gets a group of uncoupled convection equations (operators), over all $\xi$. Generally for $\boldsymbol{v}(t)$, discrete signals on $x$ and $p$ from $v(x, p, t)$, one gets a Hamiltonian system by taking the discrete fourier transform (DFT) on $p$. For Eq.\ref{Equ_3}, the Hamiltonian system can be represented by further taking DFT on $x$: \begin{equation}
    \frac{d}{dt} \hat{\tilde{\boldsymbol{v}}}(t) = i(D_\mu^2 \otimes D_{\mu}) \hat{\tilde{\boldsymbol{v}}}(t), \label{Equ_6}
\end{equation} where $\hat{\tilde{\boldsymbol{v}}}(t)$ is derived after taking DFT for $\boldsymbol{v}(t)$ on $p$ and $x$, and the same notation $D_{\mu} = diag(\mu_{-N_p/2}, ..., \mu_{N_p/2})$, which denotes a diagonal matrix with $N_p$ grid points in one dimension, is employed since little confusion will arise. Here, the diagonal coefficient complex matrix $i(D_\mu^2 \otimes D_{\mu})$ is from taking DFTs on the $x$ and $p$ domain for $- \partial_p \partial_{xx}$ in Eq.\ref{Equ_3}.

\begin{example} [Advection Equation and Other Linear PDEs] \label{example_2} Similar process and result on advection equation as Example \ref{example_1} with introducing corresponding warped phase transformation $v_0 = sin(p)u_0, \ p \in [-\pi, \pi]$. For more discussion of other PDE types, refer to \cite{23:schroedingerisation_JIN}.

\end{example}

%% file: Appendix/4_Proof.tex
\section{Properties of SKNO}

We highlight several properties of SKNO below.

\paragraph{Complexity}  
To analyze computational complexity across general dimensional settings (1D, 2D, 3D, ...), let $N_x$ denote the number of grid points in the spatial domain, $N_p$ the grid size along the auxiliary dimension, and $k_{x}$ the number of aggregated slices/nodes (used in Transolver). The complexity comparison is summarized in Table~\ref{tab:complexity}.

\input{Tables/complexity}

Consider an extreme but practically relevant case in high-accuracy numerical simulations: a 3D field with grid resolution $((2)^9)^3 = (512)^3$ and $N_p = 32$ (while Transolver typically requires $N_p = 128$ for stable performance). Even with the minimal $k_x=32$, Transolver’s kernel implementation with a large linear constant ($k_x + N_p > 128$) leads to a computational cost much higher than SKNO ($\log N_x = 27$) or FNO ($\log N_xN_p =27 + 5 = 32$).

In lower dimensions like 1D or 2D cases, where $N_x$ is exponentially smaller, our kernel retains the linear-complexity advantages while achieving superior performance with significantly smaller scale constants compared to the SOTA Transolver (under the log benefit from Fast Fourier Transform). Here, we report the FLOPs of the kernel integration on representative experiments in Table~~\ref{tab:flops_rate}. Notably, when the positional encodings are removed in Table \ref{tab:complexity_lifting}, the prediction error of Transolver increases dramatically (from $6e-3$ to $6e-1$ in relative $L_2$ error on Burgers), whereas the other two models maintain nearly the same accuracy level.

\input{Tables/flop_memory}

\begin{figure}[h]\centering
\includegraphics[width=0.8\linewidth]{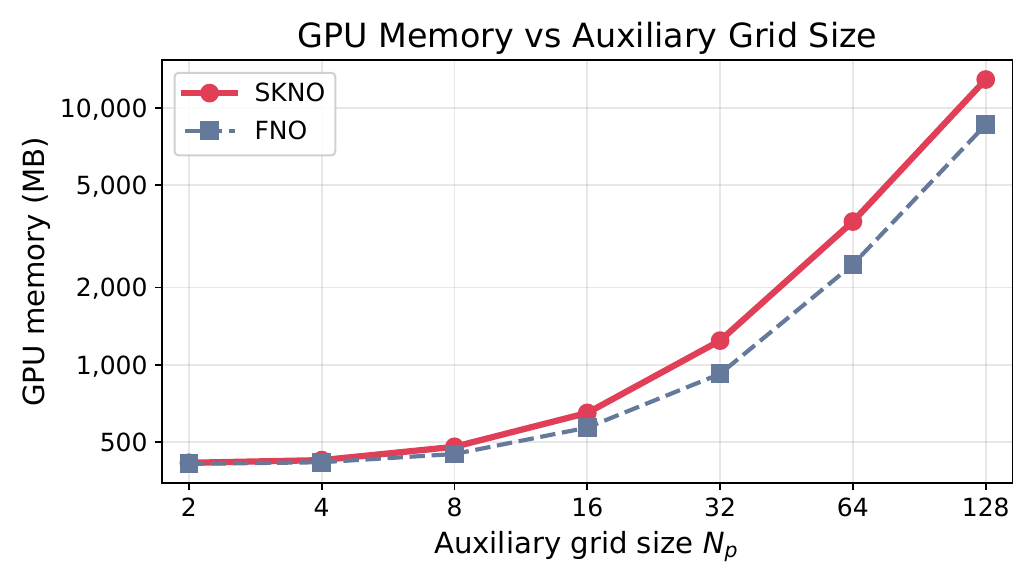}
\caption{GPU memory of the model on 3D compressible NS versus the auxiliary grid size
$N_p$.}
\label{fig:np_scan}
\end{figure}

\paragraph{Spectral Operator along $p$} Since our SKNO adopts a spectral implementation, we primarily compare it to the FNO \cite{21:fno} as a reference framework. This choice also facilitates generalization to other spectral models like \cite{22:afno,24:GINO}. To avoid notational clutter, we begin with the $d=1$ case, noting that the following arguments extend straightforwardly to higher dimensions.

\begin{theorem} [Neural Operator along the Auxiliary Dimension] \label{theorem_1}

Let \(v(x,p)\) be taken from a compact set \(V\) in the appropriate Sobolev space so that its truncated Fourier representatives
\[
\hat{\boldsymbol v}_i := F^x_{k_{x}}[P^x_{k_{x}} v](\hat{\boldsymbol x}_i,\boldsymbol p)\in \hat{V}_i\subset\mathbb C^{N_p},
\qquad
\tilde{\hat{\boldsymbol v}}_i:=F^p[R^p_{k_{p}}(F^x_{k_{x}}[P^x_{k_{x}} v]](\hat{\boldsymbol x}_i,\tilde{\boldsymbol p}))\in\tilde{\hat{V}}_i\subset\mathbb C^{N_p},
\]
are compact sets in a finite coefficient space. Then
\begin{enumerate}
  \item Approximating an operator \(\mathcal M\) is equivalent (via the Fourier-conjugate decomposition) to approximating the conjugate mapping(s) \(\{\widehat{\mathcal M}_i\}_{i=1}^{k_{x}}\) (and, after also Fourier-transforming in \(p\), the conjugates \(\{\widetilde{\widehat{\mathcal M}}_i\}\)).
  \item The forward DFT/frequency-projection and the inverse discrete transform admit neural operator approximation on the spectral domain (See Lemma 7 and Lemma 8 in \cite{21:universal_fno}, hence the operator learning problem reduces to a finite-dimensional neural-network approximation on the compact coefficient sets \(\hat{V}_i,\ \tilde{\hat{V}}_i\).
\end{enumerate}
In particular, treating the auxiliary \(p\)-dimension via an additional DFT (as SKNO does) simply produces an enlarged conjugate mapping; the same approximation strategy (as Lemma 7 $+$ Lemma 8 $+$ finite NN on compact coefficients) applies.
\end{theorem}

\textit{Note}. $\widehat{\mathcal{M}}_{i}: \hat{V}_{i} \subset \mathbb{R}^{2N_p} \rightarrow \mathbb{R}^{2N_p}$ for $i = 1, ..., k_{x}$ and $\widetilde{\widehat{\mathcal{M}}}_{i}: \tilde{\hat{V}}_{i} \subset \mathbb{R}^{2N_p} \rightarrow \mathbb{R}^{2N_p}$. Usually, we take $k_{p} = N_p$ and the $\frac{length(\boldsymbol{p}) - k_{p}}{length(\boldsymbol{p})}$ random dropout $R^p_{k_{p}} = I^p$.

\textit{Proof}. By the Fourier-conjugate decomposition (Eq.\,(12) in \cite{21:universal_fno}), the operator can be written as
\begin{align}
    \mathcal M &=\; (F^{x}_{k_{x}})^{-1} \circ \widehat{\mathcal M}_{k_{x}} \circ F^x_{k_{x}} \circ P^x_{k_{x}} \nonumber \\
    &=\; (F^{x}_{k_{x}})^{-1} \circ (F^p)^{-1} \circ \widetilde{\widehat{\mathcal M}}_{k_{x}} \circ F^p \circ F^x_{k_{x}} \circ P^x_{k_{x}}. \label{Equ_44}
\end{align}

Here, the second line follows from additionally applying a DFT in the auxiliary variable \(p\).

The mapping in Eq.\ref{Equ_44} \(F^pF^x_{k_{x}}[P^x_{k_{x}}(\cdot)]\) sends the compact input family \(V\) into a compact subset of Fourier coefficient space. Lemma~7 and Lemma~8 of \cite{21:universal_fno} guarantee that the left-hand and right-hand parts of both $\widehat{\mathcal M}_{k_{x}}$ and $\widetilde{\widehat{\mathcal M}}_{k_{x}}$ can be approximated within the desired accuracy.

Thus, the problem reduces to approximating the conjugate operators \(\widehat{\mathcal M}_{k_{x}}\) or \(\widetilde{\widehat{\mathcal M}}_{k_{x}}\), both of which are continuous maps defined on compact subsets of \(\mathbb R^{2N_p}\). By the universal approximation theorem for neural networks \cite{02:universal_nn_barron,89:universal_nn}, any such continuous map can be approximated to arbitrary accuracy by a finite neural network. Hence, there exists finite NNs that approximate the conjugate mappings on \(\hat{V}_i\) (or \(\tilde{\hat{V}}_i\)) within the desired accuracy. From the Theorem 5 in \cite{21:universal_fno}, SKNO universally approximates any $d$-dimensional operator $\mathcal{G}$.

Next, we try to explain why the auxiliary update used by SKNO can be more parameter-efficient than duplicating a standard same-basis channel-mixing branch.

\begin{proposition} [Complementary-basis enlargement under a finite auxiliary-operator budget]

Fix one spatial Fourier mode and write the discretized auxiliary profile as
$z\in\mathbb{C}^{N_p}$. Let $\Phi\in\mathbb{C}^{N_p\times N_p}$ be an invertible
basis transform along the auxiliary coordinate $p$; in SKNO, $\Phi=F_p$ is the DFT.
Let $\mathcal{S}\subseteq\mathbb{C}^{N_p\times N_p}$ denote the linear subspace
realizable by one finite-budget auxiliary branch, e.g., a diagonal, block, truncated,
Hermitian-constrained, or otherwise structured parameterization. Define the same-basis
two-branch class and the complementary-basis class by
\begin{align}
\mathcal{H}_{B+B}
&:= \{S_1+S_2:\; S_1,S_2\in\mathcal{S}\},\\
\mathcal{H}_{A+B}
&:= \{S+\Phi^{-1}T\Phi:\; S,T\in\mathcal{S}\}.
\end{align}
If $\mathcal{S}$ is a linear subspace, then
\begin{equation}
\mathcal{H}_{B+B}=\mathcal{S}.
\end{equation}
In contrast,
\begin{equation}
\dim(\mathcal{H}_{A+B})
=
\dim(\mathcal{S})
+
\dim(\Phi^{-1}\mathcal{S}\Phi)
-
\dim(\mathcal{S}\cap \Phi^{-1}\mathcal{S}\Phi).
\end{equation}
Therefore, whenever the transformed subspace is not contained in the original one,
i.e., $\Phi^{-1}\mathcal{S}\Phi\not\subseteq\mathcal{S}$, we have
\begin{equation}
\mathcal{H}_{B+B}\subsetneq \mathcal{H}_{A+B}.
\end{equation}

\end{proposition}

\textit{Proof.}
Since $\mathcal{S}$ is closed under addition, $S_1+S_2\in\mathcal{S}$ for all
$S_1,S_2\in\mathcal{S}$, and every element of $\mathcal{S}$ is obtained by choosing
$S_2=0$. Hence $\mathcal{H}_{B+B}=\mathcal{S}$. The expression for
$\dim(\mathcal{H}_{A+B})$ is the standard dimension formula for the sum of two
linear subspaces, applied to $\mathcal{S}$ and $\Phi^{-1}\mathcal{S}\Phi$.
If $\Phi^{-1}\mathcal{S}\Phi\not\subseteq\mathcal{S}$, then there exists
$T\in\mathcal{S}$ such that $\Phi^{-1}T\Phi\notin\mathcal{S}$; this element belongs
to $\mathcal{H}_{A+B}$ by choosing $S=0$, but it does not belong to
$\mathcal{H}_{B+B}=\mathcal{S}$. Thus the inclusion is strict. \qed

For example, let
$\mathcal{S}$ be the diagonal matrices, i.e., operators acting pointwise in $p$.
Then $\Phi^{-1}\mathcal{S}\Phi$ is the set of circulant matrices (convolutions
along $p$), and $\mathcal{S}\cap\Phi^{-1}\mathcal{S}\Phi=\mathrm{span}\{I\}$, so
$\dim\mathcal{H}_{A+B}=2N_p-1$ while $\dim\mathcal{H}_{B+B}=N_p$. This is an
operator-level form of the discrete Fourier uncertainty
principle~\cite{89:Uncertain_prin}: no non-trivial operator is local in both
$p$ and its Fourier conjugate, so a branch in one basis cannot reach what is
compact in the other. Table~\ref{tab11:capacity_efficiency} confirms this prediction on trained networks:
replacing the Fourier-$p$ branch by a same-basis duplicate, SKNO$(B{+}B)$,
doubles the error on 2D Gray--Scott
($2.447\mathrm{e}{-2}$ vs.\ $1.256\mathrm{e}{-2}$, 6 seeds), even though
SKNO$(A{+}B)$ uses fewer parameters and FLOPs.

\begin{corollary} [Finite-budget approximation error]

For any target auxiliary linear update \(L^\star\),
\[
\inf_{H\in\mathcal{H}_{A+B}}\|L^\star-H\|_F
\le
\inf_{H\in\mathcal{H}_{B+B}}\|L^\star-H\|_F .
\]
\begin{equation}
\inf_{H\in\mathcal{H}_{A+B}}\|L^\star-H\|_F
\le
\inf_{H\in\mathcal{H}_{B+B}}\|L^\star-H\|_F,
\end{equation}
with strict improvement whenever $L^\star$ has a component in
$\Phi^{-1}\mathcal{S}\Phi$ outside $\mathcal{S}$. Thus, adding a same-basis branch
can be redundant, whereas adding a conjugated branch can enlarge the reachable
finite-budget operator family. This explains the role of SKNO's
$F_p^{-1}\tilde A F_p+B$ update: the raw-$p$ and Fourier-$p$ terms provide two
coordinate views of the same auxiliary profile.

\end{corollary}

\begin{remark}[Unitary stability]
If \(\Phi\) is unitary, as for the DFT, then
\[
\|\Phi^{-1}T\Phi\|_2=\|T\|_2,\qquad
\|\Phi^{-1}T\Phi\|_F=\|T\|_F .
\]
Thus the complementary branch changes represented directions without increasing operator norms.
\end{remark}

\input{Tables/control_scale}

\input{Figures/code/sparse_dictionary}

\paragraph{Sparsity.}

Due to the limited network width, $N_p$ should be finite and as small as possible. Benefiting from the PDE-aligned propagator design, SKNO constructs a sparser dictionary on the output side of the $d+1$ dimensional propagation pipes that is more efficient in capturing the energy of the solutions than FNO.

For a clear and fair comparison, we consider SKNO without the differential operator and the baseline FNO. The lifting and recovering operators in both models are chosen as unbiased linear layers without positional encoding. Under the same configurations, we compare the dictionaries of SKNO and FNO at the last output linear layer in Figure~\ref{fig:sparse_dict}, which illustrates the sparsity advantage of the SKNO dictionary.

Beyond the experiments, we also provide a formal proof in a finite-dimensional matrix–vector setting, which aligns well with the neural implementation, to clarify the efficient energy capture of SKNO. The extension to function spaces amounts to working in the corresponding infinite-dimensional setting.

\begin{lemma}[Preserved Energy after Top-$r$ Truncation] \label{energy_preserve_trun_svd}
For a general matrix–vector multiplication $\boldsymbol{u}=G\boldsymbol{y}$ where $G \in \mathbb{R}^{N_x \times N_p}$ and $N_x \geq N_p$, if we perform the top-$r$ approximation $G_r$ using the singular value decomposition (SVD) of $G$, the energy preserved in the prediction $G_r\boldsymbol{y}$ is
\begin{equation}
    \lVert G_r\boldsymbol{y} \rVert_2^2 = \sum_{m=1}^r \lambda_m^2 \langle \boldsymbol{y}, \phi_m\rangle^2, \quad r \leq N_x,
\end{equation}
where $G_r\boldsymbol{y} = \Psi_r \Lambda_r \Phi_r^T\boldsymbol{y}=\sum_{m=1}^r \lambda_m \langle \boldsymbol{y}, \phi_m\rangle \psi_m$, and $\psi_m$, $\phi_m$, and $\lambda_m$ are the $m$-th left singular vector, right singular vector, and singular value, respectively.
\end{lemma}

\textit{Proof.} This follows directly from $G_r\boldsymbol{y}$ representing with the orthonormal basis $\{\psi_m\}$ of $\mathrm{range}(r)$.

\begin{proposition}[Conditional energy-capture dominance under limited width] \label{remark_3}
Assume the two models use the same lifting and recovery operators and that, under the considered
width budget, the realizable last-layer dictionary class of FNO is contained in that of SKNO,
\[
\mathcal H_{\mathrm{FNO}}(r)\subseteq \mathcal H_{\mathrm{SKNO}}(r).
\]
Define the best achievable top-\(r\) captured energy
\[
E^\star_{\mathrm M}(r)
:=
\sup_{V_L^{(\mathrm M)}\in\mathcal H_{\mathrm M}(r)}
\frac{\|(V_L^{(\mathrm M)})_r\boldsymbol{\chi}^{(\mathrm M)}\|_2^2}{\|\boldsymbol{u}\|_2^2}.
\]
Then
\[
E^{(\mathrm{SKNO})}(r)\ge E^{(\mathrm{FNO})}(r),
\]
and therefore the corresponding best rank-\(r\) truncation error of SKNO is no larger than that of FNO.
\end{proposition}

\textit{Note.} (1) The length $N_x$ in the following discussion is obtained by flattening the $d$-dimensional tensor while keeping the sequential order. (2) Without loss of generality, we discuss the case $d_u = 1$.

\textit{Proof.} Denote the vector $\boldsymbol{u} \in \mathbb{R}^{N_x}$ discretized on $D_x$, the vector $\boldsymbol{\chi}^{(\mathrm{M})} \in \mathbb{R}^{N_p}$ discretized on $N_p$, the matrix $V_L^{(\mathrm{M})} \in \mathbb{R}^{N_x \times N_p}$ where $N_x \geq N_p$. We focus on the last output (linear) layer in neural operators $\mathrm{M}$. If $N_p$ is sufficiently large, we have

\begin{align}
    \boldsymbol{u} &= V_L^{(\mathrm{M})}\boldsymbol{\chi}^{(\mathrm{M})} \nonumber \\
    &= \Psi_L^{(\mathrm{M})} \Lambda_L^{(\mathrm{M})} (\Phi_L^{(\mathrm{M})})^T \boldsymbol{\chi}^{(\mathrm{M})},
\end{align}

with singular values $\lambda_1^{(\mathrm{M})} \geq \lambda_2^{(\mathrm{M})} \geq \dots \geq \lambda_{N_x}^{(\mathrm{M})}$. The matrices $\Psi_L^{(\mathrm{M})} = (\psi_1^{(\mathrm{M})}, \psi_2^{(\mathrm{M})}, \dots, \psi_{N_x}^{(\mathrm{M})})$ and $\Phi_L^{(\mathrm{M})} = (\phi_1^{(\mathrm{M})}, \phi_2^{(\mathrm{M})}, \dots, \phi_{N_p}^{(\mathrm{M})})$ are composed with their column vectors $\{\psi_m\}_{m=1}^{N_x}$ and $\{\phi_m\}_{m=1}^{N_p}$, respectively.

In practice, we apply the $r$-truncation on $N_p$ as the optimal approximation under the metric in Eq. \ref{loss_l_2}. Utilizing the rank-$r$ truncated SVD, we have

\begin{align} \label{rank_N_p_SVD}
    (V_L^{(\mathrm{M})})_r &:= (\Psi_L^{(\mathrm{M})})_r (\Lambda_L^{(\mathrm{M})})_r (\Phi_L^{(\mathrm{M})})_r^T \nonumber \\
    &= \sum_{m=1}^{r} \lambda_m^{(\mathrm{M})} \psi_m^{(\mathrm{M})} (\phi_m^{(\mathrm{M})})^T, \ r \le N_x.
\end{align}

From Lemma \ref{energy_preserve_trun_svd}, the (relative) energy of $u$ captured by the top-$r$ SVD modes of model $\mathrm{M}$ is

\begin{align}   \label{Equ_rel_energy}
    E^{(\mathrm{M})}(r) &:= \frac{\rVert  (V_L^{(\mathrm{M})})_r \boldsymbol{\chi}^{(\mathrm{M})}\lVert_2^2}{\rVert  \boldsymbol{u}\lVert_2^2} \nonumber \\
    &= \frac{\sum_{m=1}^{r} (\lambda_m^{(\mathrm{M})})^2 \langle \boldsymbol{\chi}^{(\mathrm{M})}, \phi_m^{(\mathrm{M})} \rangle^2}{\sum_{m=1}^{N_x} (\lambda_m^{(\mathrm{M})})^2 \langle \boldsymbol{\chi}^{(\mathrm{M})}, \phi_m^{(\mathrm{M})} \rangle^2}.
\end{align}

Since singular values are non-negative, the sequence $E^{(\mathrm{M})}(r)$ is non-decreasing with respect to $r$ for each model $\mathrm{M}$. And the truncation error is

\begin{equation}
    \lVert \boldsymbol{u} - (V_L^{(\mathrm{M})})_r \boldsymbol{\chi}^{(\mathrm{M})} \rVert_2 = \rVert  \boldsymbol{u}\lVert_2 \sqrt{1-E^{(\mathrm{M})}(r)}, \quad r = 1, \dots, N_x.
\end{equation}

Specifically, the error difference between FNO and SKNO is

\begin{align}
    \epsilon(r) &= \frac{\lVert \boldsymbol{u} - (V_L^{(\mathrm{FNO})})_r \boldsymbol{\chi} \rVert_2 - \lVert \boldsymbol{u} - (V_L^{(\mathrm{SKNO})})_r \boldsymbol{\chi} \rVert_2}{\rVert  \boldsymbol{u}\lVert_2} \nonumber \\
    &:= \sqrt{1-E^{(\mathrm{FNO})}(r)} - \sqrt{1-E^{(\mathrm{SKNO})}(r)}.
\end{align}

From Eq.\ref{Equ_rel_energy}, $\epsilon(r) \rightarrow 0$ as $r \rightarrow N_p$.

Let $\mathcal{H}^{(\mathrm{M})}$ denotes the set of $V_L^{(\mathrm{M})}$ that can be realized by NO $\mathrm{M}$ under the same width budget $r$. We have

\begin{equation}
    E^{(\mathrm{M})}(r)
    := \sup_{V_L^{(\mathrm{M})} \in \mathcal{H}_{\mathrm{M}}}
    \frac{\big\| (V_L^{(\mathrm{M})})_r \boldsymbol{\chi}^{(\mathrm{M})}\big\|_2^2}{\|\boldsymbol{u}\|_2^2},
    \qquad \mathrm{M} \in \{\mathrm{FNO},\mathrm{SKNO}\},\ r \le N_p.
\end{equation}

Since $\mathcal{H}_{\mathrm{FNO}} \subset \mathcal{H}_{\mathrm{SKNO}}$, the supremum over $\mathcal{H}_{\mathrm{SKNO}}$ cannot be smaller
than the supremum over $\mathcal{H}_{\mathrm{FNO}}$, which implies

\begin{equation}
    E^{(\mathrm{SKNO})}(r) \geq E^{(\mathrm{FNO})}(r), \ \forall \ r \le N_p.
\end{equation}

\paragraph{Adjoint Backward Propagation}

Under the squared loss function like Eq. \ref{loss_l_2}, we show the adjoint backward propagation property of SKNO.

\begin{proposition} [Discrete backward updates for SKNO auxiliary branches] \label{Backward_Update} Here we show the backward updates around linear kernel integrations in $\mathrm{SKNO}$. Assume $d_a = d_u = 1$ and denote the matrix multiplication on the last output layer $V_L^{(\mathrm{SKNO})}\boldsymbol{\chi}^{(\mathrm{SKNO})}$ and the backward input residual $\boldsymbol{r}^{(\mathrm{SKNO})}$, where the vector $\boldsymbol{\chi}^{(\mathrm{SKNO})} \in \mathbb{R}^{N_p}$ is discretized on $N_p$, and the matrix $V_L^{(\mathrm{SKNO})} \in \mathbb{R}^{N_x \times N_p}$ where $N_x > N_p$. The backpropagation \cite{86:backprop} update

\begin{align}
    \Delta\boldsymbol{\chi}^{(\mathrm{SKNO})} &= \sum_i(V_L^{(\mathrm{SKNO})})_{i\cdot}^T \cdot \boldsymbol{r}_i^{(\mathrm{SKNO})} \\
    \mathrm{Grad}(V_L^{(\mathrm{SKNO})}) &= {\boldsymbol{r}}^{(\mathrm{SKNO})} \cdot (\boldsymbol{\chi}^{(\mathrm{SKNO})})^T \\ \label{Gradient_V_L}
    FC:(\Delta \widehat{\mathbf{B}})_{ijj'} &= \mathrm{conj}(\widehat{V}_{L-1})_{ij} \cdot \hat{\boldsymbol{r}}_i^{(\mathrm{SKNO})} \cdot \mathbf{M}_{ij'}^{(\mathrm{SKNO})} \cdot (\boldsymbol{\chi}^{(\mathrm{SKNO})})^T_{j'} \\ \label{Delta_tensor_B}
    \mathrm{Grad}(\widehat{V}_{L-1}^{(\mathrm{SKNO})})_{ij} &= \sum_{j'} \widehat{\mathbf{B}}_{ijj'} \cdot \hat{\boldsymbol{r}}_i^{(\mathrm{SKNO})} \cdot \mathbf{M}_{ij'}^{(\mathrm{SKNO})} \cdot (\boldsymbol{\chi}^{(\mathrm{SKNO})})^T_{j'}\\ \label{Gradient_VL_-1}
    (\Delta \widehat{\widetilde{\mathbf{A}}})_{ijj'} &= \mathrm{conj}(\widehat{\widetilde{V}}_{L-1})_{ij} \cdot \hat{\boldsymbol{r}}_i^{(\mathrm{SKNO})} \cdot \mathbf{M}_{ij'}^{(\mathrm{SKNO})} \cdot (\tilde{\boldsymbol{\chi}}^{(\mathrm{SKNO})})^T_{j'} \\
    \mathrm{Grad}(\widehat{\widetilde{V}}_{L-1}^{(\mathrm{SKNO})})_{ij} &= \sum_{j'} \widehat{\widetilde{\mathbf{A}}}_{ijj'} \cdot \hat{\boldsymbol{r}}_i^{(\mathrm{SKNO})} \cdot \mathbf{M}_{ij'}^{(\mathrm{SKNO})} \cdot (\tilde{\boldsymbol{\chi}}^{(\mathrm{SKNO})})^T_{j'}\\ \label{Gradient_VL_-1_}
    Diag:(\Delta \widehat{{\mathbf{B}}})_{ij} &= \mathrm{conj}(\widehat{{V}}_{L-1})_{ij} \cdot \hat{\boldsymbol{r}}_i^{(\mathrm{SKNO})} \cdot \mathbf{M}_{ij}^{(\mathrm{SKNO})} \cdot ({\boldsymbol{\chi}}^{(\mathrm{SKNO})})^T_j \\
    \mathrm{Grad}(\widehat{V}_{L-1}^{(\mathrm{SKNO})})_{ij} &= \widehat{\mathbf{B}}_{ij} \cdot \hat{\boldsymbol{r}}_i^{(\mathrm{SKNO})} \cdot \mathbf{M}_{ij}^{(\mathrm{SKNO})} \cdot (\boldsymbol{\chi}^{(\mathrm{SKNO})})^T_{j}\\ \label{Gradient_VL_-1_dia}
    (\Delta \widehat{\widetilde{\mathbf{A}}})_{ij} &= \mathrm{conj}(\widehat{\widetilde{V}}_{L-1})_{ij} \cdot \hat{\boldsymbol{r}}_i^{(\mathrm{SKNO})} \cdot \mathbf{M}_{ij}^{(\mathrm{SKNO})} \cdot (\tilde{\boldsymbol{\chi}}^{(\mathrm{SKNO})})^T_j \\
    \mathrm{Grad}((\widehat{\widetilde{V}}_{L-1}^{(\mathrm{SKNO})})_{ij} &= \widehat{\widetilde{\mathbf{A}}}_{ij} \cdot \hat{\boldsymbol{r}}_i^{(\mathrm{SKNO})} \cdot \mathbf{M}_{ij}^{(\mathrm{SKNO})} \cdot (\tilde{\boldsymbol{\chi}}^{(\mathrm{SKNO})})^T_{j}, \label{Gradient_VL_-1_dia_}
\end{align}

where the weight $\Delta\boldsymbol{\chi} \in \mathbb{R}^{N_p}$, complex tensors $\Delta \widehat{\mathbf{B}} \in \mathbb{C}^{N_x \times N_p \times N_p} \ or \ \mathbb{C}^{N_x \times N_p}$ and $\Delta \widehat{\widetilde{\mathbf{A}}} \in \mathbb{C}^{N_x \times N_p \times N_p} \ or \ \mathbb{C}^{N_x \times N_p}$ are the changing values of parameters. $\mathrm{conj}(\cdot)$ denotes element-wise taking the conjugate value of the input. $\mathbf{M}_{ij}^{(\mathrm{SKNO})} = \mathbf{1}[(V_L'^{(\mathrm{SKNO})})_{ij}>0]$ where $V_L^{(\mathrm{SKNO})} = \mathrm{ReLU}(V_L'^{(\mathrm{SKNO})})$.
    
\end{proposition}

\textit{Proof.} Derive these updating values by taking partial derivatives with respect to the parameters in propagators $ \widehat{\mathbf{B}}$ and $ \widehat{\widetilde{\mathbf{A}}}$ and in the recovering operator $\boldsymbol{\chi}$. A notable point here is the conjugate gradient backpropagation path to the Fourier domain like the $\hat{\boldsymbol{r}}_i^{(\mathrm{SKNO})}$ like in Eq.\ref{Delta_tensor_B}.

The adjoint backward residual propagation and conjugate parameter updates shown above are directly derived from SKNO's spectral implementation. From Proposition \ref{Backward_Update}, we can generalize the backward propagation framework for $d+1$-dimensional neural operators as follows

\begin{align}
    \Delta v_L(x, p) &= \mathcal{Q}^{*}[r](x, p) = \chi^T(p)r(x) \\
    \mathcal{K}^{*}[\Delta v](x,p) &= \mathcal{K}[\Delta v](x,p) = \iint_{D_{{x}} \times {D_p}} \kappa({x}, {y}, p, p') \Delta v({y},p')d{y}dp' \\
    \Delta a(x) &= \mathcal{P}^{*}[\Delta v_0](x) = \int_{D_p}\omega(p)\Delta v_0(x,p)dp,
\end{align}

where $r(x) := \Delta u(x) = u(x) - \mathcal{G}[a](x)$ denotes the residual between the solution $u(x)$ and the corresponding forward prediction $\mathcal{G}[a](x)$, and we define the adjoint operator of the lifting $\mathcal{P}^{*}$ to have a $p$-integral form as the recovering operator $\mathcal{Q}$, and vice versa. See Lemma \ref{adjoint_proof}.

For previous $d$ dimensional kernel integration designs, an additional matrix inversion is required when formulating the backpropagation process, whereas $d+1$ dimensional neural operators are allowed to share the same kernel for the forward and adjoint operators, i.e., $\mathcal{K} = \mathcal{K}^{*}$. When $d+1$ dimensional neural operators converge, the backpropagation updates \cite{86:backprop} stabilize within a PDE-specific equivalence class of parameters, up to a certain error tolerance.

\begin{lemma}[Adjoint of the recovery operator \(\mathcal{Q}\)] \label{adjoint_proof}
Let $H_x := L^2(D_x)$ and $H_{x,p} := L^2(D_x\times D_p)$ with the standard $L^2$ inner products. Then $\mathcal{Q}^*$ is the adjoint of $\mathcal{Q}$, i.e.
\[
\langle \mathcal{Q}[v], r \rangle_{H_x}
=
\langle v, \mathcal{Q}^* [r] \rangle_{H_{x,p}}
\quad \forall v\in H_{x,p},\ r\in H_x.
\]
\end{lemma}

\textit{Proof.} By definition of the $L^2$ inner product on $H_x$,
\begin{align}
\langle \mathcal{Q}[v], r \rangle_{H_x}
& = \int_{D_x} (\mathcal{Q}[v])(x)\,r(x)\,dx \nonumber \\
& = \int_{D_x} \left( \int_{D_p} \chi(p)\,v(x,p)\,dp \right) r(x)\,dx \nonumber \\
& = \int_{D_x}\int_{D_p} v(x,p)\,\chi(p)\,r(x)\,dp\,dx.
\end{align}

By definition of the $L^2$ inner product on $H_{x,p}$,
\begin{align}
    \langle v, \mathcal{Q}^* [r] \rangle_{H_{x,p}}
& = \int_{D_x}\int_{D_p} v(x,p)\,(\mathcal{Q}^* [r])(x,p)\,dp\,dx \nonumber \\
& = \int_{D_x}\int_{D_p} v(x,p)\,\chi(p)\,r(x)\,dp\,dx.
\end{align}

Thus $\langle \mathcal{Q}[v], r \rangle_{H_x}
= \langle v, \mathcal{Q}^* [r] \rangle_{H_{x,p}}$ for all $v,r$, so $\mathcal{Q}^*$ is the adjoint of $\mathcal{Q}$. Similar proof for the adjoint between $\mathcal{P}$ and $\mathcal{P}^{*}$.

\paragraph{Entanglement}

Moreover, we compute the entanglement entropy values after each $d+1$ dimensional propagation, and plot some randomly selected cases of them in Figure~\ref{fig:entanglement_entropy}: Compared to FNO, SKNO introduces stronger perturbations on the singular spectrum after $\mathcal{L}_l[v_l]$, cooperated with effectively controlling the sparsity of the energy distribution by $\sigma_l$. In addition, the PDE-aligned design in the $\mathcal{L}_l[v_l]$ of SKNO makes it more flexible to reduce entanglement in the final linear stage further, while FNO is purely driven by the squared-loss objective to substantially perturb the $(d+1)$-dimensional signal without an implicit structural regularization along $p$, which leads to increased entanglement entropy in some cases.

\input{Figures/code/entanglement_entropy}

%% file: Tables/complexity.tex
\begin{table}[H]
\centering
\caption{Complexity comparison of different kernel implementations.}
\setlength{\tabcolsep}{8pt}
\renewcommand{\arraystretch}{1.3}
\begin{tabular}{l|c}
\hline
\textbf{Models} & \textbf{Complexity} \\
\hline
Transolver & $\mathcal{O}\!\left[N_p (N_x (k_x + N_p) +k_x^2)\right] \; (k_x \geq 32, \; N_p = 128 \text{ or } 256)$ \\
FNO        & $\mathcal{O}\!\left[N_p (N_x \log(N_x) + N_pk_x)\right] \; (k_x \leq 16, N_p \leq 32)$ \\
\textbf{SKNO (Ours)} & $\mathcal{O}\!\left[N_p (N_x (\log N_x + \log N_p) + N_p k_x) \right] \; (k_x \leq 16, N_p \leq 32)$ \\
\hline
\end{tabular}
\label{tab:complexity}
\end{table}

\begin{table}[H]
\centering
\caption{Complexity comparison of different lifting implementations with positional encoding. (Assume all choose one linear layer)}
\setlength{\tabcolsep}{8pt}
\renewcommand{\arraystretch}{1.3}
\begin{tabular}{l|c}
\hline
\textbf{Models} & \textbf{Complexity} \\
\hline
Transolver & $\mathcal{O}\!\left[N_p (d_a + (d_{ref})^d)\right] \; (d_{ref} = 8, \; N_p = 128 \text{ or } 256)$ \\
FNO        & $\mathcal{O}\!\left[N_p (d_a + d)\right] \; (N_p \leq 32)$ \\
\textbf{SKNO (Ours)} & $\mathcal{O}\!\left[N_p (d_a + d) \right] \; (N_p \leq 32)$ \\
\hline
\end{tabular}
\label{tab:complexity_lifting}
\end{table}


%% file: Tables/flop_memory.tex
\begin{table}[H]
\centering
\caption{Computational cost comparison (FLOPs) of the kernel integration and relative rate (normalized by FNO).}
\label{tab:flops_rate}
\setlength{\tabcolsep}{8pt}
\renewcommand{\arraystretch}{1.15}
\begin{tabular}{lccc}
\toprule
\diagbox{Model}{Metric} & {FNO} & {Transolver} & {SKNO} \\
\midrule
\multicolumn{4}{c}{\textit{2D Navier--Stokes}} \\
\midrule
FLOPs & $1.16\times 10^{8}$ & $1.19\times 10^{9}$ & $2.12\times 10^{8}$ \\
\midrule
Rate  & $1.000$            & $10.292$            & $1.831$ \\
\midrule
\multicolumn{4}{c}{\textit{3D Rayleigh--Taylor Instability}} \\
\midrule
FLOPs & $8.98\times 10^{8}$ & $9.00\times 10^{9}$ & $1.59\times 10^{9}$ \\
\midrule
Rate  & $1.000$             & $10.029$            & $1.774$ \\
\bottomrule
\end{tabular}
\end{table}

%% file: Tables/control_scale.tex
\begin{table*}[t]
\centering
\caption{Capacity Efficiency on 2D Gray-Scott. Comparison of SKNO against FNO under systematically varied embedding sizes ($1\times$ to $4\times$), parallel stacking (2/3/4-Stack), and branch architecture (B+B vs.\ A+B), evaluated over 6 random seeds with 500 training epochs. SKNO (A+B) achieves the lowest error while using fewer parameters and fewer FLOPs than FNO at any tested scaling configuration. FNO scaling is non-monotonic, peaking at $2\times$ embedding size and degrading beyond. SKNO(B+B), which replaces the Fourier-domain branch along $p$-dim with a duplicate standard branch at matched parameter budget, fails to improve over FNO, confirming the architectural origin of SKNO's gains.}
\label{tab11:capacity_efficiency}
\small
\begin{tabular*}{\textwidth}{@{\extracolsep{\fill}} l r r r }
\toprule
\textbf{Model} & \textbf{\# Params} & \textbf{Rel.\ $L_2$ Err.\ $\pm$ Std.} & \textbf{FLOPs} \\
\midrule
\multicolumn{4}{c}{\textit{FNO with varying embedding size}} \\
\midrule
FNO & 1,337,450 & 2.169e-2 $\pm$ 3.35e-3 & 940,318,720 \\
FNO ($+$1 emb.) & 1,451,177 & 2.042e-2 $\pm$ 1.24e-3 & 998,184,960 \\
FNO (1.33$\times$ emb.) & 2,377,186 & 1.869e-2 $\pm$ 2.21e-3 & 1,445,109,760 \\
FNO (1.5$\times$ emb.) & 3,008,414 & 1.834e-2 $\pm$ 2.67e-3 & 1,733,386,240 \\
FNO (2$\times$ emb.) & 5,347,538 & 1.685e-2 $\pm$ 3.73e-3 & 2,741,739,520 \\
FNO (3$\times$ emb.) & 12,030,266 & 1.952e-2 $\pm$ 1.83e-3 & 5,404,303,360 \\
FNO (4$\times$ emb.) & 21,385,634 & 1.839e-2 $\pm$ 1.10e-3 & 8,928,010,240 \\
\midrule
\multicolumn{4}{c}{\textit{Parallel stacking in the same basis}} \\
\midrule
2-Stack FNO & 2,685,962 & 2.532e-2 $\pm$ 4.29e-3 & 2,328,371,200 \\
3-Stack FNO & 4,027,370 & 2.214e-2 $\pm$ 2.26e-3 & 3,429,376,000 \\
4-Stack FNO & 5,368,778 & 2.063e-2 $\pm$ 3.30e-3 & 4,530,380,800 \\
\midrule
\multicolumn{4}{c}{\textit{SKNO branch ablation}} \\
\midrule
SKNO (B+B) & 2,690,762 & 2.447e-2 $\pm$ 9.65e-4 & 2,521,047,040 \\
\textbf{SKNO (A+B)} & \textbf{2,141,844} & \textbf{1.256e-2 $\pm$ 1.14e-3} & \textbf{2,112,141,236} \\
\bottomrule
\end{tabular*}
\end{table*}

%% file: Figures/code/sparse_dictionary.tex
\begin{figure*}[ht]
    \centering
    \includegraphics[width=\textwidth]{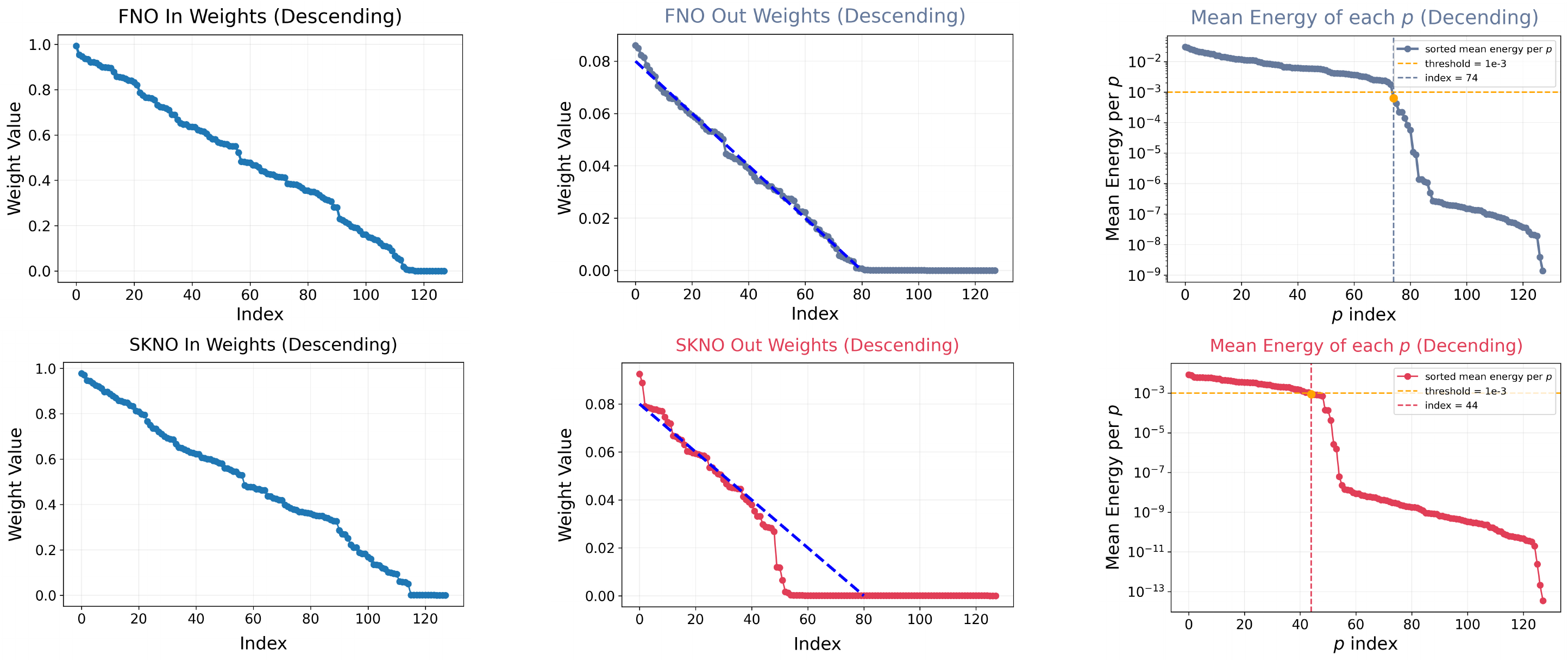}
    \caption{Dictionary comparison of FNO and SKNO. We train both models with a sufficiently large recovering budget $N_p$ on the Burgers benchmark and visualize the absolute values of the input/output weights, together with the mean energy of the dictionary of pattern fields. While the input weights of FNO (top left) and SKNO (bottom left) exhibit nearly identical distributions, SKNO yields a noticeably sparser structure in its output weights (bottom middle) and in the mean energy of each pattern field (bottom right) compared to FNO. The yellow horizontal line denotes a mean-energy threshold of $10^{-3}$, where modes below this threshold contribute only marginally to the final prediction. For a complete visualization of the output-weight and pattern-field pairs in the dictionary, see Figure~\ref{fig:pattern_scale_128_sorted_weight}.}
    \label{fig:sparse_dict}
\end{figure*}

%% file: Figures/code/entanglement_entropy.tex
\begin{figure*}
    \centering
    \includegraphics[width=\textwidth]{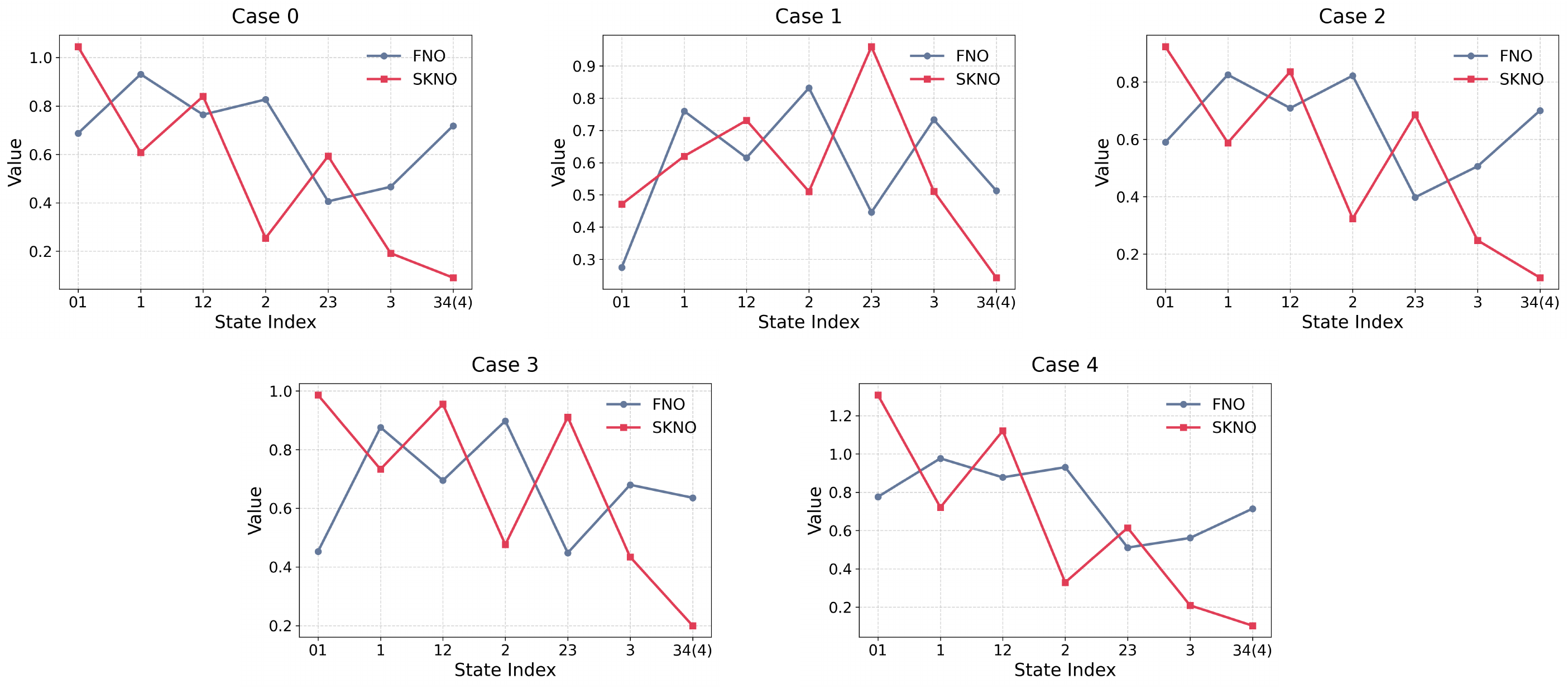}
    \caption{
    Entanglement entropy of intermediate layer outputs. For five randomly selected test cases from the test set, the final entanglement entropy of SKNO is consistently lower than that of FNO, indicating that the $(d+1)$-dimensional design of SKNO yields representations that are less entangled and exhibit a more energy-concentrated distribution. On the horizontal axis, a two-digit index (e.g., "$01$") denotes the output after the $\mathcal{L}_0$ and before the $\sigma_0$ between $v_0$ and $v_1$, whereas a single-digit index (e.g., "$1$") denotes $v_1$ as the output of the first propagator ($\sigma_0 \circ \mathcal{L}_0$). Note that the initial index “0” corresponds to $v_0$. Under a purely linear lifting operator without positional encoding, $v_0$ is typically separable, so all curves effectively start from an unentangled state at the origin and evolve through the forward propagation.
}
    \label{fig:entanglement_entropy}
\end{figure*}


%% file: Appendix/0_Table_of_Notations.tex
\section{Table of Notations}

A table of key notations is given in Table \ref{table:notations}.

%% file: Appendix/3_Visualization.tex
\section{Visualization} \label{Appendix:Visual}

\paragraph{Patterns with different scales} We visualize cases to highlight the patterns with different scales of our SKNO. These subplots are plotted over all $p$ values after the $d+1$ dimensional evolution. Cases include the 1D Burgers, 2D Darcy Flow, and 2D incompressible Navier-Stokes benchmarks, shown in Fig.~\ref{fig:pattern_scale_128_sorted_weight}, Fig.~\ref{fig10:over_p_darcy} and Fig.~\ref{fig11:over_p_ns} respectively.

\begin{algorithm} 

\caption{Entanglement Entropy Numerical Calculation}

\begin{algorithmic}[ht] \label{Entang_entro}

\REQUIRE $V_l, \ l = 0, 1, ..., L$. \ENSURE Entanglement Entropy $S_l$. \STATE \textbf{Perform singular value decomposition:} \[ V_l = \Psi_l \Lambda_l (\Phi_l)^T \] where $\Sigma_l = \mathrm{diag}((\lambda_1)_l, (\lambda_2)_l, \dots , (\lambda_{N_p})_l)$. \STATE \textbf{Normalize Schmidt coefficients:} \[ (c_m)_l \gets \frac{(\lambda_m)_l}{\sqrt{\sum_{m'} (\lambda_{m'})_l^2}}. \] \STATE \textbf{Compute entanglement entropy:} \[ S_l \gets - \sum_{m} (c_m)_l^{\,2} \log (c_m)_l^{\,2}. \]

\end{algorithmic}

\end{algorithm}

\input{Tables/notation}

\input{Tables/Transolver_head}

\input{Figures/code/challenge_data}

\input{Tables/2D_NS_6seeds}

\input{Figures/code/Different_Archi_Clarify}

\input{Figures/code/Full_ablations}

\input{Figures/code/Statistical_Plot_123D}

\input{Tables/superesolution}

\input{Tables/era5_exp}

\input{Tables/local_prop_number}

\input{Tables/Bench_Summery}

\input{Tables/Parameters}

\input{Figures/code/over_p_darcy}

\input{Figures/code/over_p_ns}

\input{Figures/code/GSUV_vis}

\input{Figures/code/2D_shallow_mixing}

\input{Figures/code/RT_3D}

\input{Figures/code/Pattern_Scale_128_Burgers}

%% file: Tables/notation.tex
\begin{table}[H]
\centering
\caption{Table of key notations.}
\label{table:notations}

\renewcommand{\arraystretch}{1.5} 

\begin{tabular}{p{0.13\linewidth}|p{0.82\linewidth}}
   \toprule
   {\bf Notation} & {\bf Meaning} \\
   \midrule
   \multicolumn{2}{c}{\bf Problem Setup} \\
   \midrule
   $D_{{x}} \subset \mathbb{R}^d$ & Original domain where functions are defined with $d$ dimensions \\
   ${x} \in D_{{x}}$ & Query position of in/output functions in the $d$ dimensional domain \\
   ${a}({x}) \in  \mathcal{A}$ & Input function \\
   $d_a$ & The length of concatenated input vectors \\
   ${u}({x}) \in  \mathcal{U}$ & Output function \\
   $d_u$ & The length of concatenated output vectors \\
   $\mathcal{G}[\cdot]$ & Solution operator \\
   \midrule
   \multicolumn{2}{c}{\bf $d+1$ dimensional NO Framework} \\
   \midrule
    $p$ & The auxiliary dimension variable composing $d+1$ dimensional PDEs \\
    ${D}_p$ & Domain where the auxiliary variable $p$ lives \\ 
   $v({x}, p)$ & Transformed function in the $d+1$ dimensional domain $D_x \times D_p$\\
   $\mathcal{P}[\cdot]$ & Operator lifts input signals into ones with grids in $D_x \times D_p$ \\
   $\mathcal{L}[\cdot] / \mathcal{L}_l[\cdot]$ & The ($l$-th) learnable linear block capturing $d+1$ dimensional operators\\
   $\sigma_l(\cdot)$ & The ($l$-th) element-wise activation function or 2-layer MLP\\
   $\mathcal{Q}[\cdot]$ & Operator recovers evolved $d+1$ dimensional signals into outputs \\ 
   \midrule
   \multicolumn{2}{c}{\bf Others} \\
   \midrule
    $\hat{(\cdot)}$ & The original function/signal represented in Fourier domain of $x$ \\ 
    $\tilde{(\cdot)}$ & The original function/signal represented in Fourier domain of $p$ \\
    $\widehat{(\cdot)}$ & The original operator/matrix represented in Fourier domain of $x$ \\ 
    $\widetilde{(\cdot)}$ & The original operator/matrix represented in Fourier domain of $p$ \\
    $N_x$ & The number of grid indices on $D_x$ \\
    $N_p$ & The number of grid indices on $D_p$ \\
    $k_{x}$ & The number of grid indices on the transformed domain after aggregation on $D_x$\\
    $k_{p}$ & The number of grid indices on the transformed domain after aggregation on $D_x$\\
   \bottomrule
\end{tabular}

\end{table}

%% file: Tables/Transolver_head.tex
\begin{table}[ht]
\centering
\caption{Relative $L_2$ and $L_1$ errors of Transolver on the 2D Darcy benchmark with different numbers of attention heads.}
\label{tab:transolver_darcy_heads}
\begin{tabular}{ccc}
\toprule
\# Heads & Rel. $L_2$ Error & Rel. $L_1$ Error \\
\midrule
1 & 3.035e-2 & 2.894e-2 \\
2 & 1.848e-2 & 1.684e-2 \\
4 & 1.007e-2 & 8.297e-3 \\
8 & 5.853e-3 & 4.899e-3 \\
\bottomrule
\end{tabular}
\end{table}

%% file: Figures/code/challenge_data.tex
\begin{figure*}[ht]
    \centering
    \includegraphics[width=\textwidth]{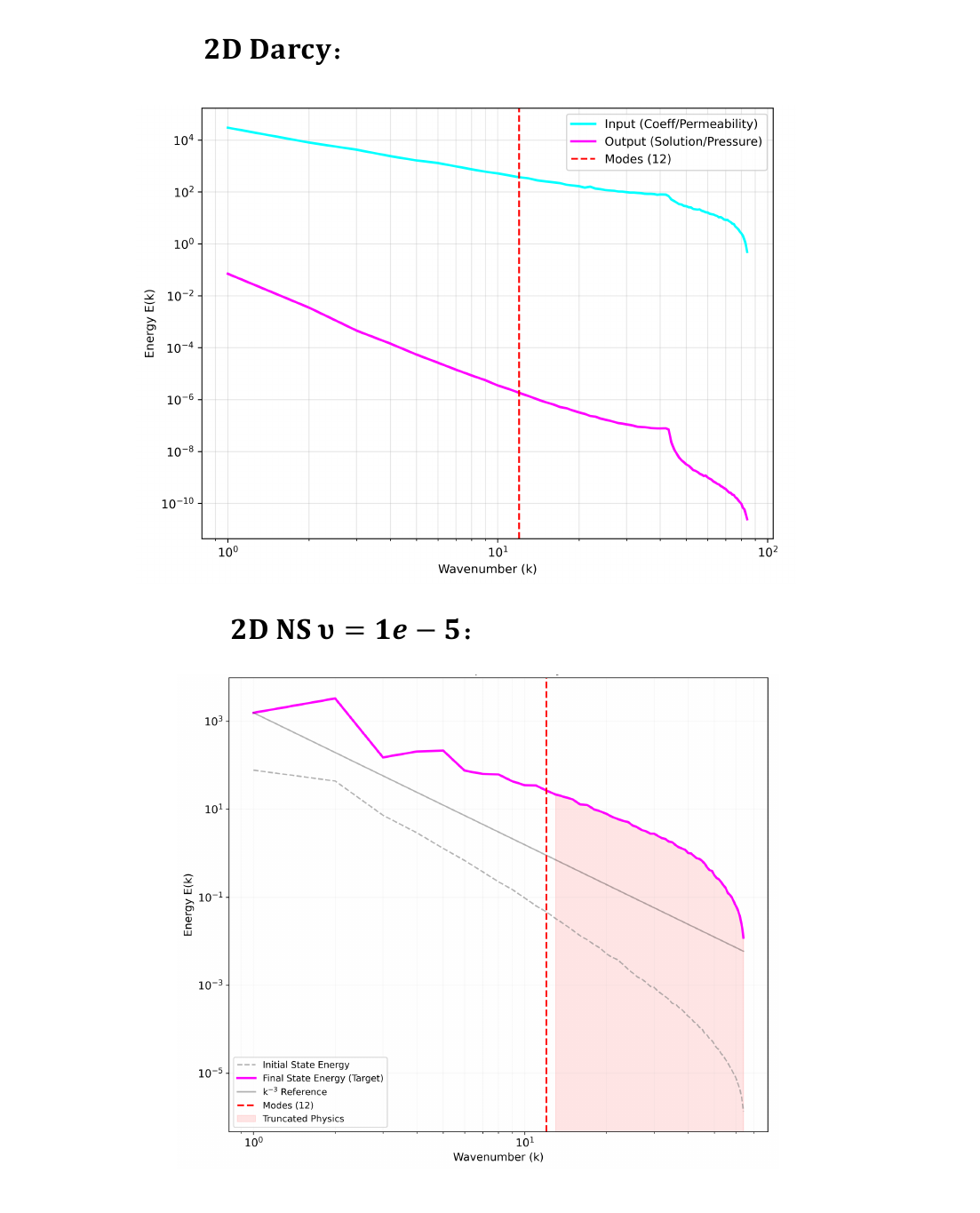}
    \caption{Energy spectral decay for the 2D Darcy flow and 2D Navier-Stokes datasets. The spectral distribution of the data provides insight into the complexity of the operator learning task. Notably, both datasets maintain substantial energy in high-frequency modes, spanning a broad and highly energetic frequency band (note that the wavenumber is plotted on a logarithmic scale). Furthermore, increased operator nonlinearity results in a more irregular spectral decay in the solution (magenta line). In both scenarios, the SKNO model demonstrates superior performance, achieving significantly lower error levels than competing models.}
    \label{fig9:time_unseen}
\end{figure*}

\input{Tables/dplus1_FNO_Transolver}

%% file: Tables/dplus1_FNO_Transolver.tex


\begin{table*}[ht]
\centering
\caption{Performance comparison on the 2D Darcy flow problem evaluating alternative implementations within $d$ dimensional and $(d+1)$ dimensional frameworks. The results are categorized into three distinct blocks. \textbf{Top:} Baseline $d$ dimensional Neural Operators (NOs). \textbf{Middle:} Architectures utilizing lifting within the vector space. Specifically, \textit{Augmented FNO} represents an FNO with expanded channel width ($+1$). \textit{Hyperbolic FNO} introduces an auxiliary vector-dimension modeled as a Lorentz hyperboloid, which creates a curved feature space with constant negative curvature to capture hierarchical scale levels. \textit{2Stack-FNO} stacks two Fourier layers on the $d$-dimensional domain (parallel), omitting specialized operator design along the $p$-dimension. \textbf{Bottom:} \textit{SKNO-Attention} replaces the $p$-dimensional spectral operator in SKNO with the attention-based aggregation module from Transolver, compared against our proposed SKNO.}
\label{tab12:darcy_NO_family}

\resizebox{\textwidth}{!}{%
\begin{tabular}{l|ccccccc}
\toprule
\textbf{Model} & DeepONet & GNOT & OFormer & LSM & CNO & FNO & Transolver \\
\midrule
Rel. $L_2$ Err. & 0.0609 & 0.0102 & 0.0121 & 0.0065 & 0.0060 & 0.0062 & 0.0058 \\
Training Time   & \textbf{5.20 min} & 22.58 min & 21.91 min & 20.37 min & 63.34 min & 10.21 min & 115.48 min \\
\bottomrule
\end{tabular}%
}

\vspace{1em} 

\begin{tabular*}{0.75\textwidth}{l|@{\extracolsep{\fill}}ccc}
\toprule
\textbf{Model} & \textit{Augmented FNO} & \textit{Hyperbolic FNO} & \textit{2Stack-FNO} \\
\midrule
Rel. $L_2$ Err. & 0.0067 & 0.0080 & 0.0061 \\
Training Time   & 10.82 min & 15.83 min & 13.35 min \\
\bottomrule
\end{tabular*}

\vspace{1em}

\begin{tabular*}{0.5\textwidth}{l|@{\extracolsep{\fill}}cc}
\toprule
\textbf{Model} & \textit{SKNO-Attention} & SKNO (Ours) \\
\midrule
Rel. $L_2$ Err. & 0.0058 & \textbf{0.0055} \\
Training Time   & 76.90 min & 17.59 min \\
\bottomrule
\end{tabular*}

\end{table*}

%% file: Tables/2D_NS_6seeds.tex
\begin{table*}[ht]
    \centering
    \caption{Comparison of training times and relative $L_2$ errors for various models evaluated across six independent random seeds. FNO (time dim)$^*$: a variant specifically designed for spatio-temporal function propagation, which preprocesses the data by repeating the spatial field to introduce a pseudo time dimension in the $d$-dimensional physical domain. Within the context of this study, this variant operates directly on $d$ dimensional inputs rather than on the auxiliary embedding dimension after lifting.}
    \label{tab13:ns_stat}
    \renewcommand\arraystretch{1.2}
    \begin{tabular}{l|cc|cc}
    \toprule
    \multirow{2}{*}{\textbf{Model}} 
    & \multicolumn{2}{c|}{\textbf{Seed 0}} 
    & \multicolumn{2}{c}{\textbf{Seed 1}} \\
    \cmidrule(lr){2-3} \cmidrule(lr){4-5}
    & Training Time & Rel. $L_2$ Err. 
    & Training Time & Rel. $L_2$ Err. \\
    \midrule
    DeepONet \cite{21:deeponet_new} & 874.50 & 5.414e-1 & 873.16 & 3.486e-1 \\
    AFNO \cite{22:afno} & 1192.37 & 3.085e-1 & 1127.28 & 3.105e-1 \\
    TFNO \cite{24:TFNO} & 749.36 & 2.270e-1 & 769.06 & 2.245e-1 \\
    U-FNO \cite{22:ufno} & 981.50 & 1.976e-1 & 986.20 & 1.970e-1 \\
    FFNO \cite{23:FFNO} & 754.98 & 2.172e-1 & 753.16 & 2.304e-1 \\
    BregmanNO \cite{25:bregmanNO} & 754.87 & 2.111e-1 & 756.86 & 2.202e-1 \\
    FNO \cite{21:fno} & 843.37 & 1.277e-1 & 856.40 & 1.285e-1 \\
    FNO (time dim) \cite{21:fno} $^{*}$ & 787.14 & 1.642e-1 & 786.21 & 1.663e-1 \\
    Transolver \cite{24:Transolver} & 1263.33 & 1.002e-1 & 1269.94 & 2.086e-1 \\
    \textbf{SKNO (Ours)} & \textbf{720.28} & \textbf{8.374e-2} & \textbf{706.68} & \textbf{8.726e-2} \\
    
    \midrule
    \midrule
    
    \multirow{2}{*}{\textbf{Model}} 
    & \multicolumn{2}{c|}{\textbf{Seed 2}} 
    & \multicolumn{2}{c}{\textbf{Seed 3}} \\
    \cmidrule(lr){2-3} \cmidrule(lr){4-5}
    & Training Time & Rel. $L_2$ Err. 
    & Training Time & Rel. $L_2$ Err. \\
    \midrule
    DeepONet & 885.19 & 3.534e-1 & 872.83 & 3.482e-1 \\
    AFNO & 1139.20 & 3.081e-1 & 1117.37 & 3.062e-1 \\
    TFNO & 733.09 & 2.263e-1 & 770.26 & 2.314e-1 \\
    U-FNO & 999.03 & 1.949e-1 & 1002.90 & 2.000e-1 \\
    FFNO & 770.89 & 2.283e-1 & 772.64 & 2.297e-1 \\
    BregmanNO & 761.34 & 2.195e-1 & 775.92 & 2.212e-1 \\
    FNO & 841.03 & 1.356e-1 & 849.42 & 1.312e-1 \\
    FNO (time dim)$^{*}$ & 803.59 & 1.618e-1 & 789.53 & 1.630e-1 \\
    Transolver & 1469.21 & 8.909e-2 & 1384.41 & 8.406e-2 \\
    \textbf{SKNO (Ours)} & \textbf{736.79} & \textbf{8.355e-2} & \textbf{708.93} & \textbf{8.379e-2} \\
    
    \midrule
    \midrule
    
    \multirow{2}{*}{\textbf{Model}} 
    & \multicolumn{2}{c|}{\textbf{Seed 4}} 
    & \multicolumn{2}{c}{\textbf{Seed 5}} \\
    \cmidrule(lr){2-3} \cmidrule(lr){4-5}
    & Training Time & Rel. $L_2$ Err. 
    & Training Time & Rel. $L_2$ Err. \\
    \midrule
    DeepONet & 878.85 & 3.444e-1 & 880.26 & 3.481e-1 \\
    AFNO & 1110.54 & 3.052e-1 & 1144.27 & 3.078e-1 \\
    TFNO & 755.34 & 2.363e-1 & 768.82 & 2.355e-1 \\
    U-FNO & 984.78 & 2.027e-1 & 997.70 & 1.917e-1 \\
    FFNO & 769.01 & 2.204e-1 & 740.73 & 2.247e-1 \\
    BregmanNO & 767.40 & 2.206e-1 & 755.68 & 2.224e-1 \\
    FNO & 863.67 & 1.403e-1 & 859.44 & 1.370e-1 \\
    FNO (time dim)$^{*}$ & 784.37 & 1.632e-1 & 796.33 & 1.655e-1 \\
    Transolver & 1285.63 & 1.008e-1 & 1278.90 & 9.993e-2 \\
    \textbf{SKNO (Ours)} & \textbf{738.20} & \textbf{8.460e-2} & \textbf{720.98} & \textbf{8.077e-2} \\
    \bottomrule
    \end{tabular}
\end{table*}

%% file: Figures/code/Different_Archi_Clarify.tex
\begin{figure*}[ht]
    \centering
    \includegraphics[width=\textwidth]{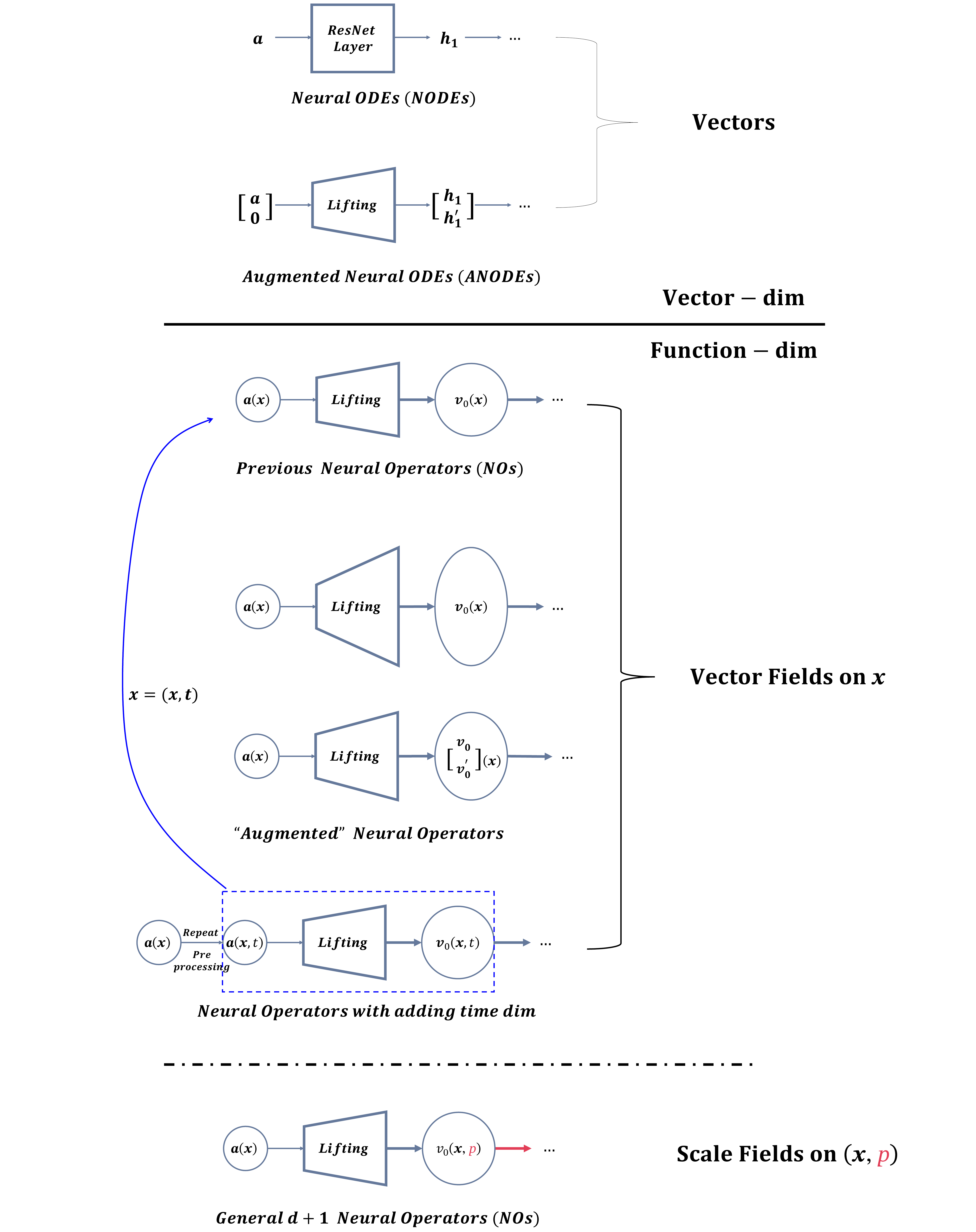}
    \caption{Architectural classifications by input and "dimension" type. \textbf{Top}: Neural ODEs (NODEs) and Augmented NODEs (ANODEs) operate on finite-dimensional vector spaces, using manual concatenation for lifting. \textbf{Middle}: Existing Neural Operators (NOs) propagate $d$ dimensional vector fields, including variants that append a temporal dimension. \textbf{Bottom}: General $d+1$ Neural Operators utilize an auxiliary embedding dimension to propagte $(d+1)$ dimensional scalar fields.}
    \label{fig6:compare}
\end{figure*}

%% file: Figures/code/Full_ablations.tex
\begin{figure*}[ht]
    \centering
    \includegraphics[width=\textwidth]{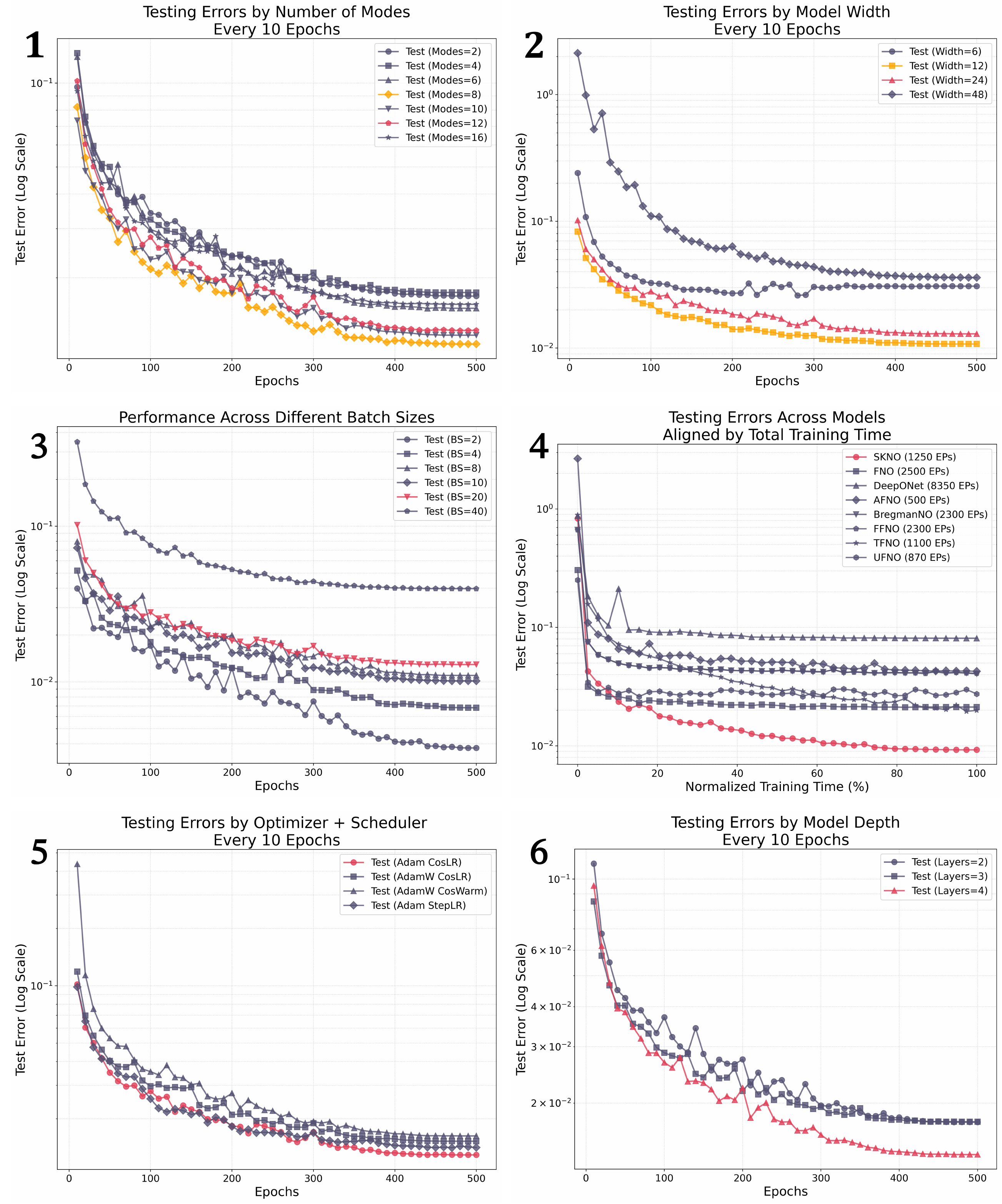}
    \caption{Ablation Studies and Performance Comparison of SKNO on the 2D Gray-Scott. All red lines represent the default configurations of SKNO used in baseline comparisons. \textbf{1} and \textbf{2}: These panels illustrate SKNO's performance under different truncation modes in the original domain and varying sizes along $p$-dim (model width). The yellow lines highlight that SKNO can achieve optimal performance with a smaller number of frequency truncation modes and a more compact $p$-dimension size. \textbf{3}, \textbf{5}, and \textbf{6}: These plots show the influence of batch size, choice of optimizer and learning rate scheduler, and network depth on the model's convergence and final test error. \textbf{4}: This panel displays the test error decay curves for various models aligned by the same total training time, where SKNO successfully surpasses others by above 55.79\% error reduction (\textit{Best testing performance: SKNO with} 9.245e-3 \textit{and FNO with} 2.091e-2).}
    \label{fig:time_unseen}
\end{figure*}

%% file: Figures/code/Statistical_Plot_123D.tex
\begin{figure*}[ht]
    \centering
    \includegraphics[width=\textwidth]{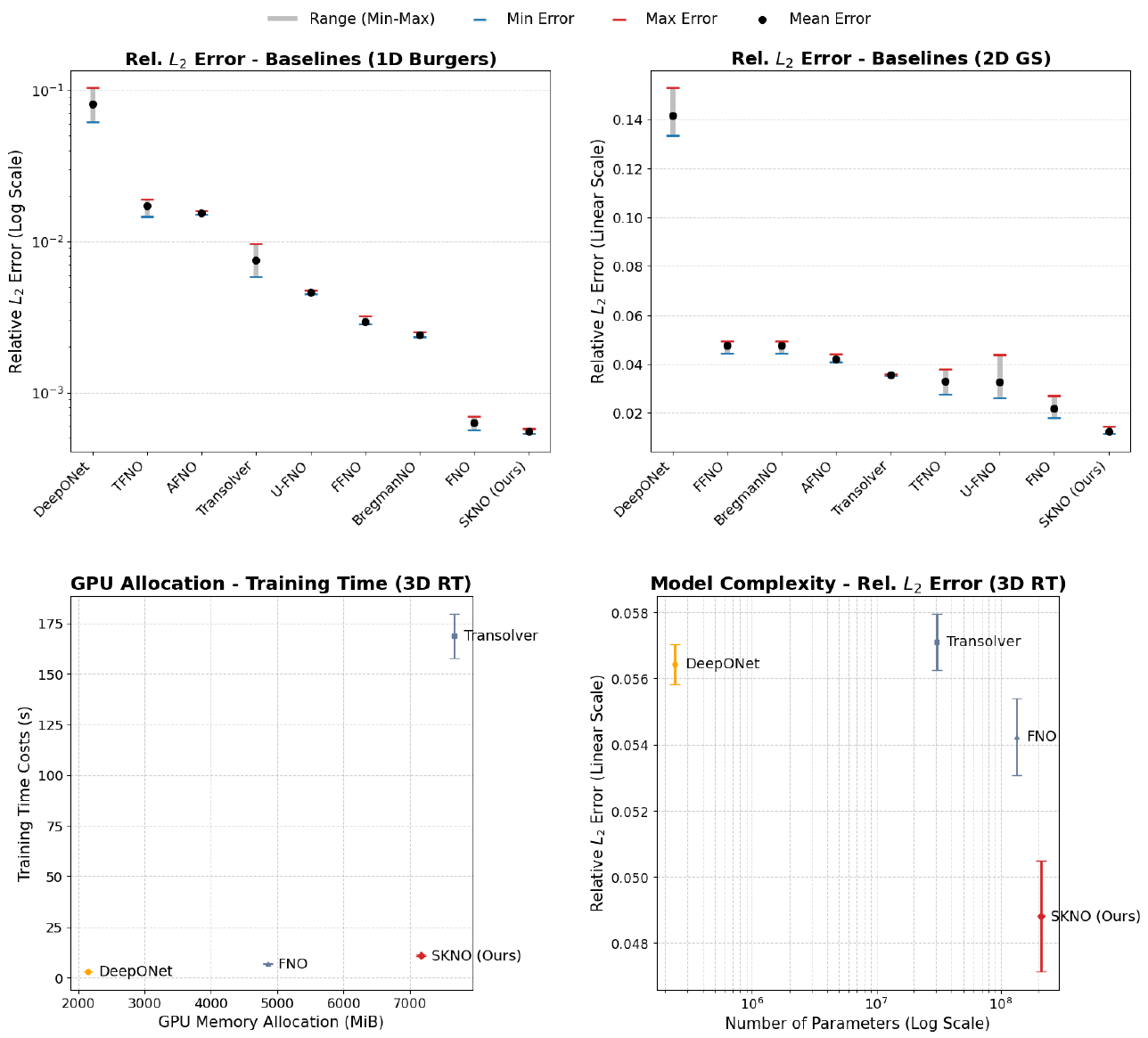}
    \caption{Visualizations of statistical performance across multiple runs with six random seeds. \textbf{Top Row}: On both the 1D Burgers and 2D Gray-Scott problems, our SKNO achieves the best performance within the same number of training epochs compared to all baselines. \textbf{Bottom Row}: On the 3D Rayleigh-Taylor Instability problem, while Transolver employs multi-head mechanisms to mitigate the explosive growth in training time caused by long embeddings, SKNO utilize in the transformed domain of the introduced auxiliary dimension, which drastically reduces training time while delivering a performance leap that remains unattainable for other baseline methods.}
    \label{fig8:time_unseen}
\end{figure*}

%% file: Tables/superesolution.tex
\begin{table}[htbp]
    \centering
    \caption{Prediction errors of neural operators on the 1D Burgers equation and 2D Darcy flow across different grid sizes. The table also reports the variance of errors over grid resolutions, highlighting SKNO’s superior accuracy and stability.}
    \resizebox{\linewidth}{!}{
    \begin{tabular}{c|ccccccc|c}
        \toprule
        \multicolumn{9}{c}{\textit{1D Burgers Equation}} \\
        \midrule
        \textbf{Grid Size} & 128 & 256 & 512 & 1024 & 2048 & 4096 & 8192 & \textbf{Variance} \\
        \midrule
        Transolver & 6.277E-03 & 7.186E-03 & 8.058E-03 & 8.546E-03 & 8.800E-03 & 8.928E-03 & 8.993E-03 & 9.04472E-07 \\
        FNO & 6.479E-04 & 6.476E-04 & 6.480E-04 & 6.477E-04 & 6.454E-04 & 6.455E-04 & 6.455E-04 & 1.33912E-12 \\
        \textbf{SKNO} & \textbf{5.475E-04} & \textbf{5.489E-04} & \textbf{5.490E-04} & \textbf{5.489E-04} & \textbf{5.488E-04} & \textbf{5.487E-04} & \textbf{5.487E-04} & \textbf{2.53997E-13} \\
        \midrule
        \multicolumn{9}{c}{\textit{2D Darcy Flow}} \\
        \midrule
        \textbf{Grid Size} & $(22)^2$ & $(29)^2$ & $(43)^2$ & $(85)^2$ & $(141)^2$ & $(211)^2$ & $(421)^2$ & \textbf{Variance} \\
        \midrule
        FNO & 9.813E-02 & 7.718E-02 & 4.956E-02 & 6.165E-03 & 3.044E-02 & 3.897E-02 & 4.939E-02 & 7.83082E-04 \\
        \textbf{SKNO} & \textbf{7.739E-02} & \textbf{6.302E-02} & \textbf{4.122E-02} & \textbf{5.551E-03} & \textbf{2.366E-02} & \textbf{3.271E-02} & \textbf{4.112E-02} & \textbf{4.90633E-04} \\
        \bottomrule
    \end{tabular}}
    \label{tab14:superes_table}
\end{table}

%% file: Tables/era5_exp.tex
\begin{table}[htbp]
    \centering
    \caption{Zero-shot super-resolution evaluation on ERA5 wind field prediction. As the grid resolution increases, SKNO exhibits substantially lower growth in relative $L_2$ error compared to FNO, leading to a pronounced accuracy advantage on the finest grid $(512)^2$.}
    \resizebox{\linewidth}{!}{
    \begin{tabular}{c|ccccc|c}
        \toprule
        \multicolumn{7}{c}{\textit{Wind U}} \\
        \midrule
        \textbf{Grid Size} & $(32)^2$ & $(64)^2$ & $(128)^2$ & $(256)^2$ & $(512)^2$ & \textbf{Variance} \\
        \midrule
        FNO & 6.272E-02 & 6.525E-02 & 6.872E-02 & 7.097E-02 & 7.212E-02 & 1.23506E-05 \\
        \textbf{SKNO} & \textbf{6.164E-02} & \textbf{6.237E-02} & \textbf{6.366E-02} & \textbf{6.445E-02} & \textbf{6.495E-02} & \textbf{1.54763E-06} \\
        \midrule
        \multicolumn{7}{c}{\textit{Wind V}} \\
        \midrule
        \textbf{Grid Size} & $(32)^2$ & $(64)^2$ & $(128)^2$ & $(256)^2$ & $(512)^2$ & \textbf{Variance} \\
        \midrule
        FNO & 2.165E-01 & 2.326E-01 & 2.430E-01 & 2.476E-01 & 2.498E-01 & 1.49635E-04 \\
        \textbf{SKNO} & \textbf{1.969E-01} & \textbf{2.027E-01} & \textbf{2.108E-01} & \textbf{2.156E-01} & \textbf{2.181E-01} & \textbf{6.30471E-05} \\
        \bottomrule
    \end{tabular}}
    \label{tab:superes_table_era5}
\end{table}

\newcommand{\sd}[1]{{\scriptsize$\pm#1$}}
\begin{table*}[tb!]
    \centering
    \caption{Main benchmarks with each baseline's embedding size matched to SKNO's parameter count. Each benchmark block reports Rel. $L_2$ Error and Training Time, as mean (top) and $\pm$ standard deviation (bottom) over 4 seeds. SKNO has the lowest mean error on all eight benchmarks and trains $1.6$--$68\times$ faster than Transolver.}
    \label{tab:param_matched}
    \renewcommand\arraystretch{1.15}
    \setlength{\tabcolsep}{4pt}

    \resizebox{\textwidth}{!}{%
    \begin{tabular}{l|cc|cc|cc|cc}
    \toprule
    \textbf{Model}
    & \multicolumn{2}{c|}{\textbf{1D Burgers}}
    & \multicolumn{2}{c|}{\textbf{2D Stress}}
    & \multicolumn{2}{c|}{\textbf{2D Strain}}
    & \multicolumn{2}{c}{\textbf{2D Darcy}} \\
    \cmidrule(lr){2-3}
    \cmidrule(lr){4-5}
    \cmidrule(lr){6-7}
    \cmidrule(lr){8-9}
    & Rel. $L_2$ & Time
    & Rel. $L_2$ & Time
    & Rel. $L_2$ & Time
    & Rel. $L_2$ & Time \\
    \midrule

    DeepONet
    & $1.500\mathrm{e}{-1}$ & \textbf{2.49 min}
    & $4.187\mathrm{e}{-1}$ & 14.98 min
    & $5.495\mathrm{e}{-1}$ & 17.09 min
    & $6.508\mathrm{e}{-2}$ & \textbf{11.23 min} \\
    & \sd{1.0\mathrm{e}{-2}} & \sd{0.03}
    & \sd{3.1\mathrm{e}{-3}} & \sd{0.16}
    & \sd{4.4\mathrm{e}{-3}} & \sd{0.95}
    & \sd{2.4\mathrm{e}{-3}} & \sd{0.20} \\

    FNO
    & $5.596\mathrm{e}{-4}$ & 8.96 min
    & $6.127\mathrm{e}{-2}$ & \textbf{11.51 min}
    & $5.577\mathrm{e}{-2}$ & \textbf{11.58 min}
    & $6.847\mathrm{e}{-3}$ & 12.04 min \\
    & \sd{2.0\mathrm{e}{-5}} & \sd{0.11}
    & \sd{5.7\mathrm{e}{-4}} & \sd{0.17}
    & \sd{1.3\mathrm{e}{-3}} & \sd{0.09}
    & \sd{2.9\mathrm{e}{-4}} & \sd{0.14} \\

    Transolver
    & $8.162\mathrm{e}{-3}$ & 88.64 min
    & $4.880\mathrm{e}{-2}$ & 131.56 min
    & $4.879\mathrm{e}{-2}$ & 132.65 min
    & $6.060\mathrm{e}{-3}$ & 108.40 min \\
    & \sd{1.3\mathrm{e}{-3}} & \sd{1.29}
    & \sd{1.4\mathrm{e}{-3}} & \sd{0.92}
    & \sd{1.0\mathrm{e}{-3}} & \sd{2.83}
    & \sd{1.5\mathrm{e}{-4}} & \sd{3.95} \\

    \textbf{SKNO}
    & $\mathbf{5.491\mathrm{e}{-4}}$ & 15.47 min
    & $\mathbf{4.877\mathrm{e}{-2}}$ & 19.44 min
    & $\mathbf{4.777\mathrm{e}{-2}}$ & 18.88 min
    & $\mathbf{5.584\mathrm{e}{-3}}$ & 17.53 min \\
    & \sd{3.9\mathrm{e}{-6}} & \sd{0.12}
    & \sd{1.5\mathrm{e}{-3}} & \sd{0.35}
    & \sd{1.3\mathrm{e}{-3}} & \sd{0.26}
    & \sd{1.3\mathrm{e}{-4}} & \sd{0.12} \\

    \midrule
    \midrule

    \textbf{Model}
    & \multicolumn{2}{c|}{\textbf{2D Gray-Scott}}
    & \multicolumn{2}{c|}{\textbf{2D Inc. NS}}
    & \multicolumn{2}{c|}{\textbf{3D Comp. NS}}
    & \multicolumn{2}{c}{\textbf{3D RT}} \\
    \cmidrule(lr){2-3}
    \cmidrule(lr){4-5}
    \cmidrule(lr){6-7}
    \cmidrule(lr){8-9}
    & Rel. $L_2$ & Time
    & Rel. $L_2$ & Time
    & Rel. $L_2$ & Time
    & Rel. $L_2$ & Time \\
    \midrule

    DeepONet
    & $1.550\mathrm{e}{-1}$ & \textbf{6.06 min}
    & $4.278\mathrm{e}{-1}$ & 953.03 min
    & $5.496\mathrm{e}{-1}$ & 152.15 min
    & $5.716\mathrm{e}{-2}$ & 79.43 min \\
    & \sd{1.6\mathrm{e}{-2}} & \sd{0.01}
    & \sd{8.7\mathrm{e}{-2}} & \sd{3.45}
    & \sd{1.6\mathrm{e}{-2}} & \sd{9.00}
    & \sd{1.7\mathrm{e}{-4}} & \sd{0.08} \\

    FNO
    & $1.877\mathrm{e}{-2}$ & 11.56 min
    & $1.402\mathrm{e}{-1}$ & 973.62 min
    & $2.676\mathrm{e}{-1}$ & \textbf{18.43 min}
    & $5.504\mathrm{e}{-2}$ & \textbf{8.79 min} \\
    & \sd{7.2\mathrm{e}{-4}} & \sd{0.03}
    & \sd{3.7\mathrm{e}{-2}} & \sd{13.83}
    & \sd{4.8\mathrm{e}{-3}} & \sd{0.49}
    & \sd{4.9\mathrm{e}{-4}} & \sd{0.49} \\

    Transolver
    & $1.881\mathrm{e}{-2}$ & 235.88 min
    & $1.344\mathrm{e}{-1}$ & 1187.05 min
    & $2.966\mathrm{e}{-1}$ & 503.30 min
    & $5.732\mathrm{e}{-2}$ & 755.02 min \\
    & \sd{9.6\mathrm{e}{-4}} & \sd{0.67}
    & \sd{2.8\mathrm{e}{-2}} & \sd{19.39}
    & \sd{2.4\mathrm{e}{-2}} & \sd{9.79}
    & \sd{3.8\mathrm{e}{-4}} & \sd{13.48} \\

    \textbf{SKNO}
    & $\mathbf{1.264\mathrm{e}{-2}}$ & 20.42 min
    & $\mathbf{8.485\mathrm{e}{-2}}$ & \textbf{721.25 min}
    & $\mathbf{2.367\mathrm{e}{-1}}$ & 22.82 min
    & $\mathbf{4.803\mathrm{e}{-2}}$ & 11.06 min \\
    & \sd{7.8\mathrm{e}{-4}} & \sd{0.28}
    & \sd{2.0\mathrm{e}{-3}} & \sd{15.08}
    & \sd{7.3\mathrm{e}{-3}} & \sd{0.47}
    & \sd{4.5\mathrm{e}{-3}} & \sd{0.65} \\

    \bottomrule
    \end{tabular}%
    }
\end{table*}

\input{Tables/other_experiment}

\begin{table*}[htbp]
    \caption{Results for ERA5 wind field prediction.}
    \label{tab:era5}

    \centering

    \resizebox{0.8\textwidth}{!}{\begin{tabular}{c|ccc}
\hline
\diagbox{Model}{Performance} & Training Time & Wind U Error & Wind V Error \\ \hline
FNO & \textbf{0.76 min} & 7.064e-2 & 2.133e-1 \\ \hline
Transolver & 10.81 min & {6.291e-2} & {2.007e-1} \\ \hline
SKNO & {1.75 min} & \textbf{6.111e-2} & \textbf{1.920e-1} \\ \hline
\end{tabular}}
\end{table*}

%% file: Tables/other_experiment.tex
\begin{table}
\centering
\caption{Performance of neural operator models on 2D Shallow Water and 3D Compressible NS benchmarks. On the 2D Shallow Water, we train models on a randomly selected resolution and evaluate them at the highest resolution. --: No 3D implementation available; $^{*}$: Requires predefining the resolution, here trained on the highest resolution; $^{**}$: For fairness, training times are all reported on the highest resolutions. The CNO model is described in (\cite{23:CNO}).}
\setlength{\tabcolsep}{6pt}
\renewcommand{\arraystretch}{1.2}
\begin{tabular}{lccccc}
\hline
\textbf{Models} & {DeepONet} & {CNO} & {FNO} & {Transolver} & {SKNO} \\
\hline
Rel.\ $L_2$ Err. & 0.0902$^{*}$ & 0.0085$^{*}$ & 0.0030 & 0.0088 & \textbf{0.0027} \\
Training Time$^{**}$ & \textbf{8.59 min} & 49.26 min & 11.43 min & 128.64 min & 20.85 min \\
\hline
\end{tabular}
\label{tab:other_exp_bench}
\end{table}

%% file: Tables/local_prop_number.tex
\begin{table}[t]
\centering
\caption{Comparison of different configurations with zero, one, and two local propagators. Results show that adding local aggregation layers helps signals integrate residual information more effectively with a lower level of entanglement.}
\label{tab:darcy_local_props}
\resizebox{0.9\linewidth}{!}{%
\begin{tabular}{lccc}
\toprule
\textbf{Metric} & {w.o. Local Prop} & {1 Local Prop} & {2 Local Props} \\
\midrule
Rel. $L_2$ Err. & 5.717 & 5.555 & 5.289 \\
Entanglement Entropy (Before Recovering) & 2.286 & 1.849 & 1.277 \\
\bottomrule
\end{tabular}%
}
\end{table}

%% file: Tables/Bench_Summery.tex
\begin{table*}[t]

    \caption{Summary of benchmark dataset characteristics for the evaluated problems. The table details the number of samples in the training and testing sets, the spatial resolution or number of grid points ($N_x$), the input feature dimensionality ($d_a$), and the output field dimensionality ($d_u$).}
    \label{tab20:sum_bench}
    \begin{center}
    \begin{small}
    \begin{sc}
    \begin{tabular}{lccccc}
        \toprule
        \textbf{Benchmark} & \textbf{Train. - Test. Samp.} & $\boldsymbol{N_x}$ & $\boldsymbol{d_a}$ - $\boldsymbol{d_u}$ \\
        \midrule
        1D Heat (Time) & 1000 - 100 & 256 & 1 - 1 \\
        1D Advection (Time) & 1000 - 200 & 256 & 10 - 1 \\
        1D Burgers (Time) & 1000 - 100 & 128 & 1 - 1 \\
        2D Stress & 1000 - 100 & 2304 & 1 - 3 \\
        2D Strain & 1000 - 100 & 2304 & 1 - 3 \\
        2D Darcy & 1000 - 100 & 7225 & 1 - 1 \\
        2D Gray-Scott (Time) & 200 - 20 & 1024 & 20 - 2 \\
        2D Shallow Water (Time) & 500 - 300 & 1024/4096/16384$^\dagger$ & 10 - 1 \\
        2D Incompr. Navier-Stokes (1e-4, Time) & 1000 - 200 & 4096 & 10 - 1 \\
        2D Incom. Navier-Stokes (1e-5, Time) & 1000 - 200 & 4096 & 10 - 1 \\
        3D Com. Navier-Stokes (Time) & 500 - 100 & 32768 & 15 - 5 \\
        3D Rayleigh-Taylor Instab. (Time) & 220 - 40 & 32768 & 4 - 4 \\
        ERA5 Wind Field Prediction (Time) & 64 - 16 & 262144 & 4 - 1 \\
        \bottomrule
    \end{tabular}
    \end{sc}
    \end{small}
    
    \vspace{0.15cm}
    \begin{minipage}{0.85\textwidth} 
    \footnotesize \textsuperscript{$\dagger$}For the 2D Shallow Water benchmark, spatial resolutions ($N_x$) are randomly downsampled to grids of $32\times32$, $64\times64$, and $128\times128$ (maximum).
    \end{minipage}
    
    \end{center}
    \vskip -0.1in
\end{table*}

%% file: Tables/Parameters.tex
\begin{table*}[t]
\centering
\caption{Comparison of parameter counts across different models and problems. Follow the recommended configurations as the convention of NO comparisons \cite{24:Transolver}.}
\label{tab:model_params}
\small 
\begin{tabular*}{\textwidth}{@{\extracolsep{\fill}} l r r r r}
\toprule
\textbf{Problem} & \textbf{DeepONet} & \textbf{Transolver} & \textbf{FNO} & \textbf{SKNO} \\
\midrule
2D Gray-Scott      & 86,802   & 11,237,698  & 1,337,450 & 2,141,844 \\
2D Shallow-Water   & 16,844,624 & 2,175,473   & 599,833  & 971,636 \\
2D Navier-Stokes   & 4,261,712 & 11,232,321  & 1,337,113 & 2,149,884 \\
3D Navier-Stokes   & 249,989   & 30,339,653 & 134,235,877 & 210,028,292 \\
3D RT              & 240,420 & 30,333,764 & 134,235,396 & 210,061,700 \\
\bottomrule
\end{tabular*}
\end{table*}

%% file: Figures/code/over_p_darcy.tex
\begin{figure*}
    \centering
    \includegraphics[width=\textwidth]{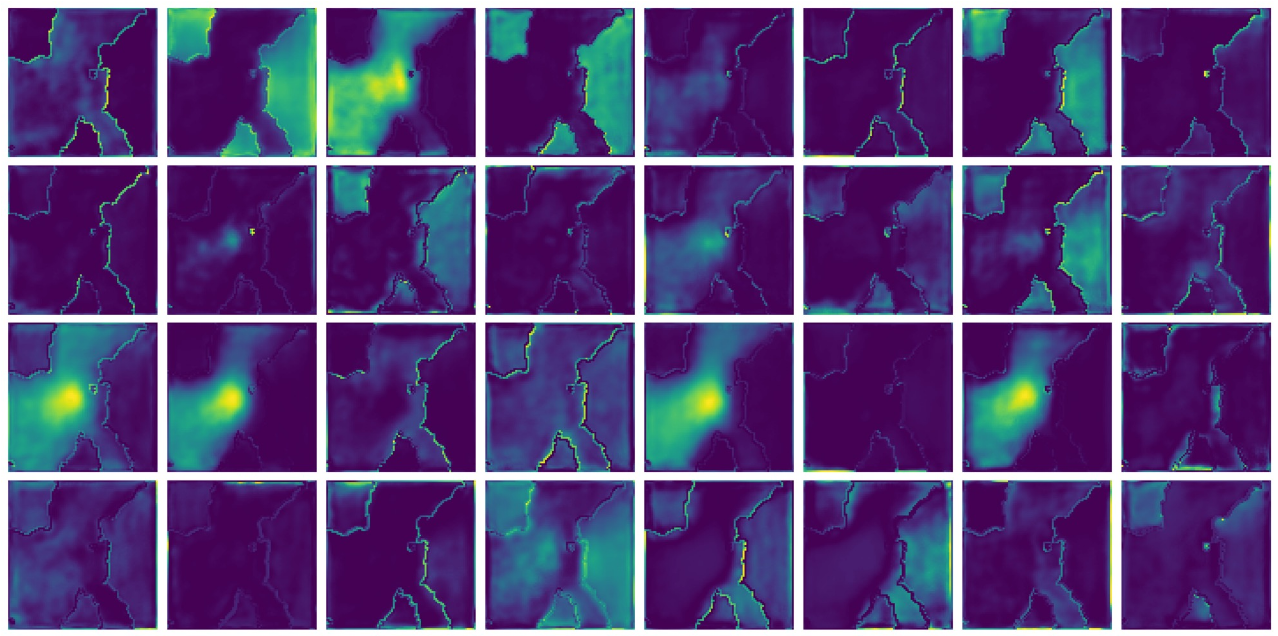}
    \caption{Pattern visualization of the 2D Darcy case. According to the discontinuities in the inputs, SKNO learns patterns at different scales.}
    \label{fig10:over_p_darcy}
\end{figure*}

%% file: Figures/code/over_p_ns.tex
\begin{figure*}
    \centering
    \includegraphics[width=\textwidth]{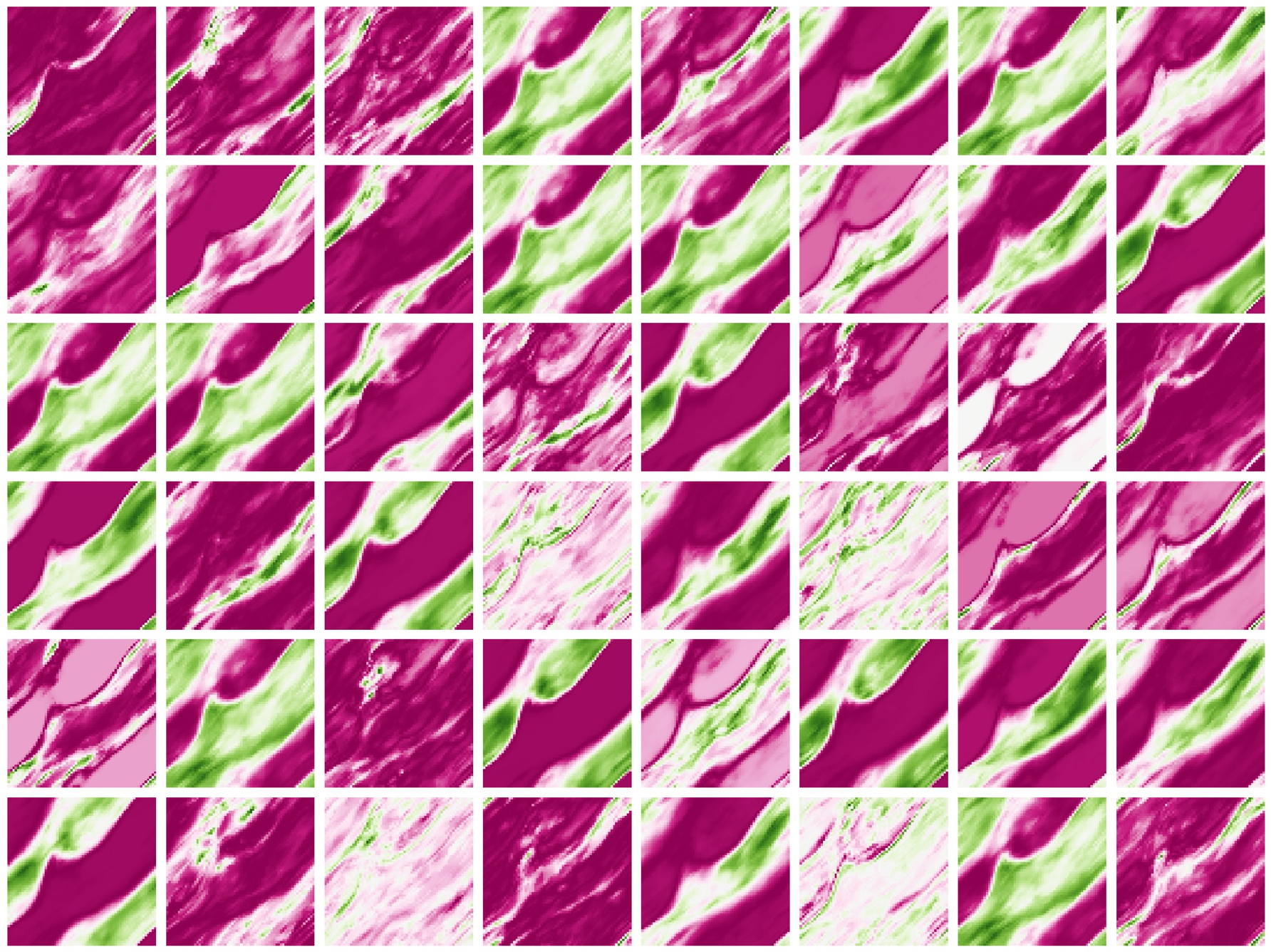}
    \caption{Pattern visualization of the 2D Navier-Stokes case.}
    \label{fig11:over_p_ns}
\end{figure*}

%% file: Figures/code/GSUV_vis.tex
\begin{figure*}
    \centering
    \includegraphics[width=\textwidth]{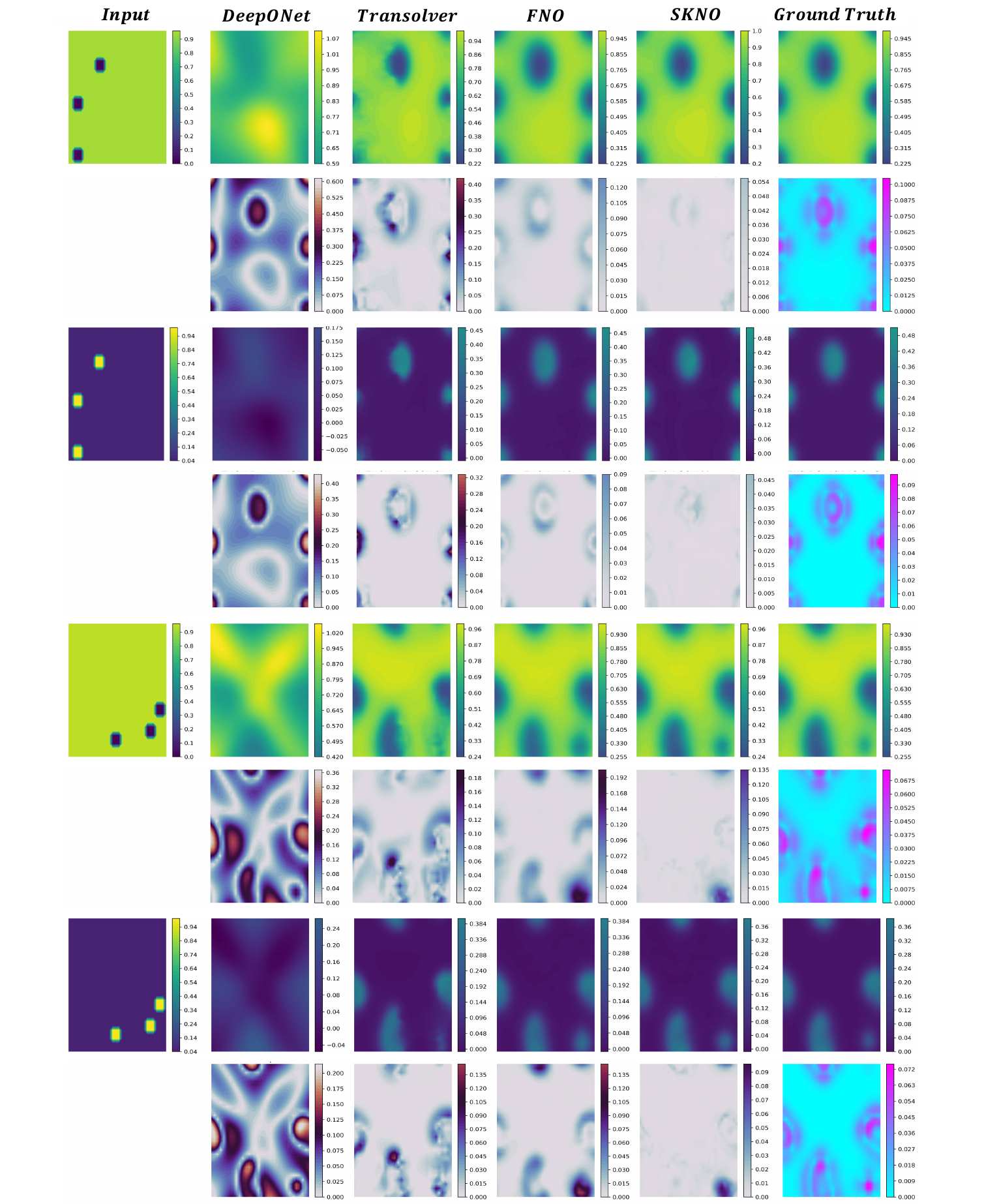}
    \caption{Visualization of model predictions for the 2D Gray-Scott dataset. The top row of each block displays the field $u(x)$, while the bottom row displays the field $v(x)$. Across all presented cases and physical fields, SKNO significantly suppresses error artifacts and reconstructs the ground truth with vastly superior precision compared to the baseline architectures.}
    \label{fig12:GS}
\end{figure*}

%% file: Figures/code/2D_shallow_mixing.tex
\begin{figure*}[ht]
    \centering
    {\includegraphics[width=\linewidth]{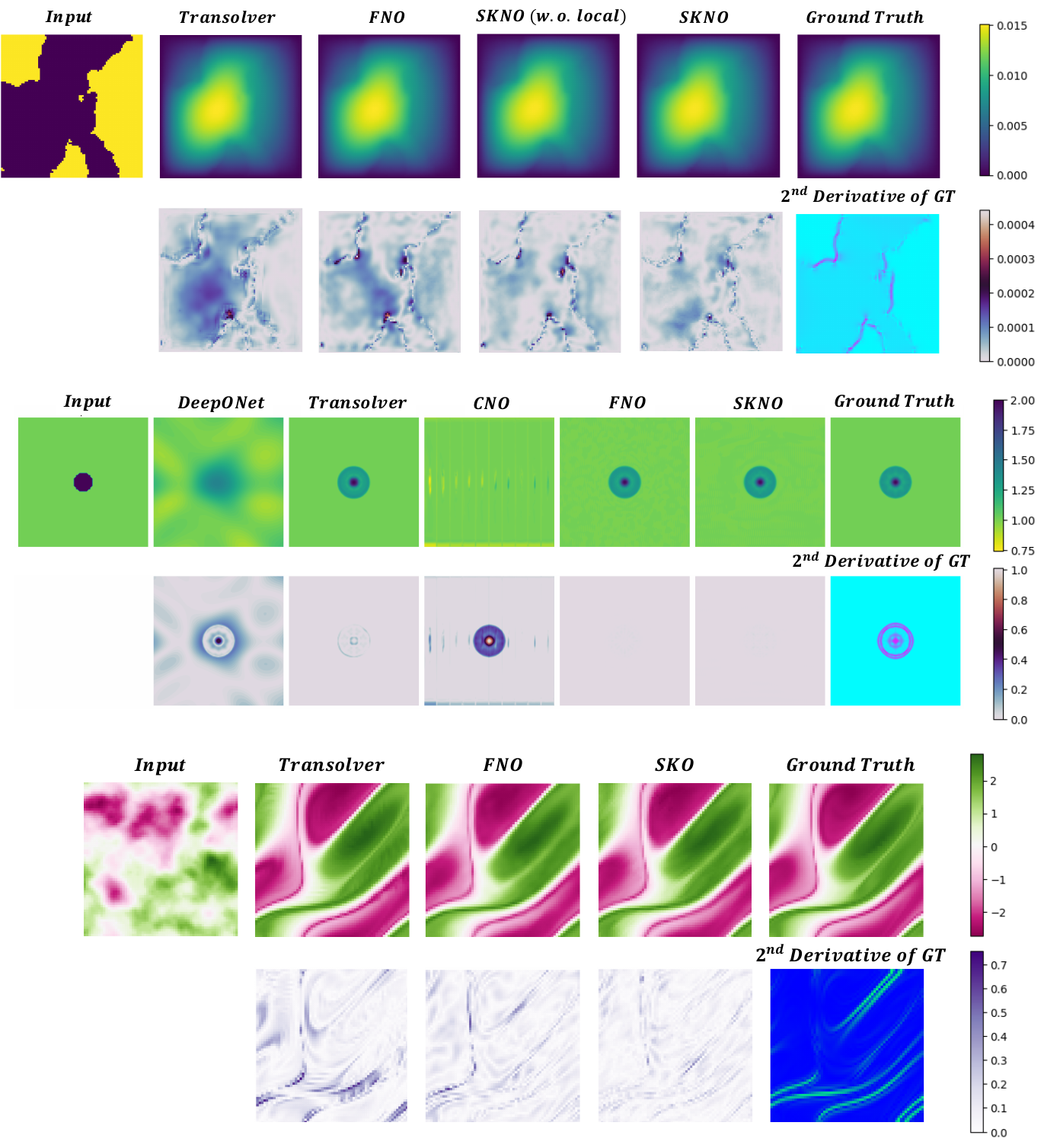}}
    \caption{Visualization of predictions, errors, and the second derivative of the ground truth for 2D Darcy flow, 2D Shallow-Water equations (using randomly mixed training data), and 2D Navier-Stokes (NS) equations. In each block, the top row displays the input field, predictions from various baseline models alongside our proposed method (SKNO), and the ground truth. The bottom row presents the corresponding error visualizations for each model's prediction, as well as the second derivative of the ground truth. Overall, these comparisons illustrate the consistently low error of SKNO on physical field mapping operators across both microscopic and macroscopic scales.}
    \label{fig:showcase_darcy}
\end{figure*}

%% file: Figures/code/RT_3D.tex
\begin{figure*}
    \centering
    \includegraphics[width=\textwidth]{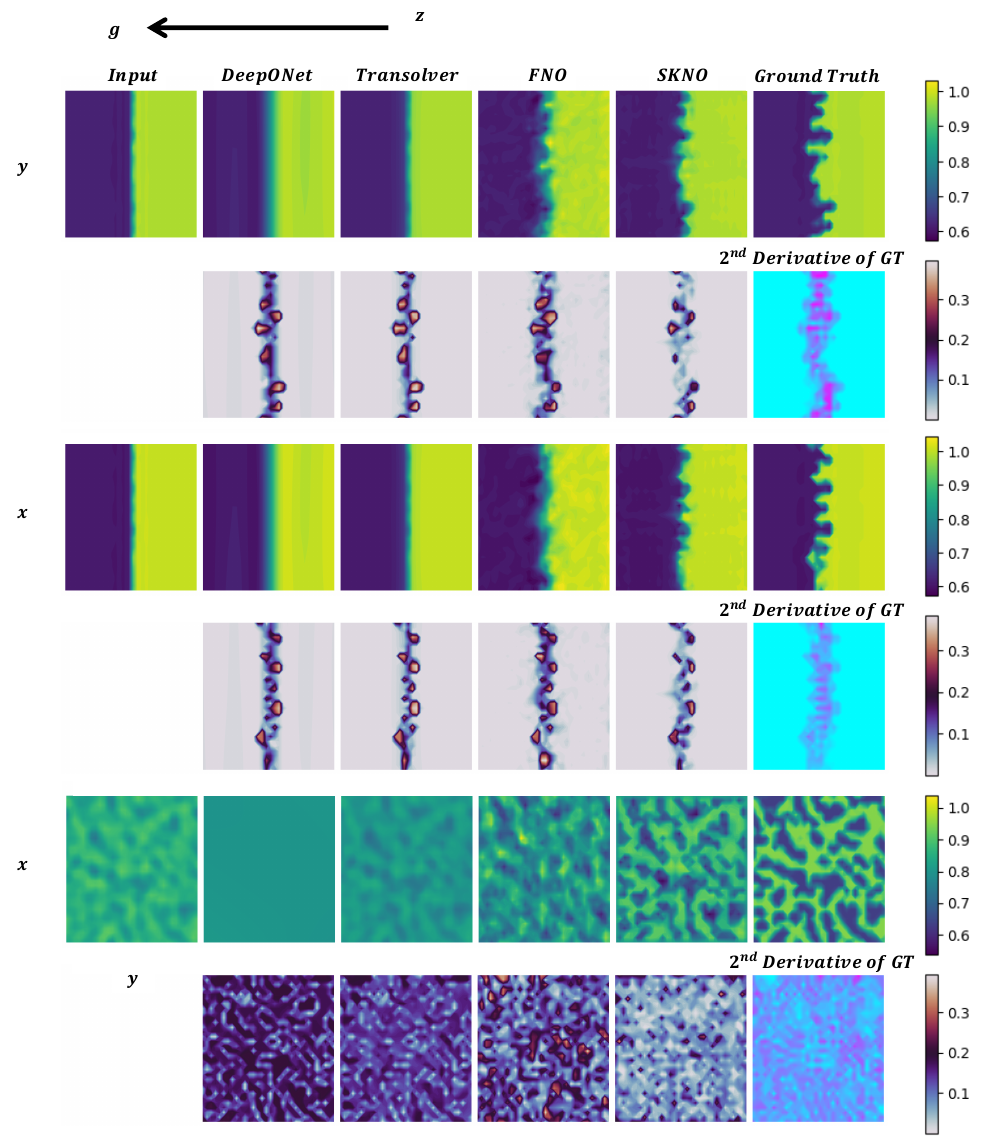}
    \caption{Visualization of the 3D Rayleigh–Taylor instability case. All slices are parallel to the coordinate planes, and their intersection points are chosen as the grid points nearest to the center of the box. We show a representative extreme case, namely predicting the output field after 20 steps from a perturbed initial condition. While other lightweight methods largely fail, SKNO can still capture the overall flow trends and the interfacial surfaces between the fluids with different densities. $\boldsymbol{g}$ represents the gravity direction.}
    \label{fig:RT}
\end{figure*}

%% file: Figures/code/Pattern_Scale_128_Burgers.tex
\begin{figure*}
    \centering
    \includegraphics[width=\textwidth]{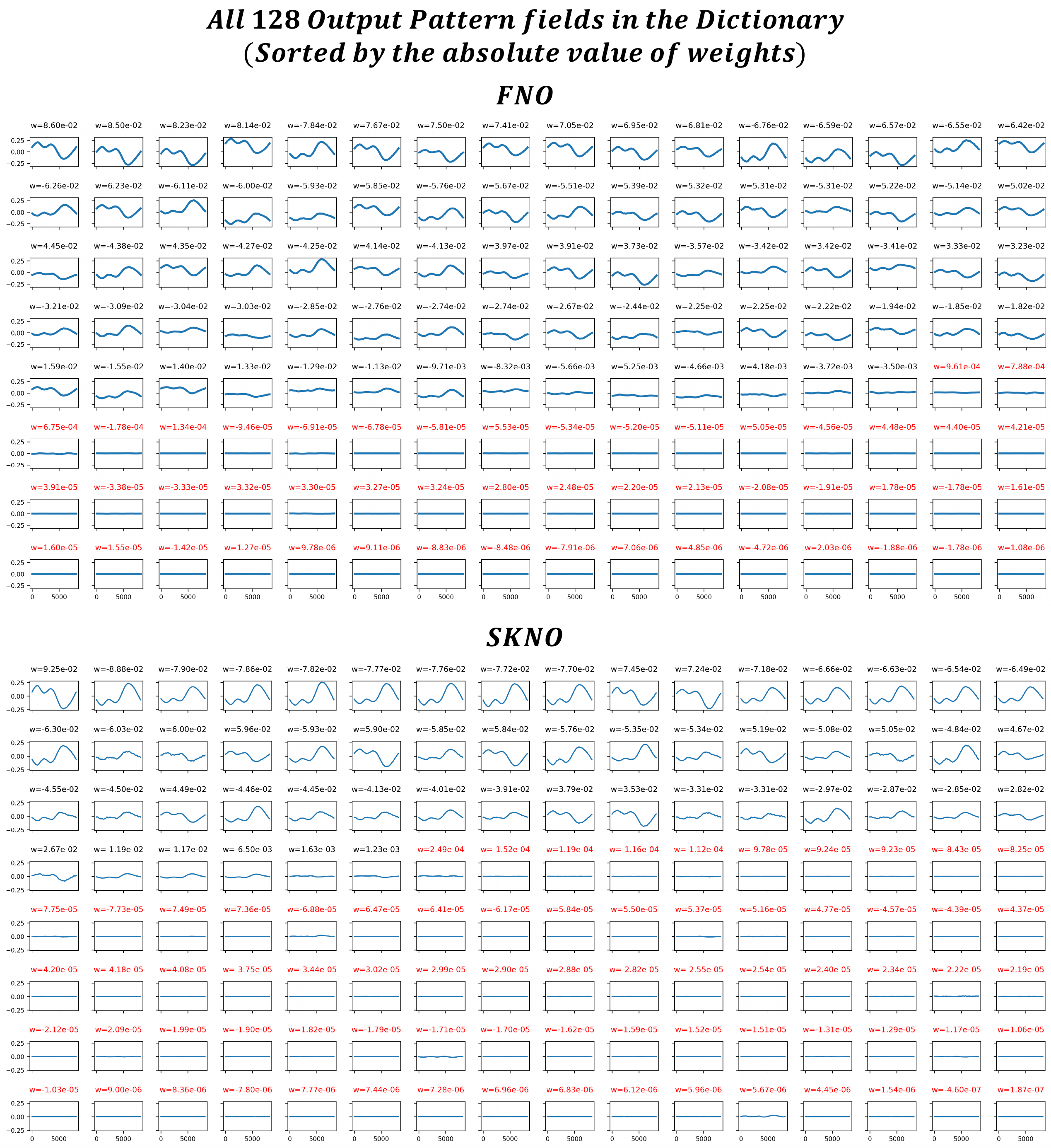}
    \caption{Visualization of paired output weights and pattern fields. All pattern fields are sorted in descending order of the absolute value of their associated output weights, and values marked in red lie below the contribution threshold of $10^{-3}$. Even with the same sufficiently large $N_p$, the relative $L_2$ error of SKNO on the testing data $6.546e-4$ is still smaller than FNO $6.603e-4$ (Redundant $p$-width but still benefit from the basis-diversified evolution).}
    \label{fig:pattern_scale_128_sorted_weight}
\end{figure*}

%% file: reference.bib
@article{89:universal_nn,
  title={Multilayer feedforward networks are universal approximators},
  author={Hornik, Kurt and Stinchcombe, Maxwell and White, Halbert},
  journal={Neural Networks},
  volume={2},
  number={5},
  pages={359--366},
  year={1989},
  publisher={Elsevier}
}

@book{92:wavelets,
  title={Wavelets and operators},
  author={Meyer, Yves},
  number={37},
  year={1992},
  publisher={Cambridge university press}
}

@article{95:universal_no,
  title={Universal approximation to nonlinear operators by neural networks with arbitrary activation functions and its application to dynamical systems},
  author={Chen, Tianping and Chen, Hong},
  journal={IEEE Transactions on Neural Networks},
  volume={6},
  number={4},
  pages={911--917},
  year={1995},
  publisher={IEEE}
}

@article{95:fast_level_set,
  title={A fast level set method for propagating interfaces},
  author={Adalsteinsson, David and Sethian, James A},
  journal={Journal of computational physics},
  volume={118},
  number={2},
  pages={269--277},
  year={1995},
  publisher={Elsevier}
}

@article{98:ANN_for_ODEPDE,
  title={Artificial neural networks for solving ordinary and partial differential equations},
  author={Lagaris, Isaac E and Likas, Aristidis and Fotiadis, Dimitrios I},
  journal={IEEE Transactions on Neural Networks},
  volume={9},
  number={5},
  pages={987--1000},
  year={1998},
  publisher={IEEE}
}

@book{02:symmetry_analy,
  title={Introduction to symmetry analysis},
  author={Cantwell, Brian J},
  year={2002},
  publisher={Cambridge University Press}
}

@article{01:level_set,
  title={Level set methods: an overview and some recent results},
  author={Osher, Stanley and Fedkiw, Ronald P},
  journal={Journal of Computational physics},
  volume={169},
  number={2},
  pages={463--502},
  year={2001},
  publisher={Elsevier}
}

@article{03:level_riemann,
  title={A level set method for the computation of multi-valued solutions to quasi-linear hyperbolic PDE's and Hamilton-Jacobi equations},
  author={Jin, Shi and Osher, Stanley},
  journal={Communications in Mathematical Sciences},
  volume={1},
  number={1},
  pages={575--591},
  year={2003}
}

@book{04:galerkin_fem,
  title={Theory and practice of finite elements},
  author={Ern, Alexandre and Guermond, Jean-Luc},
  volume={159},
  year={2004},
  publisher={Springer}
}

@article{04:levelset,
  title={Level set methods and dynamic implicit surfaces},
  author={Osher, Stanley and Fedkiw, Ronald and Piechor, K},
  journal={Appl. Mech. Rev.},
  volume={57},
  number={3},
  pages={B15--B15},
  year={2004}
}

@article{04:rom,
  title={Reduced-order modeling: new approaches for computational physics},
  author={Lucia, David J and Beran, Philip S and Silva, Walter A},
  journal={Progress in aerospace sciences},
  volume={40},
  number={1-2},
  pages={51--117},
  year={2004},
  publisher={Elsevier}
}

@article{11:pgd,
  title={A short review on model order reduction based on proper generalized decomposition},
  author={Chinesta, Francisco and Ladeveze, Pierre and Cueto, Elias},
  journal={Archives of Computational Methods in Engineering},
  volume={18},
  number={4},
  pages={395--404},
  year={2011},
  publisher={Springer}
}

@inproceedings{14:adam,
  author       = {Kingma, Diederik P. and Lei Ba, Jimmy},
  title        = {Adam: {A} Method for Stochastic Optimization},
  booktitle    = {the 3rd International Conference
on Learning Representations},
  year         = {2015}
}

@inproceedings{16:resnet,
  title={Deep residual learning for image recognition},
  author={He, Kaiming and Zhang, Xiangyu and Ren, Shaoqing and Sun, Jian},
  booktitle={Proceedings of the IEEE conference on computer vision and pattern recognition},
  pages={770--778},
  year={2016}
}

@article{17:transformer,
  title={Attention is all you need},
  author = {Vaswani, Ashish and Shazeer, Noam and Parmar, Niki and Uszkoreit, Jakob and Jones, Llion and Gomez, Aidan N. and Kaiser, \L{}ukasz and Polosukhin, Illia},
  journal={Advances in Neural Information Processing Systems},
  year={2017}
}

@inproceedings{19:AdamW,
  author={Loshchilov, Ilya and Hutter, Frank},
  title        = {Decoupled Weight Decay Regularization},
  booktitle    = {the 7th International Conference on Learning Representations},
  year         = {2019}
}

@inproceedings{20:cfd_net,
  title={CFDNet: A deep learning-based accelerator for fluid simulations},
  author={Obiols-Sales, Octavi and Vishnu, Abhinav and Malaya, Nicholas and Chandramowliswharan, Aparna},
  booktitle={Proceedings of the 34th ACM International Conference on Supercomputing},
  pages={1--12},
  year={2020}
}

@inproceedings{20:GNO,
  title={Neural Operator: Graph Kernel Network for Partial Differential Equations},
  author={Li, Zongyi and Kovachki, Nikola and Azizzadenesheli, Kamyar and Liu, Burigede and Bhattacharya, Kaushik and Stuart, Andrew and Anandkumar, Anima},
  booktitle={the 8th International Conference on Learning Representations},
  year={2020},
  series = {ICLR 2020 Workshop: Integration of Deep Neural Models and Differential Equations}
}

@inproceedings{20:MGNO,
  author={Li, Zongyi and Kovachki, Nikola and Azizzadenesheli, Kamyar and Liu, Burigede and Stuart, Andrew and Bhattacharya, Kaushik and Anandkumar, Anima},
  title        = {Multipole Graph Neural Operator for Parametric Partial Differential Equations},
  booktitle    = {Advances in Neural Information Processing Systems},
  year         = {2020}
}

@article{21:deeponet_new,
  title={Learning nonlinear operators via DeepONet based on the universal approximation theorem of operators},
  author={Lu, Lu and Jin, Pengzhan and Pang, Guofei and Zhang, Zhongqiang and Karniadakis, George Em},
  journal={Nature Machine Intelligence},
  volume={3},
  number={3},
  pages={218--229},
  year={2021},
  publisher={Nature Publishing Group UK London}
}

@article{21:universal_fno,
  title={On universal approximation and error bounds for Fourier neural operators},
  author={Kovachki, Nikola and Lanthaler, Samuel and Mishra, Siddhartha},
  journal={Journal of Machine Learning Research},
  volume={22},
  number={290},
  pages={1--76},
  year={2021}
}

@inproceedings{21:fno,
  author={Li, Zongyi and Kovachki, Nikola and Azizzadenesheli, Kamyar and Liu, Burigede and Bhattacharya, Kaushik and Stuart, Andrew and Anandkumar, Anima},
  title        = {Fourier Neural Operator for Parametric Partial Differential Equations},
  booktitle    = {the 9th International Conference on Learning Representations},
  year         = {2021}
}

@inproceedings{21:Choose,
  title={Choose a Transformer: Fourier or Galerkin},
  author = {Cao, Shuhao},
  booktitle={Advances in Neural Information Processing Systems},
  year={2021},
}

@article{21:parametric_PDE_ANN,
  title={Solving parametric PDE problems with artificial neural networks},
  author={Khoo, Yuehaw and Lu, Jianfeng and Ying, Lexing},
  journal={European Journal of Applied Mathematics},
  volume={32},
  number={3},
  pages={421--435},
  year={2021},
  publisher={Cambridge University Press}
}

@article{22:cnn_solver,
  title={Learning time-dependent PDEs with a linear and nonlinear separate convolutional neural network},
  author={Qu, Jiagang and Cai, Weihua and Zhao, Yijun},
  journal={Journal of Computational Physics},
  volume={453},
  pages={110928},
  year={2022},
  publisher={Elsevier}
}

@article{22:PDNO,
  title={Pseudo-Differential Neural Operator: Generalized Fourier Neural Operator for Learning Solution Operators of Partial Differential Equations},
  author={Shin, Jin Young and Lee, Jae Yong and Hwang, Hyung Ju},
  journal={ArXiv},
  year={2022},
  volume={abs/2201.11967}
}

@article{22:schrodingerisation_JIN_tech,
  title={Quantum simulation of partial differential equations via schrodingerisation: technical details},
  author={Jin, Shi and Liu, Nana and Yu, Yue},
  journal={ArXiv},
  year={2022},
  volume={abs/2212.14703}
}

@article{22:PDEBench,
  title={Pdebench: An extensive benchmark for scientific machine learning},
  author={Takamoto, Makoto and Praditia, Timothy and Leiteritz, Raphael and MacKinlay, Daniel and Alesiani, Francesco and Pfl{\"u}ger, Dirk and Niepert, Mathias},
  journal={Advances in Neural Information Processing Systems},
  year={2022}
}

@article{22:ufno,
  title={U-FNO—An enhanced Fourier neural operator-based deep-learning model for multiphase flow},
  author={Wen, Gege and Li, Zongyi and Azizzadenesheli, Kamyar and Anandkumar, Anima and Benson, Sally M},
  journal={Advances in Water Resources},
  volume={163},
  pages={104180},
  year={2022},
  publisher={Elsevier}
}

@article{23:schroedingerisation_JIN,
  title={Quantum simulation of partial differential equations: Applications and detailed analysis},
  author={Jin, Shi and Liu, Nana and Yu, Yue},
  journal={Physical Review A},
  volume = {108},
  pages = {032603},
  numpages = {20},
  year = {2023},
  publisher = {American Physical Society},
}

@article{23:function_space_no,
  author = {Kovachki, Nikola and Li, Zongyi and Liu, Burigede and Azizzadenesheli, Kamyar and Bhattacharya, Kaushik and Stuart, Andrew and Anandkumar, Anima},
  title        = {Neural Operator: Learning Maps Between Function Spaces With Applications to PDEs},
  journal      = {Journal of Machine Learning Research},
  volume       = {24},
  pages        = {89:1--89:97},
  year         = {2023}
}

@article{23:geofno,
  title={Fourier neural operator with learned deformations for pdes on general geometries},
  author={Li, Zongyi and Huang, Daniel Zhengyu and Liu, Burigede and Anandkumar, Anima},
  journal={Journal of Machine Learning Research},
  volume={24},
  number={388},
  pages={1--26},
  year={2023}
}

@article{23:oFormer,
  title={Transformer for Partial Differential Equations’ Operator Learning},
  author={Li, Zijie and Meidani, Kazem and Farimani, Amir Barati},
  journal={Transactions on Machine Learning Research},
  year = {2023}
}

@inproceedings{23:GNOT,
  title={Gnot: A general neural operator transformer for operator learning},
  author={Hao, Zhongkai and Wang, Zhengyi and Su, Hang and Ying, Chengyang and Dong, Yinpeng and Liu, Songming and Cheng, Ze and Song, Jian and Zhu, Jun},
  booktitle={Proceedings of the 40th International Conference on Machine Learning},
  year={2023}
}

@article{23:nature,
  title={Scientific discovery in the age of artificial intelligence},
  author={Wang, Hanchen and Fu, Tianfan and Du, Yuanqi and Gao, Wenhao and Huang, Kexin and Liu, Ziming and Chandak, Payal and Liu, Shengchao and Van Katwyk, Peter and Deac, Andreea and others},
  journal={Nature},
  volume={620},
  number={7972},
  pages={47--60},
  year={2023},
  publisher={Nature Publishing Group UK London}
}

@article{23:uno,
  title={U-NO: U-shaped Neural Operators},
  author={Rahman, Md Ashiqur and Ross, Zachary E and Azizzadenesheli, Kamyar},
  journal={Transactions on Machine Learning Research},
  year = {2023}
}

@inproceedings{23:convno,
  title={Convolutional neural operators},
  author={Raonic, Bogdan and Molinaro, Roberto and Rohner, Tobias and Mishra, Siddhartha and de Bezenac, Emmanuel},
  booktitle={ICLR 2023 Workshop on Physics for Machine Learning},
  year={2023}
}

@article{24:schroedingerisation_JIN_PRL,
  title={Quantum Simulation of Partial Differential Equations via Schr{\"o}dingerization},
  author={Jin, Shi and Liu, Nana and Yu, Yue},
  journal={Physical Review Letters},
  volume={133},
  number={23},
  pages={230602},
  year={2024},
  publisher={American Physical Society}
}

@article{24:GINO,
  title={Geometry-informed neural operator for large-scale 3d pdes},
  author={Li, Zongyi and Kovachki, Nikola and Choy, Chris and Li, Boyi and Kossaifi, Jean and Otta, Shourya and Nabian, Mohammad Amin and Stadler, Maximilian and Hundt, Christian and Azizzadenesheli, Kamyar and others},
  journal={Advances in Neural Information Processing Systems},
  year={2024}
}

@article{24:analog_JIN,
  title={Analog quantum simulation of partial differential equations},
  author={Jin, Shi and Liu, Nana},
  journal={Quantum Science and Technology},
  volume={9},
  pages = {035047},
  year={2024}
}

@article{24:local_fno,
  title={Neural operators with localized integral and differential kernels},
  author={Liu-Schiaffini, Miguel and Berner, Julius and Bonev, Boris and Kurth, Thorsten and Azizzadenesheli, Kamyar and Anandkumar, Anima},
  journal={ArXiv},
  year={2024},
  volume={abs/2402.16845}
}

@article{24:blending,
  title={Blending neural operators and relaxation methods in PDE numerical solvers},
  author={Zhang, Enrui and Kahana, Adar and Kopani{\v{c}}{\'a}kov{\'a}, Alena and Turkel, Eli and Ranade, Rishikesh and Pathak, Jay and Karniadakis, George Em},
  journal={Nature Machine Intelligence},
  pages={1--11},
  year={2024},
  publisher={Nature Publishing Group UK London}
}

@inproceedings{24:ONO,
	title = {Improved {Operator} {Learning} by {Orthogonal} {Attention}},
	author = {Xiao, Zipeng and Hao, Zhongkai and Lin, Bokai and Deng, Zhijie and Su, Hang},
        booktitle={Proceedings of the 41st International Conference on Machine Learning},
	year = {2024}
}

@inproceedings{24:Transolver,
  title={Transolver: A Fast Transformer Solver for PDEs on General Geometries},
  author = {Wu, Haixu and Luo, Huakun and Wang, Haowen and Wang, Jianmin and Long, Mingsheng},
  booktitle={Proceedings of the 41st International Conference on Machine Learning},
  year={2024}
}

@article{24:laplace,
  title={Laplace neural operator for solving differential equations},
  author={Cao, Qianying and Goswami, Somdatta and Karniadakis, George Em},
  journal={Nature Machine Intelligence},
  volume={6},
  number={6},
  pages={631--640},
  year={2024},
  publisher={Nature Publishing Group UK London}
}

@inproceedings{25:AMG,
  title={Harnessing scale and physics: A multi-graph neural operator framework for pdes on arbitrary geometries},
  author={Li, Zhihao and Song, Haoze and Xiao, Di and Lai, Zhilu and Wang, Wei},
  booktitle={Proceedings of the 31st ACM SIGKDD Conference on Knowledge Discovery and Data Mining V. 1},
  pages={729--740},
  year={2025}
}

@article{18:neuralode,
  title={Neural ordinary differential equations},
  author={Chen, Ricky TQ and Rubanova, Yulia and Bettencourt, Jesse and Duvenaud, David K},
  journal={Advances in Neural Information Processing Systems},
  year={2018}
}

@inproceedings{23:CNN_PDE_solver,
  title={Clifford Neural Layers for PDE Modeling},
  author={Brandstetter, Johannes and van den Berg, Rianne and Welling, Max and Gupta, Jayesh K},
  booktitle={the 11th International Conference on Learning Representations},
  year = {2023}
}

@article{22:transformer_2,
  title={Learning operators with coupled attention},
  author={Kissas, Georgios and Seidman, Jacob H and Guilhoto, Leonardo Ferreira and Preciado, Victor M and Pappas, George J and Perdikaris, Paris},
  journal={Journal of Machine Learning Research},
  volume={23},
  number={215},
  pages={1--63},
  year={2022}
}

@article{24:boosting,
  title={Boosting Generalization in Parametric PDE Neural Solvers through Adaptive Conditioning},
  author={Kassa{\"\i} Koupa{\"\i}, Armand and Mifsut Benet, Jorge and Yin, Yuan and Vittaut, Jean-No{\"e}l and Gallinari, Patrick},
  journal={Advances in Neural Information Processing Systems},
  year={2024}
}

@article{20:era5,
  title={The ERA5 global reanalysis},
  author={Hersbach, Hans and Bell, Bill and Berrisford, Paul and Hirahara, Shoji and Hor{\'a}nyi, Andr{\'a}s and Mu{\~n}oz-Sabater, Joaqu{\'\i}n and Nicolas, Julien and Peubey, Carole and Radu, Raluca and Schepers, Dinand and others},
  journal={Quarterly journal of the royal meteorological society},
  volume={146},
  number={730},
  pages={1999--2049},
  year={2020},
  publisher={Wiley Online Library}
}

@article{24:thewell,
  title={The well: a large-scale collection of diverse physics simulations for machine learning},
  author={Ohana, Ruben and McCabe, Michael and Meyer, Lucas and Morel, Rudy and Agocs, Fruzsina and Beneitez, Miguel and Berger, Marsha and Burkhart, Blakesly and Dalziel, Stuart and Fielding, Drummond and others},
  journal={Advances in Neural Information Processing Systems},
  year={2024}
}

@article{32:wigner,
  title={On the quantum correction for thermodynamic equilibrium},
  author={Wigner, Eugene},
  journal={Physical review},
  volume={40},
  number={5},
  pages={749},
  year={1932},
  publisher={American Physical Society}
}

@inproceedings{22:afno,
  title={Adaptive fourier neural operators: Efficient token mixers for transformers},
  author={Guibas, John and Mardani, Morteza and Li, Zongyi and Tao, Andrew and Anandkumar, Anima and Catanzaro, Bryan},
  booktitle={the 10th International Conference on Learning Representations},
  year={2022}
}

@article{02:universal_nn_barron,
  title={Universal approximation bounds for superpositions of a sigmoidal function},
  author={Barron, Andrew R},
  journal={IEEE Transactions on Information theory},
  volume={39},
  number={3},
  pages={930--945},
  year={2002},
  publisher={IEEE}
}

@article{23:CNO,
  title={Convolutional neural operators for robust and accurate learning of PDEs},
  author={Raonic, Bogdan and Molinaro, Roberto and De Ryck, Tim and Rohner, Tobias and Bartolucci, Francesca and Alaifari, Rima and Mishra, Siddhartha and de B{\'e}zenac, Emmanuel},
  journal={Advances in Neural Information Processing Systems},
  year={2023}
}

@article{25:nonlocality,
  title={Nonlocality and nonlinearity implies universality in operator learning},
  author={Lanthaler, Samuel and Li, Zongyi and Stuart, Andrew M},
  journal={Constructive Approximation},
  pages={1--43},
  year={2025},
  publisher={Springer}
}

@article{86:backprop,
  title={Learning representations by back-propagating errors},
  author={Rumelhart, David E and Hinton, Geoffrey E and Williams, Ronald J},
  journal={nature},
  volume={323},
  number={6088},
  pages={533--536},
  year={1986},
  publisher={Nature Publishing Group UK London}
}

@inproceedings{25:LaMO,
  title={Latent Mamba Operator for Partial Differential Equations},
  author={Tiwari, Karn and Dutta, Niladri and Krishnan, NM Anoop and AP, Prathosh},
  booktitle={Proceedings of the 42nd International Conference on Machine Learning},
  year={2025}
}

@article{23:CoNO,
  title={Cono: Complex neural operator for continuous dynamical systems},
  author={Tiwari, Karn and Krishnan, NM and others},
  journal={ArXiv},
  year={2023},
  volume={abs/2310.02094}
}

@inproceedings{24:NMO,
  title={Neural manifold operators for learning the evolution of physical dynamics},
  author={Wu, Hao and Weng, Kangyu and Zhou, Shuyi and Huang, Xiaomeng and Xiong, Wei},
  booktitle={Proceedings of the 30th ACM SIGKDD Conference on Knowledge Discovery and Data Mining},
  pages={3356--3366},
  year={2024}
}

@inproceedings{23:LSM,
  title={Solving High-Dimensional PDEs with Latent Spectral Models},
  author={Wu, Haixu and Hu, Tengge and Luo, Huakun and Wang, Jianmin and Long, Mingsheng},
  booktitle={Proceedings of the 40th International Conference on Machine Learning},
  year={2023}
}

@inproceedings{25:quanti_NO_parabolic,
  title={Quantitative Approximation for Neural Operators in Nonlinear Parabolic Equations},
  author={Furuya, Takashi and Taniguchi, Koichi and Okuda, Satoshi},
    booktitle={the 13th International Conference on Learning Representations},
    year={2025}
}

@article{19:ANODE,
  title={Augmented neural odes},
  author={Dupont, Emilien and Doucet, Arnaud and Teh, Yee Whye},
  journal={Advances in Neural Information Processing Systems},
  year={2019}
}

@article{24:TFNO,
  title={Multi-Grid Tensorized Fourier Neural Operator for High-Resolution PDEs},
  author={Kossaifi, Jean and Kovachki, Nikola Borislavov and Azizzadenesheli, Kamyar and Anandkumar, Anima},
  journal={Transactions on Machine Learning Research},
  year={2024}
}

@inproceedings{23:FFNO,
  title={Factorized Fourier Neural Operators},
  author={Tran, Alasdair and Mathews, Alexander and Xie, Lexing and Ong, Cheng Soon},
  booktitle={the 11st International Conference on Learning Representations},
  year={2023}
}

@inproceedings{25:bregmanNO,
  title={A Bregman Proximal Viewpoint on Neural Operators},
  author={Mezidi, Abdel-Rahim and Patracone, Jordan and Salzo, Saverio and Habrard, Amaury and Pontil, Massimiliano and Emonet, R{\'e}mi and Sebban, Marc},
  booktitle={the 13th International Conference on Learning Representations},
  year={2025}
}

@article{25:yi_PINN,
  title={Transforming physics-informed machine learning to convex optimization},
  author={Yi, Letian and Yang, Siyuan and Cui, Ying and Lai, Zhilu},
  journal={Engineering Applications of Artificial Intelligence},
  volume={161},
  pages={112149},
  year={2025},
  publisher={Elsevier}
}

@article{89:Uncertain_prin,
  title={Uncertainty principles and signal recovery},
  author={Donoho, David L and Stark, Philip B},
  journal={SIAM Journal on Applied Mathematics},
  volume={49},
  number={3},
  pages={906--931},
  year={1989},
  publisher={SIAM}
}
